\documentclass[acmsmall,screen]{acmart}
\usepackage{booktabs} % For formal tables
\usepackage{multirow}
\usepackage{booktabs} % For formal tables
\usepackage{multirow}
\usepackage{subfigure}
\usepackage{array}
\usepackage{float}

\usepackage[utf8]{inputenc} % allow utf-8 input
\usepackage[T1]{fontenc}    % use 8-bit T1 fonts
\usepackage{hyperref}       % hyperlinks

\usepackage{setspace}
\usepackage{threeparttable}
\usepackage{supertabular}
\usepackage{bm}
\usepackage{amsmath}
\usepackage{amsthm}
\usepackage{mathrsfs}
\usepackage{booktabs}
\usepackage{siunitx}
\usepackage{footmisc}
\usepackage{footnote}
\usepackage{enumitem}
\usepackage{array}
\usepackage{CJK}
\usepackage{footnote}
\usepackage{enumitem}
\usepackage{tabularx}
\usepackage{colortbl}
\usepackage{tikz}
\usepackage[edges]{forest}

\usetikzlibrary{trees,positioning,shapes,shadows,arrows.meta}

\usepackage[edges]{forest}
\definecolor{hidden-draw}{RGB}{0,0,0}
\definecolor{hidden-blue}{RGB}{194,232,247}
\definecolor{hidden-orange}{RGB}{243,202,120}
\definecolor{hidden-yellow}{RGB}{242,244,193}
\definecolor{tree-level-1}{RGB}{245,20,85}
\definecolor{tree-level-2}{RGB}{246,86,118}
\definecolor{tree-level-3}{RGB}{248,177,193}
\definecolor{tree-leaf}{RGB}{176,230,198}

\AtBeginDocument{%
  \providecommand\BibTeX{{%
    \normalfont B\kern-0.5em{\scshape i\kern-0.25em b}\kern-0.8em\TeX}}}

\begin{document}

\title{LLMs-as-Judges: A Comprehensive Survey on LLM-based Evaluation Methods}

\author{Haitao Li}
\email{liht22@mails.tsinghua.edu.cn}
\affiliation{%
  \institution{Department of Computer Science and Technology, Institute for Internet Judiciary, Tsinghua University}
  \city{Beijing}
  \country{China}
}

\author{Qian Dong}
\email{dq22@mails.tsinghua.edu.cn}
\affiliation{%
  \institution{Department of Computer Science and Technology, Institute for Internet Judiciary, Tsinghua University}
  \city{Beijing}
  \country{China}
}

\author{Junjie Chen}
\email{chenjj826@gmail.com}
\affiliation{%
  \institution{Department of Computer Science and Technology, Institute for Internet Judiciary, Tsinghua University}
  \city{Beijing}
  \country{China}
}

\author{Huixue Su}
\email{suhuixue@ruc.edu.cn}
\affiliation{%
  \institution{Gaoling School of Artificial Intelligence, Renmin University of China}
  \city{Beijing}
  \country{China}
}

\author{Yujia Zhou}
\email{suhuixue@ruc.edu.cn}
\affiliation{%
  \institution{Department of Computer Science and Technology, Institute for Internet Judiciary, Tsinghua University}
  \city{Beijing}
  \country{China}
}

\author{Qingyao Ai}
\email{aiqy@tsinghua.edu.cn}
\affiliation{%
  \institution{Department of Computer Science and Technology, Institute for Internet Judiciary, Tsinghua University}
  \city{Beijing}
  \country{China}
}

\author{Ziyi Ye}
\email{yeziyi1998@gmail.com}
\affiliation{%
  \institution{Department of Computer Science and Technology, Institute for Internet Judiciary, Tsinghua University}
  \city{Beijing}
  \country{China}
}
\author{Yiqun Liu}
\email{yiqunliu@tsinghua.edu.cn}
\affiliation{%
  \institution{Department of Computer Science and Technology, Institute for Internet Judiciary, Tsinghua University}
  \city{Beijing}
  \country{China}
}

\renewcommand{\shortauthors}{Li, et al.}
\newcommand{\yzy}[1]{\textcolor{blue}{#1}}

\begin{abstract}

The rapid advancement of Large Language Models (LLMs) has driven their expanding application across various fields.
One of the most promising applications is their role as evaluators based on natural language responses, referred to as ``LLMs-as-judges''.
This framework has attracted growing attention from both academia and industry due to their excellent effectiveness, ability to generalize across tasks, and interpretability in the form of natural language.
This paper presents a comprehensive survey of the LLMs-as-judges paradigm from five key perspectives: \textbf{Functionality}, \textbf{Methodology}, \textbf{Applications}, \textbf{Meta-evaluation}, and \textbf{Limitations}.
We begin by providing a systematic definition of LLMs-as-Judges and introduce their functionality~(Why use LLM judges?).
Then we address methodology to construct an evaluation system with LLMs~(How to use LLM judges?).
Additionally, we investigate the potential domains for their application (Where to use LLM judges?) and discuss methods for evaluating them in various contexts~(How to evaluate LLM judges?).
Finally, we provide a detailed analysis of the limitations of LLM judges and discuss potential future directions.

Through a structured and comprehensive analysis, we aim aims to provide insights on the development and application of LLMs-as-judges in both research and practice. We will continue to maintain the relevant resource list at \url{https://github.com/CSHaitao/Awesome-LLMs-as-Judges}.
\end{abstract}


\maketitle

\section{Introduction}

Studies on evaluation methods have long been a key force in guiding the development of modern Artificial Intelligence (AI)~\cite{chang2024survey}. 
AI researchers have continuously sought to measure and validate the intelligence of AI models through various tasks~\cite{chang2024survey,guo2023evaluating}.
In the mid-20th century, AI evaluation primarily centered on assessing algorithm performance in specific tasks, such as logical reasoning and numerical computation~\cite{nilsson2014principles}. 
Traditional machine learning tasks like classification and regression often use programmable and statistical metrics, including accuracy, precision, and recall. 
With the emergence of deep learning, the complexity of AI systems grew rapidly, prompting a shift in evaluation standards~\cite{lecun2015deep}. 
The evaluation of AI has expanded from pre-defined, programmable machine metrics to more flexible, robust evaluators for solving complex, realistic tasks.
A typical example is the Turing Test~\cite{french2000turing,turing2009computing}, which determines whether an AI model can exhibit human-like intelligent behavior through dialogue with humans. 
The Turing Test provides a fundamental guideline in the evaluation of AI models, especially on AI models' intelligence in flexible and realistic environments. 
% emphasizing the evaluation of intelligence through realistic and complex tasks.

% As AI advances, evaluation methods have continuously evolved to keep pace with increasingly complex models and diverse applications~\cite{chang2024survey,guo2023evaluating}. 
Recently, the emergence of Large Language Models (LLMs) and generative AI serves as a new milestone in the evolution of AI evaluation.
LLMs exhibit remarkable generalization and adaptability, showcasing strong transfer capabilities across previously unseen tasks and diverse domains~\cite{achiam2023gpt,bai2023qwen}. However, their powerful capabilities also present new challenges for evaluation. 
Due to the highly generative and open-ended nature of their outputs, standardized metrics are often insufficient for a comprehensive evaluation.
For example, in natural language generation (NLG) tasks, traditional metrics like BLEU~\cite{papineni2002bleu} and ROUGE~\cite{lin2004rouge} often fail to capture key aspects such as text fluency, logical coherence, and creativity. Moreover, modern AI evaluation extends beyond task performance and must account for the ability to address complex, dynamic problems in real-world scenarios, including robustness, fairness, and interpretability.
Human annotations, frequently regarded as the ``ground truth,'' can offer comprehensive insights and valuable feedback. By gathering responses from experts or users, researchers can gain a deeper understanding of a model's performance, practicality, and potential risks. However, collecting them are typically time-consuming and resource-intensive, making it challenging to scale up for large-scale evaluation.

In this context, a new paradigm has emerged to replace humans and statistical metrics with LLMs in evaluation, referred to as LLMs-as-judges~\cite{ashktorab2024aligning,tseng2024expert,bavaresco2024llms,bavaresco2024llms}.
% The key idea is that since LLMs have already been so powerful in langauge understanding and reasoning, they should be able to serve as the evaluator of other AI models or systems~\cite{ashktorab2024aligning,tseng2024expert,bavaresco2024llms,bavaresco2024llms}.
Compared to traditional evaluation methods, LLMs-as-judges show significant strengths. 
First, LLM judges can adjust their evaluation criteria based on the specific task context, rather than relying on a fixed set of metrics, making the evaluation process more flexible and refined. 
Second, LLM judges can generate interpretive evaluations, offering more comprehensive feedback on model performance and enabling researchers to gain deeper insights into the evaluater's strengths and weaknesses. 
Finally, LLM judges offer a scalable and reproducible alternative to human evaluation, significantly reducing the costs and time associated with human involvement.

Despite its great potential and significant advantages, LLMs-as-judges also face several critical challenges.
For example, the evaluation results of LLMs are often influenced by the prompt template, which can lead to biased or inconsistent assessments~\cite{xu2023llm}. 
Considering that LLMs are trained on extensive text corpus, they may also inherit various implicit biases, impacting the fairness and reliability of their assessments~\cite{ye2024justice}. 
Moreover, distinct tasks and domains require specific evaluation criteria, making it difficult for LLMs to adapt their standards dynamically to specific contexts.

Considering the vast potential of this field, 
this survey aims to systematically review and analyze the current state and key challenges of the LLMs-as-judges. 
As shown in Figures \ref{Taxonomy_1} and \ref{Taxonomy_2}, we discuss existing research across five key perspectives: 1) \textbf{Functionality}: Why use LLM judges, 2) \textbf{Methodology}: How to use LLM judges, 3) \textbf{Application}: Where to use LLM judges, 4) \textbf{Meta-evaluation}: How to evaluate LLM judges and 5) \textbf{Limitation}: Existing problems of LLM judges. We explore the key challenges confronting LLMs-as-judges and hope to provide a clearer guideline for their future development.

In summary, the main contributions of this paper are as follows:

\begin{enumerate} 
\item \textbf{Comprehensive and Timely Survey}: We present the extensive survey on the emerging paradigm of LLMs-as-judges, systematically reviewing the current state of research and developments in this field. By examining LLMs as performance evaluators based on their generated natural language, we highlight the unique role of LLMs in shaping the future of AI evaluation.

\item \textbf{Systematic Analysis Across Five Key Perspectives}: We organize our survey around five critical aspects: Functionality, Methodology, Application, Meta-evaluation, and Limitation. This structured approach allows for a nuanced understanding of how and why LLMs are utilized as evaluators, their practical implementations, and reliability concerns.

\item \textbf{Current Challenges and Future Research Directions}: We discuss the existing challenges for adopting LLMs-as-judges, highlighting potential research opportunities and directions while offering a forward-looking perspective on the future development of this paradigm, encouraging researchers to delve deeper into this exciting area. We also provide an open-source repository at \url{https://github.com/CSHaitao/Awesome-LLMs-as-Judges}, with the goal of fostering a collaborative community and advancing best practices in this area.
\end{enumerate}

The organization of this paper is as follows. In Section (\S\ref{sec:PRELIMINARIES}), we provide the formal definition of LLMs-as-judges. Then, Section (\S\ref{sec:Functionality}) reviews existing work from the perspective of ``Why use LLM judges''. Following that, Section (\S\ref{sec:Methodology}) covers ``How to use LLM judges'', summarizing the current technical developments in LLMs-as-judges. Section (\S\ref{sec:Domain}) discusses ``Where to use LLM judges'', focusing on their application domains. In Section (\S\ref{sec:Meta}), we review the metrics and benchmarks used for evaluating LLMs-as-judges. Section (\S\ref{sec:Limitation}) discusses the limitations and challenges of LLM judges. We discuss major future work in Sections (\S\ref{sec:Future})  and (\S\ref{sec:Conclusion}) to conclude the paper. 
% In Figures \ref{Taxonomy_1} and \ref{Taxonomy_2}, we present the taxonomy of LLMs-as-Judges.

\tikzstyle{my-box}= [
    rectangle,
    draw=hidden-draw,
    rounded corners,
    text opacity=1,
    minimum height=1.5em,
    minimum width=5em,
    inner sep=2pt,
    align=center,
    fill opacity=.5,
]
\tikzstyle{leaf}=[my-box, minimum height=1.5em,
    fill=blue!15, text=black, align=left,font=\scriptsize,
    inner xsep=2pt,
    inner ysep
=4pt,
]
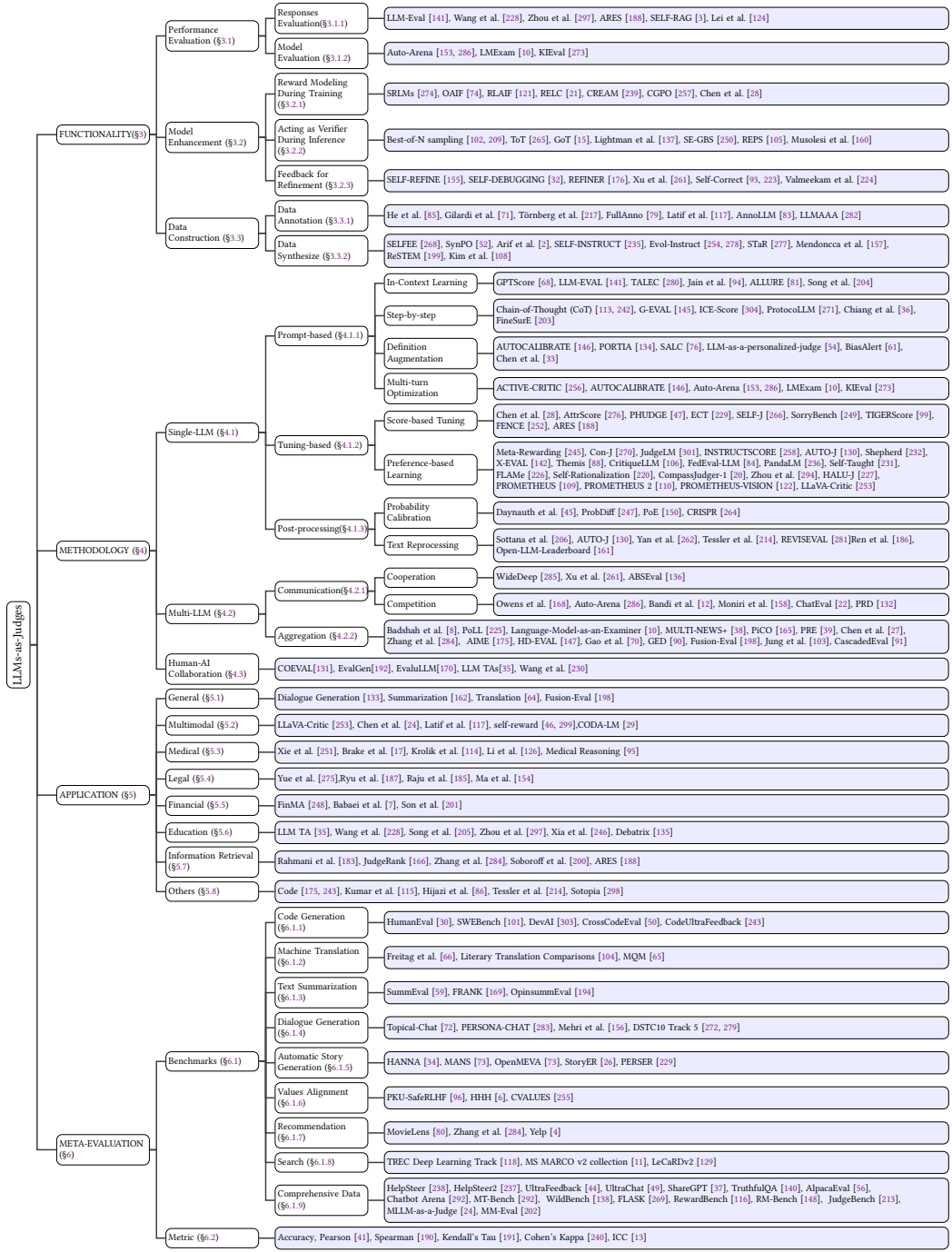
\begin{figure*}[t]

    \centering
    \resizebox{\textwidth}{!}{
        \begin{forest}
            forked edges,
            for tree={
                grow=east,
                reversed=true,
                anchor=base west,
                parent anchor=east,
                child anchor=west,
                base=left,
                font=\small,
                rectangle,
                draw=hidden-draw,
                rounded corners,
                align=left,
                minimum width=4em,
                edge+={darkgray, line width=1pt},
                s sep=3pt,
                inner xsep=2pt,
                inner ysep=3pt,
                ver/.style={rotate=90, child anchor=north, parent anchor=south, anchor=center},
            },
            where level=1{text width=6.4em,font=\scriptsize,}{},
            where level=2{text width=6.4em,font=\scriptsize,}{},
            where level=3{text width=6.4em,font=\scriptsize,}{},
            where level=4{text width=6.4em,font=\scriptsize,}{},
[LLMs-as-Judges, ver
    [FUNCTIONALITY(\S \ref{sec:Functionality})
        [Performance \\Evaluation (\S \ref{sec:Performance Evaluation})
            [Responses \\Evaluation(\S \ref{sec:Responses Evaluation})
                [LLM-Eval~\cite{lin2023llm}{, }Wang et al.~\cite{wang2024automated}{, }Zhou et al.~\cite{zhou2024llm}{, }ARES~\cite{saad2023ares}{, }SELF-RAG~\cite{asai2023self}{, }Lei et al.~\cite{lei2024recexplainer}, leaf, text width=41em]]
            [Model \\Evaluation (\S \ref{sec:Model Evaluation})
                [Auto-Arena~\cite{zhao2024auto,luo2024videoautoarena}{, }LMExam~\cite{bai2024benchmarking}{, }KIEval~\cite{yu2024kieval}, leaf, text width=41em]]]
        [Model \\Enhancement (\S \ref{sec:Model Enhancement})
            [Reward Modeling \\During Training \\ (\S \ref{sec: Reward Modeling During Training})
                [SRLMs~\cite{yuan2024self}{, }OAIF~\cite{guo2024direct}{, }RLAIF~\cite{lee2023rlaif}{, }RELC~\cite{cao2024enhancing}{, }CREAM~\cite{wang2024cream}{, }CGPO~\cite{xu2024perfect}{, }Chen et al.~\cite{chen2023adaptation}, leaf, text width=41em]]
            [Acting as Verifier \\During Inference \\(\S \ref{sec: Acting as Verifier During Inference})
                [Best-of-N sampling~\cite{jinnai2024regularized,sun2024fast}{, }ToT~\cite{yao2024tree}{, }GoT~\cite{besta2024graph}{, }Lightman et al.~\cite{lightman2023let}{, }SE-GBS~\cite{xie2024self}{, }REPS~\cite{kawabata2024rationale}{, }Musolesi et al.~\cite{musolesi2024creative}, leaf, text width=41em]]
            [Feedback for \\Refinement (\S \ref{sec: Feedback for Refinement})
                [SELF-REFINE~\cite{madaan2024self}{, }SELF-DEBUGGING~\cite{chen2023teaching}{, }REFINER~\cite{paul2023refiner}{, }Xu et al.~\cite{xu2023towards}{, }Self-Correct~\cite{huang2023large,tyen2023llms}{, }Valmeekam et al.~\cite{valmeekam2023can}, leaf, text width=41em]]]
        [Data \\ Construction  (\S \ref{sec:Data Construction})
           [Data \\Annotation (\S \ref{sec:Data Annotation})
                [He et al.~\cite{he2024if}{, }Gilardi et al.~\cite{gilardi2023chatgpt}{, }Törnberg et al.~\cite{tornberg2023chatgpt}{, }FullAnno~\cite{hao2024fullanno}{, }Latif et al.~\cite{latif2023can}{, }AnnoLLM~\cite{he2023annollm}{, }LLMAAA~\cite{zhang2023llmaaa}, leaf, text width=41em]]
           [Data \\Synthesize (\S \ref{sec:Data Synthesize})
               [SELFEE~\cite{ye2023selfee}{, }SynPO~\cite{dong2024self}{, }Arif et al.~\cite{arif2024fellowship}{, }SELF-INSTRUCT~\cite{wang2022self}{, }Evol-Instruct~\cite{xu2023wizardlm,zeng2024automatic}{, }STaR~\cite{zelikman2024star}{, }Mendoncca et al.~\cite{mendoncca2024soda}{, }\\ReSTEM~\cite{singh2023beyond}{, }Kim et al.~\cite{kim2024aligning}, leaf, text width=41em]]]]
    [
    METHODOLOGY (\S \ref{sec:Methodology})
    [
        Single-LLM  (\S \ref{sec:Single-LLM System})
        [
            Prompt-based (\S \ref{sec:Prompt Engineering})
            [
                In-Context Learning
                [
                    GPTScore~\cite{fu2023gptscore}{, }LLM-EVAL~\cite{lin2023llm}{, }TALEC~\cite{zhang2024talec}{, }Jain et al.~\cite{jain2023multi}{, }ALLURE~\cite{hasanbeig2023allure}{, }Song et al.~\cite{song2024can}, leaf, text width=33em
                ]
            ]
            [
                Step-by-step
                [
                    Chain-of-Thought (CoT)~\cite{wei2022chain,kotonya2023little}{, }G-EVAL~\cite{liu2023g}{, }ICE-Score~\cite{zhuo2023ice}{, }ProtocoLLM~\cite{yi2024protocollm}{, }Chiang et al.~\cite{chiang2023closer}{, }\\FineSurE~\cite{song2024finesure}, leaf, text width=33em
                ]
            ]
            [
                Definition \\Augmentation 
                [
                    AUTOCALIBRATE~\cite{liu2023calibrating}{, }PORTIA~\cite{li2023split}{, }SALC~\cite{gupta2024unveiling}{, }LLM-as-a-personalized-judge~\cite{dong2024can}{, }BiasAlert~\cite{fan2024biasalert}{, }\\Chen et al.~\cite{chen2024llms}, leaf, text width=33em
                ]
            ]
            [
                Multi-turn \\Optimization
                [
                    ACTIVE-CRITIC~\cite{xu2024large}{, }AUTOCALIBRATE~\cite{liu2023calibrating}{, }Auto-Arena~\cite{zhao2024auto,luo2024videoautoarena}{, }LMExam~\cite{bai2024benchmarking}{, }KIEval~\cite{yu2024kieval}, leaf, text width=33em
                ]
            ]]
            [
                Tuning-based (\S \ref{sec:Tuning})
                [
                    Score-based Tuning
                    [
                        Chen et al.~\cite{chen2023adaptation}{, }AttrScore~\cite{yue2023automatic}{, }PHUDGE~\cite{deshwal2024phudge}{, }ECT~\cite{wang2023learning}{, }SELF-J~\cite{ye2024self}{, }SorryBench~\cite{xie2024sorry}{, }TIGERScore~\cite{jiang2023tigerscore}{, }\\FENCE~\cite{xie2024improving}{, }ARES~\cite{saad2023ares}, leaf, text width=33em
                    ]
                ]
                [
                    Preference-based \\Learning
                    [
                        Meta-Rewarding~\cite{wu2024meta}{, }Con-J~\cite{ye2024beyond}{, }JudgeLM~\cite{zhu2023judgelm}{, }INSTRUCTSCORE~\cite{xu2023instructscore}{, }AUTO-J~\cite{li2023generative}{, }Shepherd~\cite{wang2023shepherd}{, }\\X-EVAL~\cite{liu2023x}{, }Themis~\cite{hu2024themis}{, }CritiqueLLM~\cite{ke2024critiquellm}{, }FedEval-LLM~\cite{he2024fedeval}{, }PandaLM~\cite{wang2023pandalm}{, }Self-Taught~\cite{wang2024self}{, }\\FLAMe~\cite{vu2024foundational}{, }Self-Rationalization~\cite{trivedi2024self}{, }CompassJudger-1~\cite{cao2024compassjudger}{, }Zhou et al.~\cite{zhou2024mitigating}{, }HALU-J~\cite{wang2024halu}{, }\\PROMETHEUS~\cite{kim2023prometheus}{, }PROMETHEUS 2~\cite{kim2024prometheus}{, }PROMETHEUS-VISION~\cite{lee2024prometheusvision}{, }LLaVA-Critic~\cite{xiong2024llava}, leaf, text width=33em
                    ]
                ]
            ]
            [
                Post-processing(\S \ref{sec:Post-processing})
                [
                    Probability \\Calibration
                    [
                        Daynauth et al.~\cite{daynauth2024aligning}{, }ProbDiff~\cite{xia2024language}{, }PoE~\cite{liusie2024efficient}{, }CRISPR~\cite{yang2024mitigating}, leaf, text width=33em
                    ]
                ]
                [
                    Text Reprocessing
                    [
                        Sottana et al.~\cite{sottana2023evaluation}{, }AUTO-J~\cite{li2023generative}{, }Yan et al.~\cite{yan2024consolidating}{, }Tessler et al.~\cite{tessler2024ai}{, }REVISEVAL~\cite{zhang2024reviseval}Ren et al.~\cite{ren2023self}{, }\\Open-LLM-Leaderboard~\cite{myrzakhan2024open}, leaf, text width=33em
                    ]
                ]
            ]]
        [
            Multi-LLM (\S \ref{sec:Multi-LLM System})
            [
                Communication(\S \ref{sec:Communication})
                [
                    Cooperation
                    [
                        WideDeep~\cite{zhang2023wider}{, }Xu et al.~\cite{xu2023towards}{, }ABSEval~\cite{liang2024abseval}, leaf, text width=33em
                    ]
                ]
                [
                    Competition
                    [
                        Owens et al.~\cite{owens2024multi}{, }Auto-Arena~\cite{zhao2024auto}{, }Bandi et al.~\cite{bandi2024adversarial}{, }Moniri et al.~\cite{moniri2024evaluating}{, }ChatEval~\cite{chan2023chateval}{, }PRD~\cite{li2023prd}, leaf, text width=33em
                    ]
                ]
            ]
            [ 
                Aggregation (\S \ref{sec:Aggregation})
                [
                    Badshah et al.~\cite{badshah2024reference}{, }PoLL~\cite{verga2024replacing}{, }Language-Model-as-an-Examiner~\cite{bai2024benchmarking}{, }MULTI-NEWS+~\cite{choi2024multi}{, }PiCO~\cite{ning2024pico}{, }PRE~\cite{chu2024pre}{, }Chen et al. ~\cite{chen2024automaticcostefficientpeerreviewframework}{, }\\Zhang et al.~\cite{zhang2024large}{, }
                    AIME~\cite{patel2024aime}{, }HD-EVAL~\cite{liu2024hd}{, }Gao et al.~\cite{gao2024bayesian}{, }GED~\cite{hu2024language}{, }Fusion-Eval~\cite{shu2024fusion}{, }Jung et al.~\cite{jung2024trust}{, }CascadedEval~\cite{huang2024empirical}, leaf, text width=41em
                ]
            ]]
        [
            Human-AI \\ Collaboration (\S \ref{sec:Hybrid System})
            [
                COEVAL\cite{li2023collaborative}{, }EvalGen\cite{shankar2024validates}{, }EvaluLLM\cite{pan2024human}{, }LLM TAs\cite{chiang2024large}{, }Wang et al.~\cite{wang2023large}, leaf, text width=49em
            ]
            ]
        ]
    [APPLICATION (\S \ref{sec:Domain})
        [
            General (\S \ref{sec:General})
            [
                Dialogue Generation~\cite{li2017dailydialog}{, }Summarization~\cite{narayan2018don}{, }Translation~\cite{feng2024improving}{, }Fusion-Eval~\cite{shu2024fusion}, leaf, text width=49em
            ]
        ]
        [
            Multimodal (\S \ref{sec:Multimodal})
            [
                LLaVA-Critic~\cite{xiong2024llava}{, }Chen et al.~\cite{chen2024mllm}{, }Latif et al.~\cite{latif2023can}{, }self-reward~\cite{zhou2024calibrated, deng2024efficient}{,}CODA-LM~\cite{chen2024automated}, leaf, text width=49em
            ]
        ]
        [
            Medical (\S \ref{sec:Medical})
            [ 
                Xie et al.~\cite{xie2024doclens}{, }Brake et al.~\cite{brake2024comparing}{, }Krolik et al.~\cite{krolik2024towards}{, }Li et al.~\cite{li2024automatic}{, }Medical Reasoning~\cite{jeong2024improving}, leaf, text width=49em
            ]
        ]
        [
            Legal (\S \ref{sec:Legal})
            [
                Yue et al.~\cite{yue2023disc}{,}Ryu et al.~\cite{ryu2023retrieval}{, }Raju et al.~\cite{raju2024constructing}{, }Ma et al.~\cite{ma2024leveraging}, leaf, text width=49em
            ]
        ]
        [
            Financial (\S \ref{sec:Financial})
            [
                FinMA~\cite{xie2023pixiu}{, }Babaei et al.~\cite{babaei2024gpt}{, }Son et al.~\cite{son2024krx}, leaf, text width=49em
            ]
        ]
        [
            Education (\S \ref{sec:Education})
            [
                LLM TA~\cite{chiang2024large}{, }Wang et al.~\cite{wang2024automated}{, }Song et al.~\cite{song2024automated}{, }Zhou et al.~\cite{zhou2024llm}{, }Xia et al.~\cite{xia2024evaluating}{, }Debatrix~\cite{liang2024debatrix}, leaf, text width=49em
            ]
        ]
        [
            Information Retrieval \\(\S \ref{sec: Information Retrieval})
            [
                Rahmani et al.~\cite{rahmani2024llmjudge}{, }JudgeRank~\cite{niu2024judgerank}{, }Zhang et al.~\cite{zhang2024large}{, }Soboroff et al. ~\cite{soboroff2024don}{, }ARES~\cite{saad2023ares}, leaf, text width=49em
            ]
        ]
        [
            Others (\S \ref{sec: Others})
            [
                Code~\cite{patel2024aime, weyssow2024codeultrafeedback}{, }Kumar et al.~\cite{kumar2024llms}{, }Hijazi et al.~\cite{hijazi2024using}{, }Tessler et al.~\cite{tessler2024ai}{, }Sotopia~\cite{zhou2023sotopia}, leaf, text width=49em
            ]
        ]]
    [META-EVALUATION \\ (\S \ref{sec:Meta})
        [Benchmarks (\S \ref{sec:benchmarks})
        [
            Code Generation \\(\S \ref{sec:Code Generation})
            [
                HumanEval~\cite{chen2021evaluating}{, }SWEBench~\cite{jimenez2023swe}{, }DevAI~\cite{zhuge2024agent}{, }CrossCodeEval~\cite{ding2024crosscodeeval}{, }CodeUltraFeedback~\cite{weyssow2024codeultrafeedback}, leaf, text width=41em
            ]
        ]
        [
            Machine Translation \\(\S \ref{sec:Machine Translation})
            [
                Freitag et al.~\cite{freitag2021results}{, }Literary Translation Comparisons~\cite{karpinska2023large}{, }MQM~\cite{freitag2021experts}, leaf, text width=41em
            ]
        ]
        [
            Text Summarization \\(\S \ref{sec:Text Summarization})
            [
                SummEval~\cite{fabbri2021summeval}{, }FRANK~\cite{pagnoni2021understanding}{, }OpinsummEval~\cite{shen2023opinsummeval}, leaf, text width=41em
            ]
        ]
        [
            Dialogue Generation \\(\S \ref{sec:Dialogue Generation})
            [
                Topical-Chat~\cite{gopalakrishnan2023topical}{, }PERSONA-CHAT~\cite{zhang2018personalizing}{, }Mehri et al.~\cite{mehri2020usr}{, }DSTC10 Track 5~\cite{yoshino2023overview,zhang2021automatic}, leaf, text width=41em
            ]
        ]
        [
            Automatic Story \\Generation (\S \ref{sec:Automatic Story Generation})
            [ 
                HANNA~\cite{chhun2022human}{, }MANS~\cite{guan2021openmeva}{, }OpenMEVA~\cite{guan2021openmeva}{, }StoryER~\cite{chen2023storyer}{, }PERSER~\cite{wang2023learning}, leaf, text width=41em
            ]
        ]
        [
            Values Alignment \\(\S \ref{sec:Values Alignment})
            [
                PKU-SafeRLHF~\cite{ji2024pku}{, }HHH~\cite{askell2021general}{, }CVALUES~\cite{xu2023cvalues}, leaf, text width=41em
            ]
        ]
        [
            Recommendation \\(\S \ref{sec:Recommendation})
            [
                MovieLens~\cite{harper2015movielens}{, }Zhang et al.~\cite{zhang2024large}{, }Yelp~\cite{asghar2016yelp}, leaf, text width=41em
            ]
        ]
        [
            Search (\S \ref{sec:Search})
            [
                TREC Deep Learning Track~\cite{lawrie2024overview}{, }MS MARCO v2 collection~\cite{bajaj2016ms}{, }LeCaRDv2~\cite{li2024lecardv2}, leaf, text width=41em
            ]
        ]
        [
            Comprehensive Data \\(\S \ref{sec:Comprehensive Data})
            [
                HelpSteer~\cite{wang2023helpsteer}{, }HelpSteer2~\cite{wang2024helpsteer2}{, }UltraFeedback~\cite{cui2024ultrafeedback}{, }UltraChat~\cite{ding2023enhancing}{, }ShareGPT~\cite{chiang2023vicuna}{, }TruthfulQA~\cite{lin2021truthfulqa}{, }AlpacaEval~\cite{dubois2024length}{, }\\Chatbot Arena~\cite{zheng2023judging}{, }MT-Bench~\cite{zheng2023judging}{, }
                WildBench~\cite{lin2024wildbench}{, }FLASK~\cite{ye2023flask}{, }RewardBench~\cite{lambert2024rewardbench}{, }RM-Bench~\cite{liu2024rm}{, }
                JudgeBench~\cite{tan2024judgebench}{, }\\MLLM-as-a-Judge~\cite{chen2024mllm}{, }MM-Eval~\cite{son2024mm}, leaf, text width=41em
            ]
        ]]
        [Metric (\S \ref{sec:metric})
            [Accuracy{, }Pearson~\cite{cohen2009pearson}{, }Spearman~\cite{sedgwick2014spearman}{, }Kendall's Tau~\cite{sen1968estimates}{, }Cohen's Kappa~\cite{warrens2015five}{, }ICC~\cite{bartko1966intraclass}, leaf, text width=49em ]]]]
        \end{forest}
    }
\caption{Taxonomy of LLMs-as-judges in functionality, methodology, application, meta-evaluation.}
\label{Taxonomy_1}
\end{figure*}

\tikzstyle{my-box}= [
    rectangle,
    draw=hidden-draw,
    rounded corners,
    text opacity=1,
    minimum height=1.5em,
    minimum width=5em,
    inner sep=2pt,
    align=center,
    fill opacity=.5,
]
\tikzstyle{leaf}=[my-box, minimum height=1.5em,
    fill=blue!15, text=black, align=left,font=\scriptsize,
    inner xsep=2pt,
    inner ysep
=4pt,
]
\begin{figure*}[t]

    \centering
    \resizebox{\textwidth}{!}{
        \begin{forest}
            forked edges,
            for tree={
                grow=east,
                reversed=true,
                anchor=base west,
                parent anchor=east,
                child anchor=west,
                base=left,
                font=\small,
                rectangle,
                draw=hidden-draw,
                rounded corners,
                align=left,
                minimum width=4em,
                edge+={darkgray, line width=1pt},
                s sep=3pt,
                inner xsep=2pt,
                inner ysep=3pt,
                ver/.style={rotate=90, child anchor=north, parent anchor=south, anchor=center},
            },
            where level=1{text width=6.4em,font=\scriptsize,}{},
            where level=2{text width=6.4em,font=\scriptsize,}{},
            where level=3{text width=6.4em,font=\scriptsize,}{},
            where level=4{text width=6.4em,font=\scriptsize,}{},
[LLMs-as-judges, ver
    [LIMITATION (\S \ref{sec:Limitation})
        [Biases (\S\ref{sec:biases})
            [Presentation-Related \\ (\S\ref{sec:presentationbiases})
                [Position bias~\cite{blunch1984position,raghubir2006center, ko2020look, wang2018position, llmsjuding2024openreview, zheng2023judging,chen2024humans,wang2023large,li2023generative,zheng2023large, raina2024llm,hou2024large, li2023split, li2023prd,khan2024debating, zhou2023batch, li2024calibraeval,  shi2024judging, stureborg2024large, zhao2024measuring}{, }Verbosity bias~\cite{nasrabadi2024juree, ye2024justice, ye2024beyond}, leaf, text width=41em] ]
            [Social-Related (\S\ref{sec:socialbiases})
                [Authority bias~\cite{chen2024humans,ye2024justice, zhao2023mind}{, }Bandwagon-effect bias~\cite{koo2023benchmarking,ye2024justice}{, }Compassion-fade bias~\cite{koo2023benchmarking,ye2024justice}{, }Diversity bias~\cite{chen2024humans,ye2024justice}, leaf, text width=41em] ]
            [Content-Related \\(\S\ref{sec:contentbiases})
                [Sentiment bias~\cite{ye2024justice}{, }Token Bias~\cite{jiang2024peek,li2024calibraeval,pezeshkpour2023large,raina2024llm}{, }Contextual Bias~\cite{poulain2024bias, zhou2024large,zhou2023batch,fei2023mitigating,zhao2021calibrate,han2022prototypical}, leaf, text width=41em] ]
            [Cognitive-Related \\(\S\ref{sec:cognitivebiases})
                [Overconfidence bias~\cite{khan2024debating,jung2024trust}{, }Self-enhancement bias~\cite{liu2023g,zheng2023judging,li2023prd,liu2023g, brown1986evaluations,ye2024justice, badshah2024reference}{, }Refinement-aware bias~\cite{ye2024justice, xu2024pride}\\Distraction bias~\cite{ye2024justice,koo2023benchmarking,shi2023large}{, }Fallacy-oversight bias~\cite{chen2024humans,ye2024justice}, leaf, text width=41em] ] ]
        [Adversarial Attacks \\(\S\ref{sec:attacks})
            [Adversarial Attacks \\on LLMs (\S \ref{sec:Adversarial Attacks on LLMs})
                [Text-Level Manipulations~\cite{ebrahimi2017hotflip, jiang2023prompt, branch2022evaluating, perez2022ignore}{, }Structural and Semantic Distortions~\cite{xu2023llm}{, }Optimization-Based Attacks~\cite{sun2020natural,sun2020adv, lee2022query}, leaf, text width=41em] ]
            [Adversarial Attacks \\on LLMs-as-Judges \\(\S \ref{sec:Adversarial Attacks on LLMs-as-judges})
                [Zheng et al.~\cite{zheng2024cheating}{, }Doddapaneni et al.~\cite{doddapaneni2024finding}{, }MT-Bench~\cite{zheng2023judging}{, }Raina et al.~\cite{raina2024llm}{, }Shi et al.~\cite{shi2024optimization}, leaf, text width=41em] ]  ]
        [Inherent Weaknesses \\(\S\ref{sec:weaknesses}) 
            [Knowledge Recency \\(\S \ref{sec:Knowledge Recency})
                [Zhao et al. ~\cite{zhao2023survey}{, }Luo et al.~\cite{luo2023empirical}{, }Gao et al.~\cite{gao2023retrieval}{, }Lewis et al.~\cite{lewis2020retrieval}{, }Wu et al.~\cite{wu2024continual}{, }Dierickx et al. ~\cite{dierickx2024striking}, leaf, text width=41em]]
            [Hallucination (\S \ref{sec:Hallucination})
                [Dierickx et al.~\cite{dierickx2024striking}{, }Ji et al.~\cite{ji2023survey}{, }Tonmoy et al.~\cite{tonmoy2024comprehensive}, leaf, text width=41em]]
            [Domain-Specific \\Knowledge Gaps \\(\S \ref{sec:Domain-Specific Knowledge Gaps})
                [Feng et al.~\cite{feng2023knowledge}{, }Pan et al.~\cite{pan2024unifying}{, }Gao et al.~\cite{gao2023retrieval}{, }Szymanski et al.~\cite{szymanski2024limitations}{, }Dorner et al.~\cite{dorner2024limitsscalableevaluationfrontier}, leaf, text width=41em]] ]]
[
    FUTURE WORK \\(\S\ref{sec:Future})
    [
        More Efficient (\S\ref{sec:more-Efficient})
        [
            Automated Construction of Evaluation Criteria and Tasks~\cite{bai2024benchmarking,zhao2024auto,yu2024kieval,zhang2024talec,wang2024revisiting}{, }Scalable Evaluation Systems~\cite{xu2024perfect}{, }Accelerating Evaluation Processes~\cite{lee2024aligning,liu2024aligning, chen2024internet}, leaf, text width=49em
        ]
    ]
    [
        More Effective (\S\ref{sec:more-Effective})
        [
            Integration of Reasoning and Judge Capabilities~\cite{zhuo2023ice,yi2024protocollm,stephan2024calculation}{, }Establishing a Collective Judgment Mechanism~\cite{chan2023chateval,chu2024pre}{, }Enhancing Domain Knowledge~\cite{raju2024constructing}{, }\\Cross-Domain and Cross-Language Transferability~\cite{son2024mm,hada2023large,watts2024pariksha}{, }Multimodal Integration Evaluation~\cite{chen2024mllm}, leaf, text width=49em
        ]
    ]
    [
        More Reliable (\S\ref{sec:more-Reliable})
        [
            Enhancing Interpretability and Transparency~\cite{liu2024hd}{, }
            Mitigating Bias and Ensuring Fairness~\cite{li2024calibraeval}{, }
            Enhancing Robustness~\cite{shi2024optimization,elangovan2024beyond}, leaf, text width=49em
        ]]]]
        \end{forest}
    }
    \caption{Taxonomy of LLMs-as-judges in limitation and future work.}
    \label{Taxonomy_2}
\end{figure*}
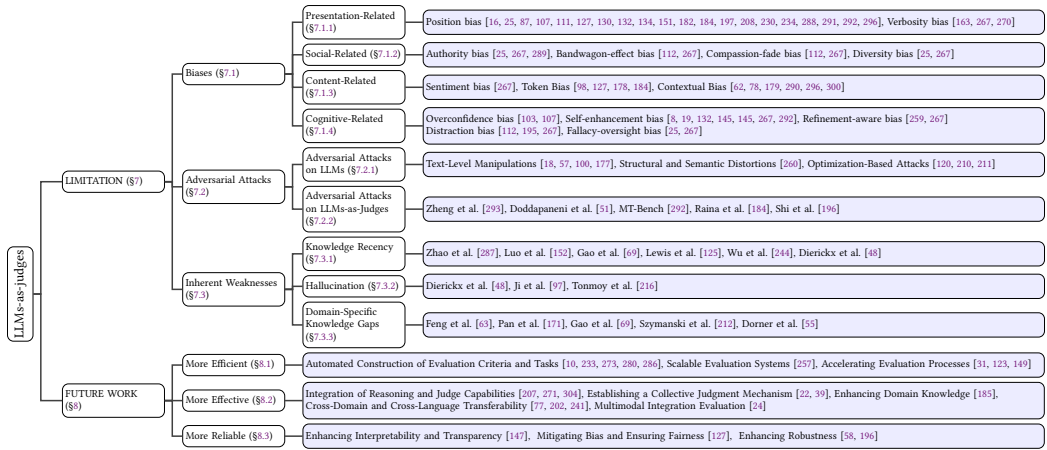
\section{PRELIMINARIES}
\label{sec:PRELIMINARIES}
\begin{figure}[t]
\includegraphics[width=\linewidth]{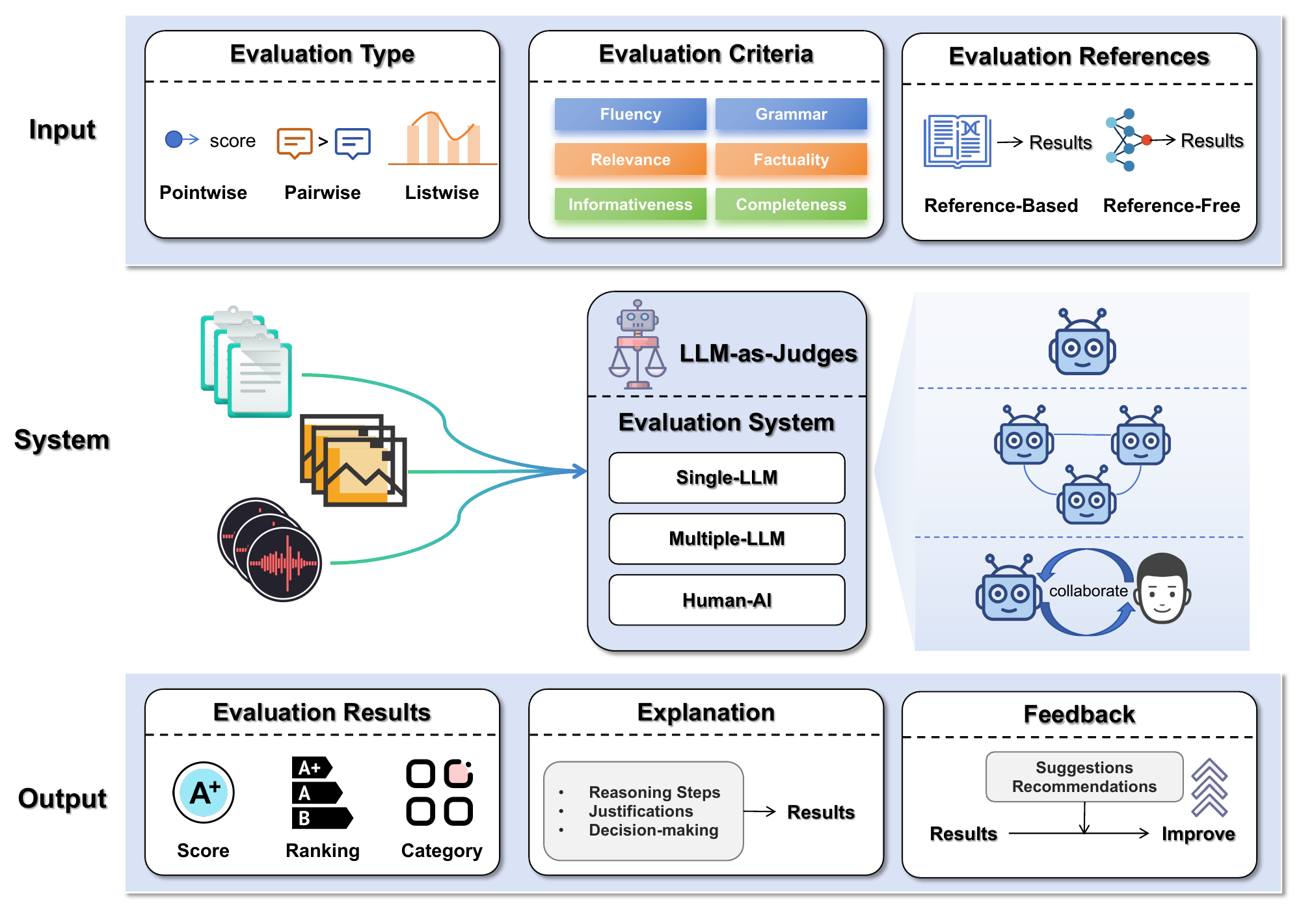}
\caption{Overview of the LLMs-as-judges system.}
% \vspace{-5mm}
\label{figure:overreview}
\end{figure}

In this section, we will provide a formal definition of LLMs-as-judges, aiming to encompass all current evaluation paradigms and methods, thereby offering readers a clear and thorough understanding. Figure \ref{figure:overreview} presents an overview of the LLMs-as-judges system.

The LLMs-as-judges paradigm is a flexible and powerful evaluation framework where LLMs are employed as evaluative tools, responsible for assessing the quality, relevance, and effectiveness of generated outputs based on defined evaluation criteria. This framework leverages the extensive knowledge and deep contextual understanding of LLMs, enabling it to flexibly adapt to various tasks in NLP and machine learning.
We formalize the input-output structure of the LLMs-as-Judges paradigm, unifying various evaluation scenarios into a unified perspective. 
Specifically, the evaluation process can be defined as follows:
\begin{equation}\label{eqn-1} 
(\mathcal{Y}, \mathcal{E}, \mathcal{F})=E(\mathcal{T}, \mathcal{C}, \mathcal{X}, \mathcal{R})
\end{equation} 
where $E$ is the evaluation function, taking the evaluation type $\mathcal{T}$, evaluation criteria $\mathcal{C}$, evaluation item $\mathcal{X}$ and optional references $\mathcal{R}$ as input. Based on these inputs, the LLM can produces three outputs: evaluation result $\mathcal{Y}$, explanation $\mathcal{E}$ and feedback $\mathcal{F}$.
Different input-output configurations correspond to distinct methods and objectives. This unified formulation brings together diverse evaluation paradigms, offering a structured framework for categorizing and understanding various approaches within LLMs-as-judges.

\subsection{Evaluation Function $E$}
The evaluation function $E$ in the context of LLMs-as-judges can be categorized into three primary configurations: Single-LLM systems, Multi-LLM systems, and Hybrid systems that combine LLMs with human evaluators. Each of these configurations serves distinct purposes, offers different advantages, and faces unique challenges.

\begin{itemize}[leftmargin=*]
    \item \textbf{Single-LLM Evaluation System~\cite{lin2023llm,zhang2024talec,liu2023g}: } A single LLM evaluation system relies on a single model to perform the evaluation tasks.  It is simple to deploy and scale, making it efficient for tasks that don't require specialized evaluation. However, its flexibility is limited, as it may struggle with tasks that demand specialized knowledge or reasoning over complex inputs. Additionally, if not properly trained, a single model may introduce biases, leading to inaccurate evaluations.
    \item \textbf{Multi-LLM Evaluation Systems~\cite{chu2024pre,chan2023chateval,li2023prd}: } A Multi-LLM evaluation system combines multiple models that work together to perform evaluation tasks.  These models may interact through various mechanisms, such as collaboration, or competition, to refine their outputs and achieve more accurate results. By leveraging the strengths of different models, a multi-model system can cover a broader range of evaluation criteria and provide a more comprehensive assessment. However, this comes at a higher computational cost and requires more resources, making deployment and maintenance more challenging, particularly for large-scale tasks. 
    Moreover, while cooperation between models often enhances evaluation results, the methods through which these models achieve consensus or resolve differences remain key areas of ongoing exploration.
    \item \textbf{Human-AI Collaboration System~\cite{li2023collaborative,wang2023large,shankar2024validates}: } In this system, LLMs work alongside human evaluators, combining the efficiency of automated evaluation with the nuanced judgment of human expertise. This configuration allows human evaluators to mitigate potential biases in the LLM's output and provide subjective insights into complex evaluation tasks. While this system offers greater reliability and depth, it comes with challenges in coordinating between the models and humans, ensuring consistent evaluation standards, and integrating feedback. Additionally, the inclusion of human evaluators increases both the cost and time required for the evaluation process, making it less scalable than purely model-based systems.
\end{itemize}

\subsection{Evaluation Input}
In the LLMs-as-judges paradigm, in addition to the evaluation item $\mathcal{X}$, LLM judges typically receive three other types of inputs: Evaluation Type $\mathcal{T}$ , Evaluation Criteria $\mathcal{C}$, and Evaluation References $\mathcal{R}$. The following provides a detailed explanation:

\subsubsection{Evaluation Type $\mathcal{T}$}
The Evaluation Type $\mathcal{T}$ defines the specific evaluation mode, determining how the evaluation will be conducted. It typically includes three approaches: pointwise, pairwise, and listwise evaluation.

\begin{itemize}[leftmargin=*]
\item \textbf{Pointwise Evaluation~\cite{wang2023learning,kim2023prometheus,ye2024self}:} This method evaluates each candidate item individually based on the specified criteria.
For example, in a text summarization task, the LLM might evaluate each generated summary separately, assigning a score based on factors like informativeness, coherence, and conciseness. Although pointwise evaluation is simple and easy to apply, it may fail to capture the relative quality differences between candidates and can be influenced by biases arising from evaluating items in isolation.

\item \textbf{Pairwise Evaluation~\cite{saad2023ares,cao2024compassjudger,he2024fedeval,hu2024rethinking}:} This method involves directly comparing two candidate items to determine which one performs better according to the specified criteria. It is commonly used in preference-based tasks. For example, given two summaries of a news article, the LLM may be asked to decide which summary is more coherent or informative. Pairwise evaluation closely mirrors human decision-making processes by focusing on relative preferences rather than assigning absolute scores. This approach is especially effective when the differences between outputs are subtle and difficult to quantify.

\item \textbf{Listwise Evaluation~\cite{10.1145/3626772.3657813,yan-etal-2024-consolidating,hou2024large,niu2024judgerank}:} 
This method is designed to collectively evaluate the entire list of candidate items, evaluating and ranking them based on the specific criteria. It is often applied in ranking tasks, such as document retrieval in search engines, where the objective is to determine the relevance of the documents in relation to a user query. Listwise evaluation takes into account the interactions between multiple candidates, making it well-suited for applications that require holistic analysis. 
\end{itemize}

In general, these three evaluation modes are not entirely independent. pointwise scores can be aggregated to create pairwise comparisons or used to construct a ranked list. Similarly, pairwise preferences can be organized into a complete ranking list for listwise analysis. 
However, these transformations are not always reliable within the LLMs-as-judges framework~\cite{liu2024aligning}. For example, in pointwise evaluation, output $A$ may receive a score of 5, while output $B$ receives a score of 4, yet, a direct pairwise comparison might not consistently yield $A > B$ due to potential bias. Additionally, LLM judges do not always satisfy transitivity in their judgments. For instance, given pairwise preferences where $z_i > z_j$  and  $z_j > z_k$ , the LLM may not necessarily yield $z_i > z_k$. These inconsistencies contribute to concerns about the reliability and trustworthiness of the LLM-as-Judge framework, which we will discuss in detail in Section (\S\ref{sec:Limitation}).

\subsubsection{Evaluation Criteria $\mathcal{C}$.}

The evaluation criteria $\mathcal{C}$ define the specific standards that determine which aspects of the output should be assessed. These criteria are designed to cover a broad range of quality attributes and can be tailored based on the nature of the task. Typically, the criteria encompass the following aspects:

\begin{itemize}[leftmargin=*]
\item \textbf{Linguistic Quality~\cite{fabbri2021summeval,zhang2018personalizing,chhun2022human}:} This category evaluates the language-related features of the output, such as fluency, grammatical accuracy, coherence, and Conciseness. Linguistic quality is crucial in tasks like text generation, machine translation, and summarization, where clarity and readability are essential.

\item \textbf{Content Accuracy~\cite{chen2021evaluating,jimenez2023swe,tan2024judgebench}:} This dimension focuses on the correctness and relevance of the content. It includes evaluating aspects such as factual accuracy, ensuring that the output does not contain misleading or incorrect information. Content accuracy is particularly crucial in tasks such as code generation and fact-checking.

\item \textbf{Task-Specific Metrics~\cite{ji2024pku,lin2024wildbench,tan2024judgebench}:} In addition to general quality metrics, many tasks require evaluation based on standards specific to their respective domains. These standards may include metrics such as informativeness (assessing whether the output provides comprehensive and valuable information) or completeness (ensuring all key aspects of the input are covered). Other criteria may include diversity, well-structured content, and logical clarity.
\end{itemize}

In addition to providing clear evaluation criteria, offering several examples can also be beneficial for the assessment. By incorporating well-structured examples, LLMs can better align its output with user expectations, especially when handling complex tasks or ambiguous queries.

\subsubsection{Evaluation References $\mathcal{R}$.}
Evaluation References $\mathcal{R}$ are optional. Depending on the availability of evaluation reference, the evaluation process can be broadly divided into reference-based and reference-free scenarios.

\begin{itemize}[leftmargin=*]
\item \textbf{Reference-Based Evaluation~\cite{freitag2021results,karpinska2023large}: }
The reference-based evaluation leverages reference data to determine whether the performance meets the expected standards. It is commonly applied in tasks where the quality of the output can be objectively judged by its similarity to established reference. In Natural Language Generation (NLG) tasks, this method is widely used to evaluate the resemblance between generated content and reference content. For example, in machine translation or text summarization, an LLM can compare the generated translations or summaries against high-quality references. The key strength of this approach is its well-defined benchmarking process; however, its effectiveness may be constrained by the quality and variety of the reference data.

\item \textbf{Reference-Free Evaluation~\cite{shen2023opinsummeval,zheng2023judging,he2023socreval}: }
The reference-free evaluation does not rely on a specific reference $\mathcal{R}$, instead, it evaluates $\mathcal{X}$ based on intrinsic quality standards or its alignment with the source context. For example, when assessing language fluency or content coherence, an LLM can autonomously generate evaluation results using internal grammatical and semantic rules. This method is widely used in fields like sentiment analysis and dialogue generation. The main advantage of this approach is its independence from specific references, providing greater flexibility for open-ended tasks. However, its drawback lies in the difficulty of obtaining satisfactory evaluations in domains where the LLM lacks relevant knowledge.
\end{itemize}

\subsection{Evaluation Output}

In the LLMs-as-judges paradigm, the LLM typically generates three types of outputs: the evaluation result $\mathcal{Y}$, the explanation $\mathcal{E}$, and the feedback $\mathcal{F}$. Below are detailed descriptions.

\begin{itemize}[leftmargin=*]
    \item \textbf{Evaluation Result $\mathcal{Y}$~\cite{zhao2024auto,saad2023ares}: } The evaluation result $\mathcal{Y}$ is the primary output, which can take the form of a numerical score, a ranking, a categorical label, or a qualitative assessment. It reflects the quality, relevance, or performance of the candidate items according to the specified evaluation criteria. For example, in a machine translation task, $\mathcal{Y}$ could be a score indicating translation quality, while in a dialogue generation task, it might be a rating of coherence and appropriateness on a scale from 1 to 5. The evaluation result $\mathcal{Y}$ provides a clear measure of performance, enabling researchers to effectively compare different models or outputs.
    
    \item \textbf{Explanation $\mathcal{E}$~\cite{ye2024beyond,xie2024improving}: } The explanation $\mathcal{E}$ provides detailed reasoning and justifications for the evaluation result. It offers insights into why certain result received higher or lower scores, highlighting specific features of the candidate item that influenced the evaluation. For example, in a summarization task, the LLM judges might explain that the score was lowered due to missing critical information or the presence of redundant content. The explanation component enhances transparency, allowing users to understand the decision-making process of the LLM and gain deeper insights into the strengths and weaknesses of the evaluated content.

    \item \textbf{Feedback $\mathcal{F}$~\cite{madaan2024self,chen2023teaching}: } The feedback $\mathcal{F}$ consists of actionable suggestions or recommendations aimed at improving the evaluated output. Unlike the evaluation result, which merely indicates performance, the feedback component is designed to guide the refinement of the content. For instance, in a creative writing task, feedback might include recommendations for enhancing the narrative flow or improving clarity. This component is especially valuable for the iterative development of the evaluated item, as it provides concrete pointers that help both the LLM and content creators enhance the quality of the generated outputs.
\end{itemize}

Depending on the intended purpose and specific requirements of the evaluation, the LLM judges can generate various combinations of the three outputs ( $\mathcal{Y}$, $\mathcal{E}$, $\mathcal{F}$ ) for a given task. 
In most cases, providing explanation $\mathcal{E}$ not only helps users better understand and trust the evaluation results but also leads to more human-aligned and accurate evaluation result $\mathcal{Y}$. Moreover, generating feedback $\mathcal{F}$ generally demands a higher level of model capability, as it requires not only assessing the quality of the input but also providing concrete, actionable recommendations for improvement.

\section{Functionality}
\label{sec:Functionality}
As an emerging evaluation paradigm, LLMs-as-judges play a significant role across various scenarios. Based on their functionality, we categorize the application of LLM evaluators into three main directions: \textbf{Performance Evaluation} (\S\ref{sec:Performance Evaluation}), \textbf{Model Enhancement} (\S\ref{sec:Model Enhancement}), and \textbf{Data Construction} (\S\ref{sec:Data Construction}). In this section, we will delve into these functionalities, explore their potential, and discuss specific implementation methods.

\begin{figure}[t]
\includegraphics[width=\linewidth]{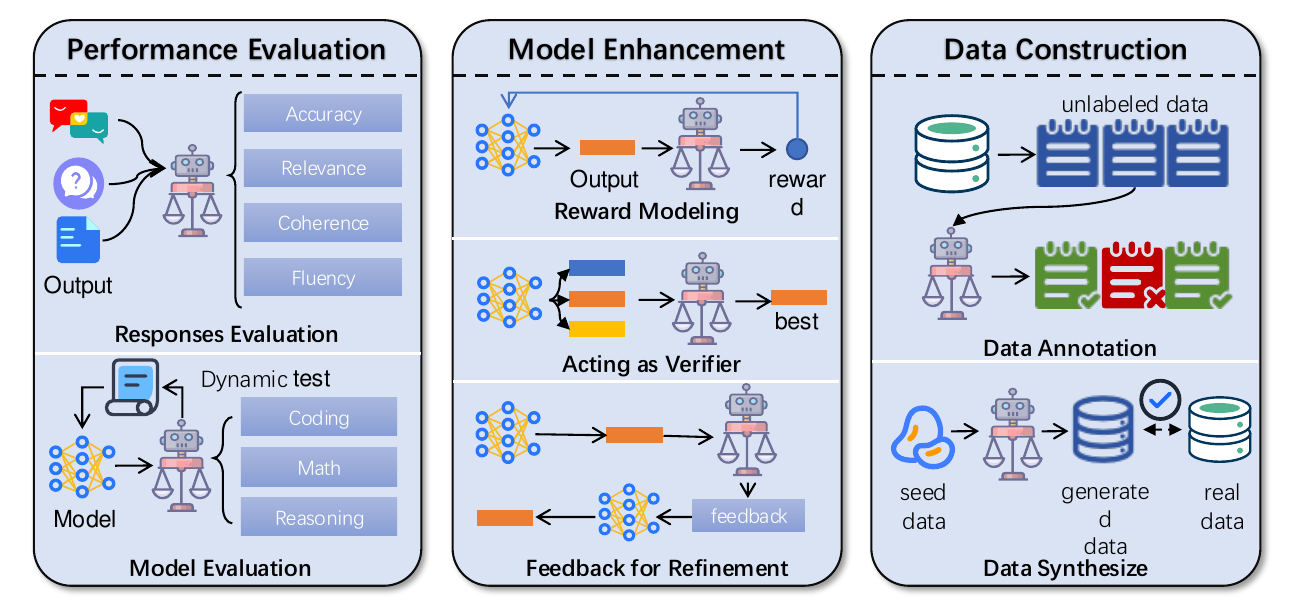}
\caption{Overview of the Functionality of LLMs-as-judges.}
% \vspace{-5mm}
\label{figure:function}
\end{figure}

\subsection{Performance Evaluation}
\label{sec:Performance Evaluation}
 Performance evaluation represents the most fundamental application objective of an LLM judges, serving as the cornerstone for understanding and optimizing their other function. It can be broadly divided into two components: \textbf{Responses Evaluation} (\S\ref{sec:Responses Evaluation}) and \textbf{Model Evaluation} (\S\ref{sec:Model Evaluation}). 
 Response Evaluation focuses on aspects such as the quality, relevance, and coherence, and fluency of the responses for a given task. In contrast, model evaluation takes a holistic approach, assessing the overall capabilities of LLMs. Although these two aspects are interconnected, they focus on different levels of analysis, providing multidimensional insights into performance. 

\subsubsection{Responses Evaluation}
\label{sec:Responses Evaluation}

The purpose of evaluating responses is to identify better answers within the context of a specific question or task, which can enhance overall decision-making.
These responses can originate from either AI models or humans. 
Evaluation criteria typically consider general attributes such as accuracy, relevance, coherence, and fluency. However, in practical applications, the evaluation of responses often requires customized metrics tailored to specific tasks. For instance, in the education domain, the focus may be more on the inspirational and educational value of the answers.

LLM judges have also been widely applied in the assessment of text response~\cite{lin2023llm,wang2024automated,zhou2024llm}. Lin et al.~\cite{lin2023llm} propose LLM-Eval, a unified framework employing a single-prompt strategy to evaluate the performance of open-domain dialogue systems across multiple dimensions, including content, grammar, relevance, and appropriateness. 
Wang et al.~\cite{wang2024automated} proposed an article scoring and feedback system tailored to different genres, such as essays, narratives, and question-answering articles. Using BERT and ChatGPT models, they enabled automated scoring and detailed feedback, showcasing the potential of LLMs in article evaluation.
Moreover, Zhou et al.~\cite{zhou2024llm} conduct a detailed evaluation of whether LLMs can serve as reliable tools for automated paper review. Their findings indicate that current LLMs are still not sufficiently reliable for such tasks, particularly in scenarios requiring logical reasoning or a deep knowledge base.

Furthermore, the evaluation of a single response is not limited to assessing the quality of the final answer but can also extend to analyzing the response process~\cite{saad2023ares,asai2023self,lei2024recexplainer}. For instance, this can include evaluating whether retrieval is necessary at a given step, the relevance of the retrieved documents, and the interpretability of the response.
For example, ARES~\cite{saad2023ares} uses LLM judges to evaluate RAG systems across three dimensions: Contextual Relevance, Answer Faithfulness, and Answer Relevance. Similarly, Asai et al. ~\cite{asai2023self} proposed SELF-RAG, which employs reflective token to determine whether retrieval is required and to self-assess the quality of generated outputs. Lei et al.~\cite{lei2024recexplainer} introduced LLMs to evaluate the quality of generated explanations, demonstrating the effectiveness of LLMs in understanding and generating explanations for recommendation tasks.

\subsubsection{Model Evaluation}
\label{sec:Model Evaluation}
Model evaluation typically begins with assessing individual responses and then extends to analyzing overall capabilities. This wider perspective aims to analyze the model's performance across various tasks or domains, such as coding ability, instruction-following proficiency, reasoning, and other specialized skills relevant to its intended applications.

A common and straightforward approach is to represent model performance using average performance on static benchmarks~\cite{zheng2023judging,lin2024wildbench,tan2024judgebench}. LLM judges assess the model's performance using a set of carefully designed metrics, which results in a performance ranking. This method is widely adopted due to its simplicity and comparability. For example, task sets can be designed to evaluate the model's knowledge coverage, reasoning depth, and language generation quality~\cite{son2024mm,liu2024rm,lambert2024rewardbench}, or real-world scenarios can be simulated to assess the model's ability to handle complex situations~\cite{liu2023agentbench,trivedi2024appworld}.

As the demand for evaluation increases, the evaluation process has gradually shifted from traditional static testing to more dynamic, interactive assessments~\cite{bai2024benchmarking,zhao2024auto,yu2024kieval}.
LLMs-as-judges has pioneered this approach, similar to Chatbot Arena~\cite{zheng2023judging}, a crowdsourced platform that collects anonymous votes on LLM performance and ranks them using Elo scores. 
Auto-Arena~\cite{zhao2024auto,luo2024videoautoarena} and LMExam~\cite{bai2024benchmarking} assess model capabilities by using LLMs as both question setters and evaluators. These frameworks innovatively combine diverse question generation, multi-turn question-answering evaluation, and a decentralized model-to-model evaluation mechanism, providing more detailed and granular performance assessments.
Additionally, KIEval~\cite{yu2024kieval} introduces an LLM-driven ``interactor'' role, which evaluates the knowledge mastery and generation abilities of LLMs through dynamic multi-turn conversations. These dynamic evaluation methods effectively address data leakage and evaluation bias issues common in traditional benchmark tests.

\subsection{Model Enhancement}
\label{sec:Model Enhancement}
In addition to Performance Evaluation, LLMs-as-judges is also widely used for Model Enhancement. From training to inference, LLMs-as-judges plays a key role in improving model performance. Its application in model enhancement offers a novel optimization pathway for artificial intelligence, fostering the refinement and personalization of intelligent systems across a broader spectrum of real-world applications.

\subsubsection{Reward Modeling During Training}
\label{sec: Reward Modeling During Training}

A primary application of LLMs-as-judges is in reward modeling during training, particularly in reinforcement learning with feedback~\cite{yuan2024self,guo2024direct,wang2024cream,xu2024perfect,cao2024enhancing}. LLM judges assign scores to model outputs by evaluating them against human-defined criteria, guiding optimization toward desired behaviors. This ensures alignment with human values, improving the quality and relevance of the generated outputs and improving the effectiveness of LLMs in real-world tasks.

A series of works, such as SRLMs~\cite{yuan2024self}, OAIF~\cite{guo2024direct}, and RLAIF~\cite{lee2023rlaif}, have enabled LLMs to become their own reward models. This overcomes the traditional RLHF dependency on fixed reward models, allowing the model to iteratively reward and self-optimize, fostering self-evolution through continuous self-assessment.
RELC~\cite{cao2024enhancing} tackles the challenge of sparse rewards in traditional RL by introducing a Critic Language Model (Critic LM) to evaluate intermediate generation steps. This dense feedback at each step helps mitigate reward sparsity, offering more detailed guidance to the model during training.

However, using the same LLM for both policy generation and reward modeling can pose challenges in ensuring the accuracy of the rewards. This dual role setup may lead to accumulated biases and preference data noise, which can undermine the training effectiveness. To address this issue, CREAM~\cite{wang2024cream} introduces cross-iteration consistency constraints to regulate the training process and prevent the model from learning unreliable preference data. This significantly enhances reward consistency and alignment performance. In addition, CGPO~\cite{xu2024perfect} groups tasks by category (such as dialogue, mathematical reasoning, safety, etc.) and uses ``Mixed Judges'' to assign a specific reward model to each task group. This ensures that the reward signals are closely aligned with the task objectives, thereby preventing conflicts between different goals.

\subsubsection{Acting as Verifier During Inference}
\label{sec: Acting as Verifier During Inference}
During inference, LLM judges serve as verifier, responsible for selecting the optimal response from multiple candidates~\cite{yao2024tree,besta2024graph,lightman2023let,musolesi2024creative,besta2024graph}. By comparing the outputs based on various metrics, such as factual accuracy and reasoning consistency, they are able to identify the best fit for the given task or context, thereby optimizing the inference process or improving the quality of the generated results.

One of the simplest applications is Best-of-N sampling~\cite{jinnai2024regularized,sun2024fast}, where the model is sampled N times, and the best result is selected to improve model performance. Similarly, Wang et al.~\cite{wang2022self} introduced a promising sampling method called self-consistency, where n samples are drawn from the judge model, and the average score is output. These sampling methods enhance inference stability by selecting the best result from multiple evaluations.
Further optimization strategies include the Tree of Thoughts (ToT)~\cite{yao2024tree} method, which models the problem-solving process as a tree structure. This allows the model to explore multiple solution paths and optimize path selection through self-assessment mechanisms.
The Graph of Thoughts (GoT)~\cite{besta2024graph} method extends this concept by introducing directed graphs, where the non-linear interactions between nodes improve the efficiency and precision of multi-step reasoning. In both methods, LLM judges play a crucial role in guiding the model to select the most promising paths, thereby enhancing the quality and accuracy of reasoning.

Similarly, Lightman et al.~\cite{lightman2023let} discuss how step-by-step validation can enhance the performance of LLMs in multi-step reasoning tasks, particularly in the domain of mathematics. SE-GBS~\cite{xie2024self} integrates self-assessment into the multi-step reasoning decoding process, generating scores that reflect logical correctness and further ensuring the accuracy and consistency of the reasoning chain. The REPS~\cite{kawabata2024rationale} improves the accuracy and reliability of reasoning validation models by comparing reasoning paths pairwise, verifying their logical consistency and factual basis.
Also, Musolesi et al. ~\cite{musolesi2024creative} proposed Creative Beam Search, with the LLM acting as a judge to simulate the human creative selection process, thereby enhancing the diversity and creativity of the generated results.

\subsubsection{Feedback for Refinement}
\label{sec: Feedback for Refinement}
After receiving the initial response, LLM judges provide actionable feedback to iteratively improve output quality. By analyzing the response based on specific task criteria, such as accuracy, coherence, or creativity, the LLM can identify weaknesses in the output and offer suggestions for improvement. This iterative refinement process plays a crucial role in applications that require adaptability~\cite{madaan2024self,paul2023refiner,chen2023teaching,xu2023towards,huang2023large}.

SELF-REFINE~\cite{madaan2024self} enables LLMs to iteratively improve output quality through feedback generated by the model itself, without requiring additional training or supervision data. On the other hand, SELF-DEBUGGING~\cite{chen2023teaching} demonstrates a practical application of self-correction in code generation by identifying and rectifying errors through self-explanation and feedback. This approach has significantly enhanced the performance of LLMs across various code generation tasks.

In addition to refining response quality, LLMs judges are also widely used to enhance reasoning abilities. For example, REFINER~\cite{paul2023refiner} optimizes the reasoning performance of LLMs through interactions between a generator model and a critic model. In this framework, the generator model is responsible for producing intermediate reasoning steps, while the critic model analyzes these steps and provides detailed feedback, such as identifying calculation errors or logical inconsistencies. Xu et al.~\cite{xu2023towards} propose a multi-agent collaboration strategy to enhance the reasoning abilities of LLMs by simulating the academic peer review process. The framework is divided into three stages: generation, review, and revision. Agents provide feedback and attach confidence scores to refine the initial answers, with the final result determined through majority voting.

While the feedback and correction mechanisms of LLMs judges are continually evolving, the limitations of self-feedback in improving quality should not be overlooked. Research on Self-Correct~\cite{huang2023large,tyen2023llms} shows that, the intrinsic self-correction capabilities of LLMs often fall short of effectively improving reasoning quality. 
Valmeekam et al.~\cite{valmeekam2023can} also raise concerns about the effectiveness of LLMs as self-validation tools in the absence of reliable external validators. Future research can focus on improving the accuracy of feedback provided by these LLM judges and incorporating external validation mechanisms to optimize their performance in complex reasoning tasks.

\subsection{Data Construction}
\label{sec:Data Construction}
Data collection is a crucial stage in the development of machine learning systems, especially those driven by the rapid advancements in deep learning. The quality of the data directly determines the performance of the trained models.
The LLMs-as-judges has significantly transformed the landscape of data collection, substantially reducing reliance on human effort.
In this section, we will explore the pivotal role of LLMs-as-judges in data collection from two key perspectives: \textbf{Data Annotation} (\S\ref{sec:Data Annotation}) and \textbf{Data Synthesize} (\S\ref{sec:Data Synthesize}).

\subsubsection{Data Annotation}
\label{sec:Data Annotation}
Data Annotation involves leveraging LLM judges to label large, unlabeled datasets efficiently~\cite{he2024if,gilardi2023chatgpt,tornberg2023chatgpt,hao2024fullanno}. By utilizing the advanced natural language understanding and reasoning capabilities of LLMs, the annotation process can be automated to a significant extent, enabling the generation of high-quality labels with reduced human intervention.

LLMs have demonstrated remarkable potential in text annotation tasks, consistently outperforming traditional methods and human annotators in various settings.
He et al.~\cite{he2024if} evaluated the performance of GPT-4 in crowdsourced data annotation workflows, particularly in text annotation tasks. Their comparative study revealed that, even with best practices, the highest accuracy achievable by MTurk workers was 81.5\%, whereas GPT-4 achieved an accuracy of 83.6\%.
Similarly, Gilardi et al.~\cite{gilardi2023chatgpt} analyzed 6,183 tweets and news articles, demonstrating that ChatGPT outperformed crowdsourced workers in tasks such as stance detection, topic detection, and framing.
Törnberg et al.~\cite{tornberg2023chatgpt} further investigated the classification of Twitter users' political leanings based on their tweet content. Their findings revealed that ChatGPT-4 not only surpassed human classifiers in accuracy and reliability but also exhibited bias levels that were comparable to or lower than those of human classifiers.

As technology advances, more and more research is exploring their application in multimodal data annotation. For example, the FullAnno~\cite{hao2024fullanno} uses the GPT-4V model to generate image annotations, significantly improving the quality of image descriptions through a multi-stage annotation process.
Furthermore, Latif et al.~\cite{latif2023can} explored the application of LLMs in speech emotion annotation, demonstrating that, with data augmentation, LLM-annotated samples can significantly enhance the performance of speech emotion recognition models. By integrating text, audio features, and gender information, the effectiveness of LLM-based annotations was further improved, highlighting their potential in advancing multimodal annotation tasks.

As LLMs perform excellently in annotation tasks, researchers are actively exploring methods to further improve annotation quality and address potential challenges. For example, AnnoLLM~\cite{he2023annollm} introducedthe ``explain-then-annotate'' method, which enhances both the accuracy and transparency of annotations by prompting the LLM to justify its label assignments. 
Additionally, the LLMAAA~\cite{zhang2023llmaaa} framework incorporates an active learning strategy to efficiently select high-information samples for annotation, thereby mitigating the effects of noisy labels and reducing the reliance on costly human annotation. These approach not only enhance the performance of task-specific models but also offer new perspectives on the efficient application of LLMs in annotation workflows.

\subsubsection{Data Synthesize}
\label{sec:Data Synthesize}
The goal of Data Synthesis is to create entirely new data, either from scratch or based on seed data, while ensuring it is similar in distribution to real data. Data Synthesis enables the generation of diverse data samples, enhancing a model's generalization ability to unseen examples while reducing reliance on sensitive real-world data~\cite{ye2023selfee,dong2024self,arif2024fellowship,wang2022self,kim2024aligning,mendoncca2024soda}.

In recent years, advancements in LLMs have led to significant improvements in both the quality and efficiency of data synthesis methods. In this domain, methods like SELFEE~\cite{ye2023selfee} and SynPO~\cite{dong2024self} have effectively enhanced the alignment capabilities of LLMs by leveraging small amounts of labeled data and iteratively generating preference-aligned data. Arif et al.~\cite{arif2024fellowship} also introduce a multi-agent workflow for generating optimized preference datasets.

SELF-INSTRUCT~\cite{wang2022self} and Evol-Instruct~\cite{xu2023wizardlm,zeng2024automatic} represent innovative approaches to improving model alignment and performance through self-generated instruction data. SELF-INSTRUCT~\cite{wang2022self}  requires minimal human annotation, instead relying on self-generated instruction data to align pre-trained models. Evol-Instruct~\cite{xu2023wizardlm,zeng2024automatic}  further enhances LLM performance by automatically generating instruction data, significantly boosting model capabilities.

STaR~\cite{zelikman2024star} and ReSTEM~\cite{singh2023beyond} are research efforts aimed at enhancing reasoning capabilities through synthetic data. STaR~\cite{zelikman2024star} employs a self-guided iterative process to improve model performance on complex reasoning tasks, offering an effective solution for tackling increasingly sophisticated reasoning challenges in the future. ReSTEM~\cite{singh2023beyond}, on the other hand, utilizes a self-training approach based on the expectation-maximization framework to enhance the problem-solving capabilities of large language models, particularly in areas such as solving mathematical problems and generating code.

\definecolor{mygray}{gray}{.9}

\section{Methodology}
\label{sec:Methodology}
\par
\begin{figure}[t]
\includegraphics[width=\linewidth]{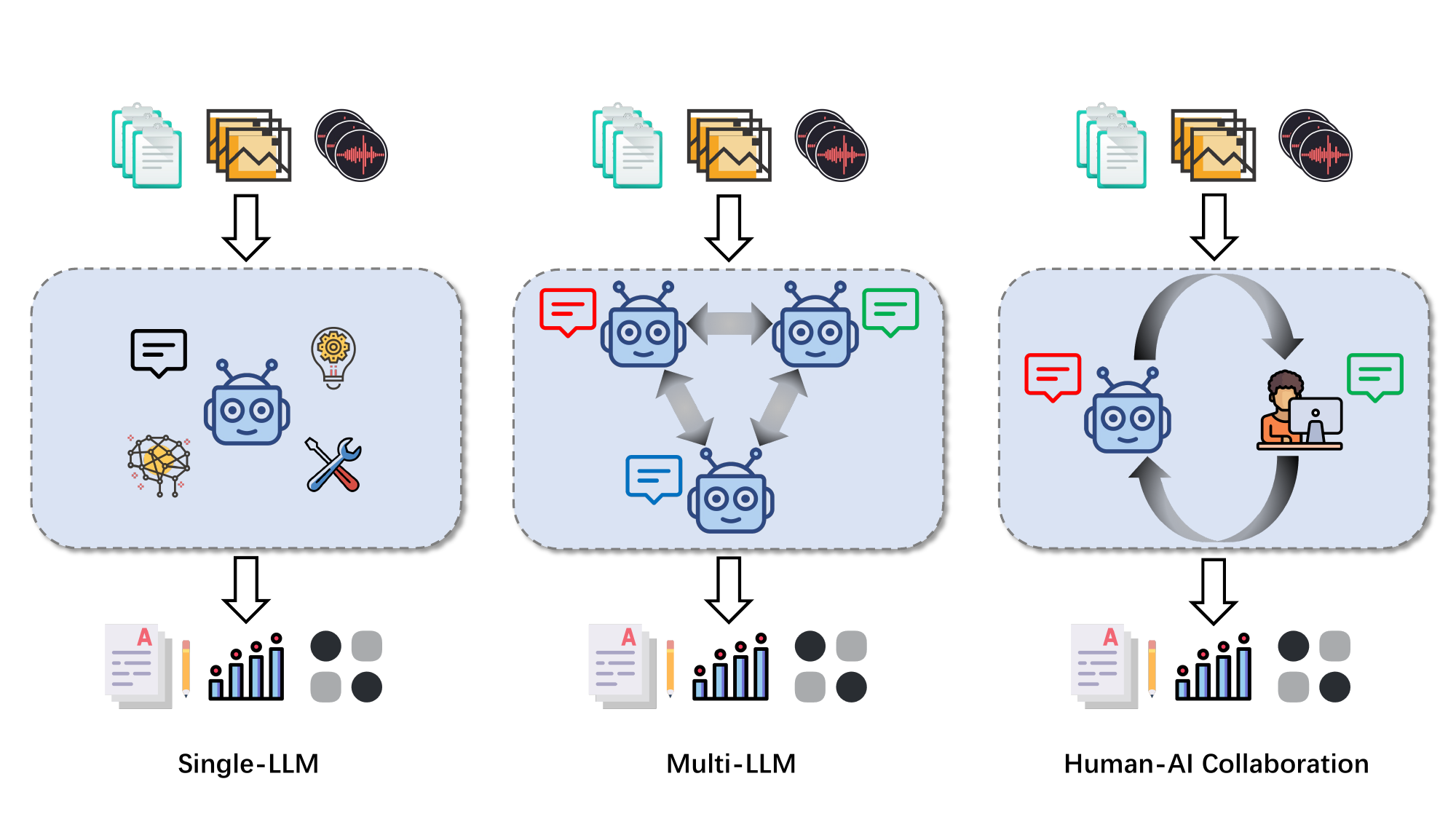}
\caption{Overview of the Methodology of LLMs-as-judges.}
% \vspace{-5mm}
\label{figure:method}
\end{figure}

 The use of LLM judges requires careful methodological considerations to ensure the accuracy and consistency of judgments. 
Researchers have developed various approaches according to the complexity and specific requirements of different judgment tasks, each offering unique advantages. 
In this section, we categorize these methodologies into three broad approaches: \textbf{Single-LLM System} (\S\ref{sec:Single-LLM System}): evaluation by a single-LLM, \textbf{Multi-LLM System} (\S\ref{sec:Multi-LLM System}): evaluation by cooperation among multi-LLMs, and \textbf{Human-AI Collaboration} (\S\ref{sec:Hybrid System}): evaluation by cooperation of LLMs and Human.
Figure ~\ref{figure:method} presents an overview of methodology.

\subsection{Single-LLM System}
\label{sec:Single-LLM System}

\par Single-LLM System relies on a single model to perform judgment tasks, with its effectiveness largely determined by the LLM's capabilities and the strategies used to process input data. This approach can generally be divided into three fundamental components: \textbf{Prompt Engineering} (\S\ref{sec:Prompt Engineering}), \textbf{Tuning} (\S\ref{sec:Tuning}), and \textbf{Post-processing} (\S\ref{sec:Post-processing}) of model outputs.

\subsubsection{Prompt-based}
\label{sec:Prompt Engineering}
\par Prompt engineering~\cite{sahoo2024systematic} involves crafting clear and structured input prompts tailored to elicit accurate and contextually appropriate responses from LLM judges. This approach is crucial for ensuring that LLMs grasp the complexities of specific tasks and provide relevant, consistent, and goal-aligned judgments. 
In many cases, well-designed prompts significantly reduce the need for extensive model training.

\textbf{In-Context Learning.} In-Context Learning (ICL) is a distinctive capability of LLMs that allows them to dynamically adapt to evaluation tasks using carefully curated examples or explanations within the prompt~\cite{dong2022survey}. Several recent methods have demonstrated the power of ICL in LLM-as-judges, showcasing how it enhances the flexibility and performance of LLMs in diverse settings. For example, GPTScore~\cite{fu2023gptscore} leverages the few-shot learning capability of generative pre-trained models to evaluate generated text. By using relevant examples to customize prompts, it provides a flexible, training-free approach to assess multiple aspects of text quality.
Similarly, LLM-EVAL~\cite{lin2023llm} incorporates carefully crafted examples into prompts, proposing a unified, multi-dimensional automatic evaluation method for open-domain dialogue.
Another notable example is TALEC~\cite{zhang2024talec}, a model-based evaluation method that leverages in-context learning to enable users to set custom evaluation criteria for LLMs in specific domains. Through careful prompt engineering, users can iteratively adjust the examples to refine the evaluation process as needed.
In addition, Jain et al.~\cite{jain2023multi} proposed the In-Context Learning-based Evaluator (ICE) for multi-dimensional text evaluation. ICE leverages LLMs and a small number of in-context examples to evaluate generated text summaries, achieving competitive results.

While ICL can enable effective evaluation, it is not without challenges. One major issue is that the model's responses may be influenced by the selection of prompt examples, potentially leading to bias~\cite{zhao2021calibrate,zhou2023batch,han2022prototypical,fei2023mitigating}.
To address this issue, Hasanbeig et al. proposed ALLURE~\cite{hasanbeig2023allure}, a comprehensive protocol designed to mitigate bias in ICL for LLMs during text evaluation. ALLURE~\cite{hasanbeig2023allure} improves evaluator accuracy by iteratively incorporating discrepancies between its assessments and annotated data into the learning context. Moreover, after uncovering the existence of symbol bias within LLM evaluators when using ICL, Song et al.~\cite{song2024can} proposed two effective mitigation strategy prompt templates, Many-Shot with Reference (MSwR) and Many-Shot without Reference (MSoR), to bolster the reliability and precision of LLM-based assessments.

\textbf{Step-by-step.}
Step-by-step involves breaking down complex evaluation tasks into fine-grained components, leveraging the reasoning capabilities of LLMs to simplify the evaluation process. The most straightforward example of which is perhaps Chain-of-Thought (CoT)~\cite{wei2022chain,kotonya2023little}. Building on that, frameworks like G-EVAL~\cite{liu2023g} have been proposed to assess the quality of NLG outputs. G-EVAL~\cite{liu2023g} combines CoT with a form-filling paradigm, allowing the LLM to assess outputs in a structured manner.
Similarly, ICE-Score~\cite{zhuo2023ice} introduces a step-by-step framework for evaluating code, in which the LLM is instructed with task definitions, evaluation criteria, and detailed evaluation steps. By breaking the task down into clear steps, ICE-Score~\cite{zhuo2023ice} improves the quality and consistency of code evaluation. Also, ProtocoLLM~\cite{yi2024protocollm} employs a similar step-by-step approach to evaluate the specialized capabilities of LLMs in generating scientific protocols. 
Portia~\cite{li2023split} achieves better evaluation results in a lightweight yet effective manner. It divides the answer into multiple parts, aligns similar content between candidate answers, and then merges them back into a single prompt for evaluation by the LLM.

Some studies break down evaluations into two steps: ``explanation-rating.'' This approach suggests that providing an explanation enhances the reliability of the rating. Chiang et al.~\cite{chiang2023closer} offer empirical guidelines to improve the quality of LLM evaluations, demonstrating that combining rating with explanation (rate-explain) or explanation with rating (explain-rate) leads to higher correlations with human ratings.
Another effective strategy is to decompose complex evaluation standards into specific, discrete criteria, allowing the LLM to assess each aspect independently. FineSurE~\cite{song2024finesure} is an advanced example of this method, offering a framework for the fine-grained evaluation of text summarization quality. It breaks down the evaluation into multiple dimensions, such as faithfulness, completeness, and conciseness. Through detailed analysis, including fact-checking and key fact alignment, FineSurE~\cite{song2024finesure} outperforms traditional methods in terms of evaluation accuracy.

\textbf{Definition Augmentation.}
The Enhanced Definition approach involves refining prompts to inject improved evaluation criteria, establish assessment principles, or incorporate external knowledge into the LLM judge's decision-making process.
Some studies focus on enriching and clarifying the prompts to ensure that the evaluation criteria are both comprehensive and well-defined.

For example, Liu et al. propose AUTOCALIBRATE~\cite{liu2023calibrating}, a multi-stage, gradient-free approach. This method involves the drafting, revision, and application of calibrated criteria, and it automatically calibrates and aligns an LLM-based evaluator to match human preferences for NLG quality assessment.
Furthermore, SALC~\cite{gupta2024unveiling} enables LLMs to autonomously generate context-aware evaluation criteria for self-assessment, overcoming the limitations of static, human-defined metrics. 
On the other hand, the LLM-as-a-Personalized-Judge approach~\cite{dong2024can} introduces a novel perspective by incorporating diverse evaluative roles and principles. This allows LLMs to adapt to complex, varied evaluation scenarios, resulting in more nuanced and context-sensitive assessments.

Another key aspect of Definition Augmentation is the retrieval of external knowledge, which helps reduce hallucinations and provides more factual support. For instance, BiasAlert~\cite{fan2024biasalert}, a tool designed to detect social bias in LLM-generated open-text outputs. It integrates external human knowledge with the LLM judge's inherent reasoning capabilities to reliably identify and mitigate bias, outperforming GPT4-as-A-Judge across various scenarios.
Moreover, Chen et al.~\cite{chen2024llms} found that within retrieval-augmented generation (RAG) frameworks, LLM judges do not exhibit a significant self-preference effect during evaluation.

\textbf{Multi-turn Optimization.}
Multi-turn optimization involves iterative interactions between the evaluator and the evaluated entity, refining evaluation results through diverse forms of feedback, thus fostering deeper analysis and a progressive improvement in evaluation quality~\cite{zhou2024fairer}.
Unlike traditional methods that rely on predefined criteria, Xu et al. proposed ACTIVE-CRITIC~\cite{xu2024large}, enabling LLMs to infer evaluation criteria from data and dynamically optimize prompts through multiple rounds of interaction. Moreover, 
Some studies~\cite{zhao2024auto,luo2024videoautoarena,bai2024benchmarking,yu2024kieval} leverage LLMs as question designers to engage in dynamic interactions with the evaluated entities, adjusting the questions and task design in real time. This allows for flexible modification of the evaluation content based on the performance of the evaluated entity, thereby enabling more comprehensive assessments.

\subsubsection{Tuning-based}
\label{sec:Tuning}
\par Tuning involves training a pre-existing LLM on a specialized dataset to adapt it to specific judgment tasks. It's especially useful when the judgment domain involves highly specialized knowledge or nuanced decision-making~\cite{huang2024limitations}.

\textbf{Score-based Tuning.} Score-based tuning involves using data with scores to train models and enhance their ability to predict judgment scores based on specific evaluation criteria~\cite{chen2023adaptation,deshwal2024phudge,wang2023learning}.
% Supervised Fine-Tuning (SFT) methods leverage labeled data to directly fine-tune language models, enhancing their performance on specific tasks by aligning them with human values and improving factual accuracy.

Many studies have explored the enhancement of LLM-as-judges by fine-tuning them on human-labeled datasets. 
For instance, PHUDGE~\cite{deshwal2024phudge}, fine-tuned from the Phi-3 model, achieves state-of-the-art performance in terms of latency and throughput when automatically evaluating the quality of outputs from LLMs. This fine-tuning process equips the model with the necessary judgment skills, enabling it to assess various types of content in a structured and accurate manner.
Additionally, ECT~\cite{wang2023learning} introduces a novel method for transferring scoring capabilities from LLMs to lighter models. This allows the lighter models to function as effective reward models for sequence generation tasks, enhancing sequence generation models through reinforcement learning and reranking approaches.
AttrScore~\cite{yue2023automatic} is another framework for evaluating attribution and identifying specific types of attribution errors, using a curated test set from a generative search engine and simulated examples from existing benchmarks.
The above research highlights that LLMs can better align their decision-making process with humans through fine-tuning with human-constructed datasets.
% By leveraging annotated data, LLMs can better align their decision-making process with human reasoning, thereby improving their ability to evaluate tasks with greater accuracy and relevance.

In addition to human-labeled data, some studies have also attempted to fine-tune models using synthetic datasets like SorryBench~\cite{xie2024sorry} generated for evaluation tasks. These datasets are often created through rule-based methods or by generating artificial evaluation examples, which also give rise to some metrics like TIGERScore~\cite{jiang2023tigerscore}.
SELF-J~\cite{ye2024self} is a self-training framework for developing judge models to evaluate LLMs' adherence to human instructions without human-annotated quality scores. SELF-J~\cite{ye2024self} proposes selective instruction following, allowing systems to decline low-quality instructions.
FENCE~\cite{xie2024improving} is another factuality evaluator designed to provide claim-level feedback to language model generators. It details a data augmentation approach that enriches public datasets with textual critiques and diverse source documents from various tools, thereby enhancing factuality without introducing lesser-known facts.
Utilizing synthetic training data to fine-tune lightweight language model judges and employing prediction-powered inference (PPI) for statistical confidence to mitigate potential prediction errors, ARES~\cite{saad2023ares} can automatically assess RAG systems.

\textbf{Preference-based Learning.} 
Preference-based learning focuses on training LLMs to make inferences and learn based on preferences, enabling the development of more adaptive and customizable evaluation capabilities.

Initially, researchers leverage these data in conjunction with advanced techniques like Direct Preference Optimization (DPO)~\cite{rafailov2024direct} to train LLMs for more nuanced evaluative capabilities. In this method, the model is trained to predict which of two outputs is preferred according to human-like values, rather than learning a scalar reward signal. Such self-improving approach is well reflected in Meta-Rewarding~\cite{wu2024meta}.
Con-J~\cite{ye2024beyond} trains a generative judge by using the DPO loss on contrastive judgments and the SFT loss on positive judgments to align LLMs with human values.
In terms of evaluating other LLMs effectively in open-ended scenarios, JudgeLM~\cite{zhu2023judgelm} addresses key biases in the fine-tuning process with a high-quality preference dataset.
Another typical method is PandaLM~\cite{wang2023pandalm}, which is trained on a reliable human-annotated preference dataset, focusing extends beyond just the objective correctness of responses, and addresses vital subjective factors.
Moreover, Self-Taught~\cite{wang2024self} is another approach to train LLMs as effective evaluators without relying on human-annotated preference judgments, using synthetic training data only. Through an iterative self-improvement scheme, LLM judges are able to produce reasoning traces and final judgments.
Not quite the same, FedEval-LLM~\cite{he2024fedeval} fine-tunes many personalized LLMs without relying on labeled datasets to provide domain-specific evaluation, mitigating biases associated with single referees. It is designed to assess the performance of LLMs on downstream tasks, at the same time, ensuring privacy preservation.

As research has progressed, newer methods have emerged that combine both score-based and preference-based data to refine model evaluation capabilities, not to mention some novel metrics like INSTRUCTSCORE~\cite{xu2023instructscore}.
FLAMe~\cite{vu2024foundational} is an example of such an approach. It's a family of Foundational Large Autorater Models which significantly improves generalization to a wide variety of held-out tasks using both pointwise and pairwise methods during training.
As generative judge model, AUTO-J~\cite{li2023generative} addresses challenges in generality, flexibility, and interpretability by training on a diverse dataset containing scoring and preference.
To critique and refine the outputs of large language models, Shepherd~\cite{wang2023shepherd} leverages a high-quality feedback dataset to identify errors and suggest improvements across various domains.
In the domain of NLG, X-EVAL~\cite{liu2023x} consists of a vanilla instruction tuning stage and an enhanced instruction tuning stage that exploits connections between fine-grained evaluation aspects.
Notably, Themis~\cite{hu2024themis} also achieved outstanding results acting as a reference-free NLG evaluation language model designed for flexibility and interpretability.
Similarly, CritiqueLLM~\cite{ke2024critiquellm} provides effective and explainable evaluations of LLM outputs, and uses a dialogue-based prompting method to generate high-quality referenced and reference-free evaluation data.
Self-Rationalization~\cite{trivedi2024self} enhances LLM performance by iteratively fine-tuning the judge via DPO, which allows LLMs to learn from their own reasoning.
Based on pointwise and pairwise dataset, CompassJudger-1~\cite{cao2024compassjudger} acts as an open-source, versatile LLM for efficient and accurate evaluation of other LLMs.
Likewise, Zhou et al.~\cite{zhou2024mitigating} introduces a systematic framework for bias reduction, employing calibration for closed-source models and contrastive training for open-source models.
Apart from that, HALU-J~\cite{wang2024halu} is designed to enhance hallucination detection in LLMs by selecting pertinent evidence and providing detailed critiques.
PROMETHEUS~\cite{kim2023prometheus} and PROMETHEUS 2~\cite{kim2024prometheus} are open-source LLMs specialized for fine-grained evaluation that can generalize to diverse, real-world scoring rubrics beyond a single-dimensional preference, supporting both direct assessment and pairwise ranking, and can evaluate based on custom criteria. What's more, the following PROMETHEUS-VISION~\cite{lee2024prometheusvision} fills the gap in the visual field.
As for various multimodal tasks, LLaVA-Critic~\cite{xiong2024llava} demonstrates its effectiveness in providing reliable evaluation scores and generating reward signals for preference learning, highlighting the potential of open-source LMMs in self-critique and evaluation.

\begin{table*}[t]
\centering
\caption{An Overview of Fine-Tuning Methods in Single-LLM Evaluation (Sorted in ascending alphabetical order).}
\scalebox{0.56}{
\begin{tabular}{cccccccc}
\hline
\multirow{2}{*}{Method} & \multicolumn{3}{c}{Data Construction} & \multicolumn{2}{c}{Tuning Method} & \multirow{2}{*}{Base LLM} \\ \cmidrule(lr){2-4} \cmidrule(lr){5-6}
&Annotator&Domain&Scale&Evaluation Type&Technique&\\ \hline

ARES~\cite{saad2023ares}&Human \& LLM&RAG System&-&Pairwise&PPI&DeBERTa-v3-Large\\

\rowcolor{mygray}
AttrScore~\cite{yue2023automatic}&Human&Various&63.8K&Pointwise&SFT&Multiple LLMs\\

AUTO-J~\cite{li2023generative}&Human \& GPT-4&Various&4396&Pointwise \& Pairwise&SFT&Llama2-13B-Chat\\

\rowcolor{mygray}
&&&&Pointwise, Pairwise,&&\\
\rowcolor{mygray}
\multirow{-2}{*}{CompassJudger-1~\cite{cao2024compassjudger}}&\multirow{-2}{*}{Human \& LLM}&\multirow{-2}{*}{Various}&\multirow{-2}{*}{900K}& \& Generative&\multirow{-2}{*}{SFT}&\multirow{-2}{*}{Qwen2.5 Series}\\

Con-J~\cite{ye2024beyond}&Human \& ChatGPT&Creation, Math, \& Code&220K&Pairwise&SFT \& DPO&Qwen2-7B-Instruct\\

\rowcolor{mygray}
CritiqueLLM~\cite{ke2024critiquellm}&Human \& GPT-4&Various&7722&Pointwise \& Pairwise&SFT&ChatGLM3-6B\\

&&Machine Translation,&&&&\\
&&Text Style Transfer,&&&&\\
\multirow{-3}{*}{ECT~\cite{wang2023learning}}&\multirow{-3}{*}{ChatGPT}&\& Summarization&\multirow{-3}{*}{-}&\multirow{-3}{*}{Pointwise}&\multirow{-3}{*}{SFT \& RLHF}&\multirow{-3}{*}{RoBERTa}\\

\rowcolor{mygray}
&&Instruct-tuning&5K, 10K,&&&\\
\rowcolor{mygray}
\multirow{-2}{*}{FedEval-LLM~\cite{he2024fedeval}}&\multirow{-2}{*}{Human}&\& Summary&per client&\multirow{-2}{*}{Pairwise}&\multirow{-2}{*}{LoRA}&\multirow{-2}{*}{Llama-7B}\\

&&Summarization,&&&&\\
\multirow{-2}{*}{FENCE~\cite{xie2024improving}}&\multirow{-2}{*}{Human \& LLM}&QA, \& Dialogue&\multirow{-2}{*}{-}&\multirow{-2}{*}{Pointwise}&\multirow{-2}{*}{SFT \& DPO}&\multirow{-2}{*}{Llama3-8B-Chat}\\

\rowcolor{mygray}
&&&&Pointwise, Pairwise,&&\\
\rowcolor{mygray}
&&&&Classification,&&\\
\rowcolor{mygray}
\multirow{-3}{*}{FLAMe~\cite{vu2024foundational}}&\multirow{-3}{*}{Human}&\multirow{-3}{*}{Various}&\multirow{-3}{*}{5.3M}&\& Open-ended generation&\multirow{-3}{*}{RLHF}&\multirow{-3}{*}{PaLM-2-24B}\\

&GPT-4-Turbo&Multiple-Evidence&&&&\\
\multirow{-2}{*}{HALU-J~\cite{wang2024halu}}&\& GPT-3.5-Turbo&Hallucination Detection&\multirow{-2}{*}{2663}&\multirow{-2}{*}{Pointwise \& Pairwise}&\multirow{-2}{*}{SFT \& DPO}&\multirow{-2}{*}{Mistral-7B-Instruct}\\

\rowcolor{mygray}
HelpSteer2~\cite{wang2024helpsteer2}&Human&Various&-&Pointwise \& Pairwise&PPI \& RLHF&Llama3.1-70B-Instruct\\

INSTRUCTSCORE~\cite{xu2023instructscore}&GPT-4&Various&40K&Pointwise \& Pairwise&SFT&Llama-7B\\

\rowcolor{mygray}
JudgeLM~\cite{zhu2023judgelm}&GPT-4&Open-ended Tasks&100K&Pairwise&SFT&Vicuna Series\\

LLaVA-Critic~\cite{xiong2024llava}&GPT-4o&Various&113K&Pointwise \& Pairwise&DPO&LLaVA-OneVision(OV) 7B \& 72B\\

\rowcolor{mygray}
Meta-Rewarding~\cite{wu2024meta}&Llama3&Various&20K&Pairwise&DPO&Llama3-8B-Instruct\\

OffsetBias~\cite{park2024offsetbias}&Human \& LLM&Bias Detection&268K&Pairwise&RLHF&Llama3-8B-Instruct\\

\rowcolor{mygray}
PandaLM~\cite{wang2023pandalm}&Human&Various&300K&Pairwise&SFT&Llama-7B\\

PHUDGE~\cite{deshwal2024phudge}&Human \& GPT-4&NLG&-&Pointwise \& Pairwise&LoRA&Phi-3\\

\rowcolor{mygray}
PROMETHEUS~\cite{kim2023prometheus}&Human&Various&100K&Pointwise&SFT&Llama2-Chat-7B \& 13B\\

PROMETHEUS2~\cite{kim2024prometheus}&Human&Various&300K&Pointwise \& Paiwise&SFT&Mistral-7B \& Mistral-8x7B\\

\rowcolor{mygray}
PROMETHEUS-VISION~\cite{lee2024prometheusvision}&GPT-4V&Various&15K&Pointwise&SFT&Llava-1.5\\

&&Common, Coding,&&&&\\
\multirow{-2}{*}{SELF-J~\cite{ye2024self}}&\multirow{-2}{*}{Human \& GPT-4}&\& Academic&\multirow{-2}{*}{5.7M}&\multirow{-2}{*}{Pointwise}&\multirow{-2}{*}{LoRA}&\multirow{-2}{*}{Llama2-13B}\\

\rowcolor{mygray}
Self-Rationalization~\cite{trivedi2024self}&LLM&Various&-&Pointwise \& Pairwise&SFT \& DPO&Llama3.1-8B-Instruct\\

Self-Taught~\cite{wang2024self}&LLM&Various&20K&Pairwise&-&Llama3-70B-Instruct\\

\rowcolor{mygray}
Shepherd~\cite{wang2023shepherd}&Human&Various&-&Pointwise \& Pairwise&SFT&Llama-7B\\

SorryBench~\cite{xie2024sorry}&Human \& GPT-4&Unsafe Topics&2.7K&Pointwise&SFT&Multiple LLMs\\

\rowcolor{mygray}
Themis~\cite{hu2024themis}&Human \& GPT-4&NLG&67K&Pointwise \& Pairwise&SFT \& DPO&Llama3-8B\\

TIGERScore~\cite{jiang2023tigerscore}&Human \& GPT-4&Text Generation&42K&Pointwise&SFT&Llama2-7B \& 13B\\

\rowcolor{mygray}
X-EVAL~\cite{liu2023x}&Human&NLG&55,602&Pointwise \& Pairwise&SFT&Flan-T5\\

\hline
\end{tabular}
}
\end{table*}

\subsubsection{Post-processing}
\label{sec:Post-processing}
\par Post-processing involves further refining evaluation results to extract more precise and reliable outcomes. This step typically includes analyzing the initial outputs to identify patterns, inconsistencies, or areas requiring improvement, followed by targeted adjustments and in-depth analysis. By addressing these issues, post-processing ensures that the evaluation results are not only accurate but also aligned with the specific objectives and standards of the task.

\textbf{Probability Calibration.}
During the post-hoc process of the model output, some studies use rigorous mathematical derivations to quantify the differences, thereby optimizing them.
For instance, Daynauth et al.~\cite{daynauth2024aligning} investigates the discrepancy between human preferences and automated evaluations in language model assessments, particularly employs Bayesian statistics and a t-test to quantify bias towards higher token counts, and develops a recalibration procedure to adjust the GPTScorers.
Apart from that, ProbDiff~\cite{xia2024language} is another novel self-evaluation method for LLMs that assesses model efficacy by computing the probability discrepancy between initial responses and their revised versions.
Moreover, Liusie et al.~\cite{liusie2024efficient} introduces a Product of Experts (PoE) framework for efficient comparative assessment using LLMs, which yield an expression that can be maximized with respect to the underlying set of candidates. This paper proposes two experts, a soft Bradley-Terry expert and a Gaussian expert that has closed-form solutions.
Unlike from frameworks above, CRISPR~\cite{yang2024mitigating} is a novel bias mitigation method for LLMs executing instruction-based tasks, which identifies and prunes bias neurons with probability calibration, reducing bad performance without compromising pre-existing knowledge.

\textbf{Text Reprocessing.} 
In LLMs-as-judges, text reprocessing methods are essential for enhancing the accuracy and reliability of evaluation outcomes. Specifically, text processing can improve the evaluation process by integrating multiple evaluation results or outcomes from several rounds of assessment.
For example, Sottana et al.~\cite{sottana2023evaluation} employs a multi-round evaluation process. Each round involves scoring model outputs based on specific criteria, with the human and GPT-4 evaluations ranking model performances from best to worst and averaging these rankings to mitigate subjectivity.
For the single-response evaluation, AUTO-J~\cite{li2023generative} employs a "divide-and-conquer" strategy. Critiques that either adhere to or deviate from the scenario-specific criteria are consolidated to form a comprehensive evaluation judgment and then generate the final assessment.
Consistent with former aforementioned studies, Yan et al.~\cite{yan2024consolidating} introduces a post-processing method to consolidate the relevance labels generated by LLMs. It demonstrates that this approach effectively combines both the ranking and labeling abilities of LLMs through post-processing.
Furthermore, REVISEVAL~\cite{zhang2024reviseval} is a novel evaluation paradigm that enhances the reliability of LLM Judges by generating response-adapted references through text revision capabilities of LLMs.
Apart from that, Tessler et al.~\cite{tessler2024ai} explores the use of AI as a mediator in democratic deliberation, aiming to help diverse groups find common ground on complex social and political issues. With the goal of maximizing group approval, the researchers developed the "Habermas Machine", which iteratively generate group statements based on individual opinions.

Another category of text reprocessing methods involves task transformation, primarily focusing on the conversion between open-ended and multiple-choice question (MCQ) formats. Ren et al.~\cite{ren2023self} explores the use of self-evaluation to enhance the selective generation capabilities of LLMs. Specifically, the authors reformulate open-ended generation tasks into token-level prediction tasks, reduce sequence-level scores to token-level scores to improve quality calibration.
Conversely, Myrzakhan et al.~\cite{myrzakhan2024open} introduces the Open-LLM-Leaderboard, a new benchmark for evaluating LLMs using open-style questions, which eliminates selection bias and random guessing issues associated with multiple-choice questions. It presents a method to identify suitable open-style questions and validate the correctness of LLM open-style responses against human-annotated ground-truths.

\subsection{Multi-LLM System}
\label{sec:Multi-LLM System}

\par Multi-LLM Evaluation harnesses the collective intelligence of multiple LLMs to bolster the robustness and reliability of evaluations. By either facilitating inter-model communication or independently aggregating their outputs, these systems can effectively mitigate biases, leverage complementary strengths across different models, refine decision-making precision, and foster a more nuanced understanding of complex judgments.

\subsubsection{Communication}
\label{sec:Communication}
\par Communication means the dynamic flow of information between LLMs, which is pivotal for sparking insights and sharing rationales during the judgment process. 
Recent research has shown that communication among LLMs can enable emergent abilities through their  interactions~\cite{xu2023towards}, leading to a cohesive decision-making process and better judgment performance.
% By breaking down the walls of isolated reasoning, these interactions bring LLMs together into a cohesive decision-making entity and lead to better responses.
The Multi-LLM system can benefit from LLM interactions in two ways: cooperation and competition.

\textbf{Cooperation.} 
Multi-LLMs can work together to achieve a common goal with information and rationales sharing through interactions to enhance the overall evaluation process.
For example, \citet{zhang2023wider} proposed an architecture named WideDeep to aggregate information at the LLM's neuro-level.
In addition, \citet{xu2023towards} introduced a multi-agent collaboration strategy that mimics the academic peer review process to enhance complex reasoning in LLMs. 
The approach involves agents creating solutions, reviewing each other’s work, and revising their initial submissions based on feedback.
Similarly, ABSEval~\cite{liang2024abseval} utilizes four agents for answer synthesize, critique, execution, and commonsense, to build the overall workflow.
Although the cooperation can complement each other's strengths between LLMs to a certain degree, this method still includes the risk of groupthink, where similar models reinforce each other’s biases rather than providing diverse insights.

\textbf{Competition.} Multi-LLMs systems can also benefit from competitive or adversarial communication, i.e., LLMs argue or debate to evaluate each other's outputs~\cite{zhao2024auto,moniri2024evaluating,chan2023chateval,li2023prd}. 
Such multi-LLMs systems could be categorized into centralized and decentralized structures~\cite{owens2024multi}.

In the centralized structure, a single central LLM acts as the orchestrator of the conversation, highlighting the efficiency of a unified decision-making process.
Auto-Arena~\cite{zhao2024auto} is such a novel framework that automates the evaluation of LLMs through agent peer battles and committee discussions, aiming to provide timely and reliable assessments. In detail, the framework conducts multi-round debates between LLM candidates, and uses an LLM judge committee to decide the winner.
Inspired by courtroom dynamics, \citet{bandi2024adversarial} propose two architectures, MORE and SAMRE, which utilize multiple advocates and iterative debates to dynamically assess LLM outputs.

In contrast, the decentralized structure emphasizes a collective intelligence where all models engage in direct communication, promoting a resilient and distributed decision-making structure.
In the domain of LLM debates, Moniri et al.~\cite{moniri2024evaluating} introduced a unique automated benchmarking framework, employing another LLM as the judge to assess not only the models' domain knowledge but also their abilities in problem definition and inconsistency recognition.
ChatEval~\cite{chan2023chateval} is another multi-agent debate framework that utilizes multiple LLMs with diverse role prompts and communication strategies on open-ended questions and traditional NLG tasks, significantly improves evaluation performance compared to single-agent methods.
Moreover, PRD~\cite{li2023prd} applied peer rank and discussion to address issues like self-enhancement and positional bias in current LLM evaluation methods, leading to better alignment with human judgments and a path for fair model capability ranking.

\subsubsection{Aggregation}
\label{sec:Aggregation}

Alternatively, in multi-LLM systems without communication, judgments are independently generated by multiple models, which are subsequently synthesized into a final decision through various aggregation strategies. Techniques such as majority vote, weighted averages, and prioritizing the highest confidence predictions, each play a crucial role. These methods allow each model to assess without interference, and eventually extract and combine the most effective elements from each model's response.

Simple voting methods, such as majority voting, by selecting the most frequent answers, offers a straightforward approach to synthesize evaluations.
For example, Badshah et al.~\cite{badshah2024reference} introduced a reference-guided verdict method for evaluating free-form text using multiple LLMs as judges. Combining these LLMs through majority vote significantly improves the reliability and accuracy of evaluations, particularly for complex tasks.
Furthermore, PoLL~\cite{verga2024replacing} demonstrates that using a diverse panel of smaller models as judges through max voting and average pooling is not only an effective method for evaluating LLM performance, but also reduces intra-model bias of a single large model.
Language-Model-as-an-Examiner~\cite{bai2024benchmarking} is another benchmarking framework to evaluate the performance of foundation models on open-ended question answering through voting. In the peer-examination mechanism, the LM serves as a knowledgeable examiner that formulates questions and evaluates responses in a reference-free manner.
What's more, multi-LLM evaluation could also be used in improving dataset quality. Choi et al.~\cite{choi2024multi} provided an enhanced dataset, MULTI-NEWS+, which is the result of a cleansing strategy leveraging CoT and majority voting to identify and exclude irrelevant documents through LLM-based data annotation.

Weighted scoring aggregation involves assigning different importance to different model outputs, either by aggregating multiple overall scores for the same response or by combining assessments of different aspects of the response to form a comprehensive evaluation.
On the one hand, through a peer-review mechanism, PiCO~\cite{ning2024pico} allows LLMs to answer unlabeled questions and evaluate each other without human annotations. It formalizes the evaluation as a constrained optimization problem, maximizing the consistency between LLMs' capabilities and corresponding weights.
Likewise, PRE~\cite{chu2024pre,chen2024automaticcostefficientpeerreviewframework} can automatically evaluate LLMs through a peer-review process. It selects qualified LLMs as reviewers through a qualification exam and aggregates their ratings using weights which is proportional to their agreement of humans, demonstrating effectiveness and robustness in evaluating text summarization tasks.
In the field of recommendation explanations, Zhang et al.~\cite{zhang2024large} suggests that ensembles like averaging ratings of multiple LLMs can enhance evaluation accuracy and stability.
On the other hand, for example, AIME~\cite{patel2024aime} is an evaluation protocol that utilizes multiple LLMs that each with a specific role independently generate an evaluation on separate criteria and then combine them via concatenation.
Similarly, a paper introduces HD-EVAL~\cite{liu2024hd}, which iteratively aligns LLM-based evaluators with human preference via Hierarchical Criteria Decomposition. By decomposing a given evaluation task into finer-grained criteria, aggregating them according to estimated human preferences, pruning insignificant criteria with attribution, and further decomposing significant criteria, HD-EVAL demonstrates its superiority.

Apart from weighting methods, there are some other advance mathematical aggregation techniques, such as Bayesian methods and graph-based approaches, offering more robust ways to handle uncertainties and inconsistencies across multiple evaluators.
Notably, a paper introduces two calibration methods, Bayesian Win Rate Sampling (BWRS) and Bayesian Dawid-Skene~\cite{gao2024bayesian}, to address the win rate estimation bias when using many LLMs as evaluators for text generation quality.
In addition to that, GED~\cite{hu2024language} addresses inconsistencies in LLM preference evaluations by leveraging multiple weak evaluators to construct preference graphs, and then utilize DAG structure to ensemble and denoise these graphs for better, non-contradictory evaluation results.

LLM-based aggregation is a grand-new perspective like Fusion-Eval~\cite{shu2024fusion}. It's a novel framework that integrates various assistant evaluators using LLMs, each of which specializes in assessing distinct aspects of responses, to enhance the correlation of evaluation scores with human judgments for natural language systems.

In addition to the above direct use of multiple model evaluation, the cascade framework employs a tiered approach, where weaker models are used initially for evaluations, and stronger models are engaged only when higher confidence is required, optimizing resource use and enhancing evaluation precision.
Jung et al.~\cite{jung2024trust} proposes "Cascaded Selective Evaluation" to ensure high agreement with human judgments while using cheaper models.
Similar to the work above, Huang et al.~\cite{huang2024limitations} proposes CascadedEval, a novel method integrating proprietary models, in order to compensate for the limitations of fine-tuned judge models.

\subsection{Human-AI Collaboration System}
\label{sec:Hybrid System}

Human-AI Collaboration Systems bridge the gap between automated LLM judgments and the essential need for human oversight, particularly in high-stakes domains such as law, healthcare, and education. Human evaluators act either as the ultimate deciders, or as intermediaries who verify and refine model outputs. By incorporating human insights, Hybrid systems can ensure the final judgment is more reliable and aligned with ethical considerations, and empower continuous model improvement through feedback loops.

In many Human-AI Collaboration systems, human evaluators play a vital role during the evaluation process itself, actively collaborating with the LLMs to review and refine the generated outputs.
For example, COEVAL\cite{li2023collaborative} introduces a collaborative evaluation pipeline where LLMs generate initial criteria and evaluations for open-ended natural language tasks. These machine-generated outputs are then reviewed and refined by human evaluators to guarantee reliability.
To address a significant positional bias in LLMs when used as evaluators, Wang et al.~\cite{wang2023large} proposes a calibration framework with three strategies: Multiple Evidence Calibration, Balanced Position Calibration, and Human-in-the-Loop Calibration.
Similarly, EvalGen\cite{shankar2024validates} integrates human feedback iteratively to refine evaluation criteria, addressing challenges such as  ``criteria drift'', where the standards of evaluation evolve as humans interact with the model.
These systems allow human evaluators to provide real-time adjustments, enhancing the accuracy and trustworthiness of the evaluation process.

While in other systems, human involvement takes place after the LLM has completed its evaluations, providing a final layer of verification and adjustment. This method ensures that the LLM's judgments are thoroughly scrutinized and aligned with human values.
EvaluLLM\cite{pan2024human} allows humans to intervene and refine the evaluation results, thereby enhancing trust in the model’s performance while also controlling for potential biases.
Additionally, Chiang et al.\cite{chiang2024large} tried LLM TAs as an assignment evaluator in a large university course. After students submit assignments and receive LLM-generated feedback, the teaching team reviews and finalizes the evaluation results. This process illustrates how human oversight after the initial automated evaluation can guarantee fairness and consistentcy with academic standards.

\section{Application}
\label{sec:Domain}

Due to the convenience and effectiveness of LLM Judges, they have been widely applied as judges across various domains.
These applications not only cover general domains but also specific domains such as multimodal, medical, legal, financial, education, information retrieval and others. In this section, we will provide a detailed introduction to these applications, demonstrating how LLMs achieve precise and efficient evaluations in different domains.

\begin{figure}[t]
    \centering
    \includegraphics[width=\textwidth]{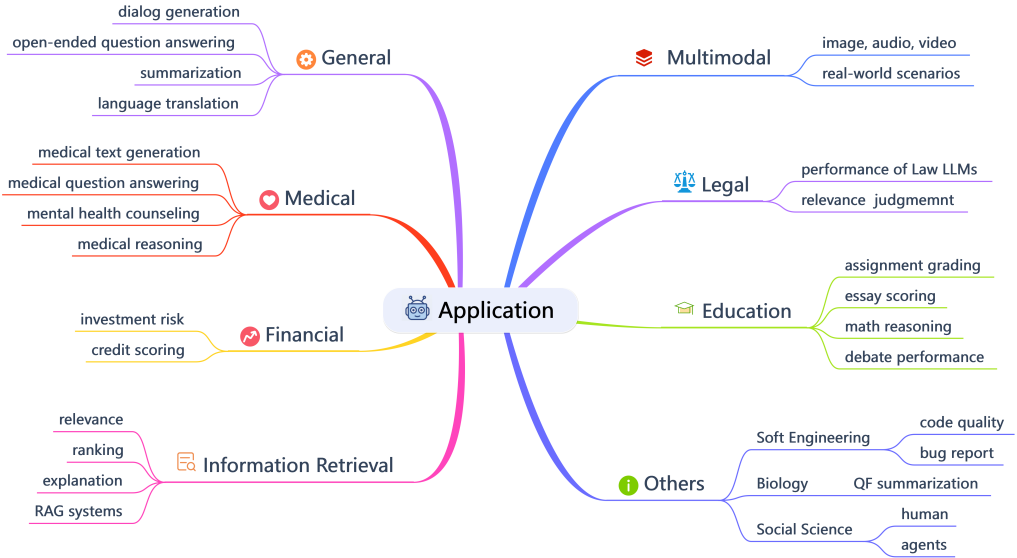}
    \caption{LLMs-as-judges are widely applied across various domains.}
    \label{figure:application}
\end{figure}

\subsection{General}
\label{sec:General}
In general domains, LLM Judges are applied to tasks requiring both understanding and generation, such as dialogue generation, open-ended question answering, summarization, and language translation. Each task follows its own set of evaluation criteria to meet its specific requirements.
For instance, in dialogue generation \cite{li2017dailydialog}, the criteria emphasize the natural flow, emotional resonance, and contextual relevance of the conversation. In summarization tasks \cite{narayan2018don}, the evaluation focuses on the coherence, consistency, fluency, and relevance of the text. In translation tasks \cite{feng2024improving}, the assessment prioritizes the quality, accuracy, fluency, and style.

As these diverse sub-tasks require specialized evaluation criteria, LLM judges provides refined evaluation methods that go beyond traditional metrics, paving the way for more comprehensive and in-depth assessments. For instance, Shu et al. \cite{shu2024fusion} introduced Fusion-Eval, an innovative approach that leverages LLMs to integrate insights from various assistant evaluators. Fusion-Eval evaluated summary quality across four dimensions—coherence, consistency, fluency, and relevance, achieving a system-level Kendall-Tau correlation of 0.962 with human judgments. For dialogue quality, it assessed six aspects: coherence, engagingness, naturalness, groundedness, understandability, and overall quality, attaining a turn-level Spearman correlation of 0.744. Furthermore, Xu et al. \cite{xu2024large} proposed the ACTIVE-CRITIC framework, which enables LLMs to actively infer the target task and relevant evaluation criteria while dynamically optimizing prompts. In the story generation task, this framework achieved superior evaluation performance.

\subsection{Multimodal}
\label{sec:Multimodal}
In the multimodal domain, the evaluation objects of LLMs are not limited to textual data but extend to various forms of information such as images, audio, and video. One of the primary challenges in evaluating multimodal tasks lies in the significant heterogeneity among these modalities, including substantial differences in data structures, representation methods, and feature distributions.

To address this challenge, advanced techniques are often required to help LLMs integrate different forms of information, ensuring that they can provide accurate and meaningful evaluations. For example, Xiong et al. \cite{xiong2024llava} trained an open-source multimodal LLM, LLaVA-Critic, specifically to evaluate model performance in multimodal scenarios. Similarly, Chen et al. \cite{chen2024mllm} developed a Multimodal LLM-as-a-judge for 14 Vision-Language tasks, providing a unified evaluation framework. In addition, LLMs-as-judges can also be used in audio. For instance, Latif et al. \cite{latif2023can} used LLMs for identifying and evaluating emotional cues in speech, achieving remarkable accuracy in the process. Beyond these efforts, some recent studies \cite{zhou2024calibrated, deng2024efficient} have also explored the potential of multimodal LLMs to self-evaluate and self-reward, enhancing their performance without the need for external evaluators or human annotations.

As the application of LLMs-as-judges continues to expand in multimodal domains, there is a growing interest in exploring their use in more specific real-world scenarios, such as autonomous driving. Chen et al. \cite{chen2024automated} proposed CODA-LM, a novel vision-language benchmark for self-driving, which provides automatic and systematic evaluation of Large Vision-Language Models (LVLMs) on road corner cases. Interestingly, they found that using the text-only LLM judges resulted in a closer alignment with human preferences than LVLMs.

\subsection{Medical}
\label{sec:Medical}
In the medical field, LLMs-as-judges have demonstrated significant potential, particularly in areas such as diagnostic support, medical text analysis, clinical decision-making, and patient education.
In this domain, high-quality evaluation requires LLM judges to possess precise interpretation capabilities for domain-specific terminology, the ability to comprehensively analyze diverse data types (such as clinical records and medical imaging), and strict compliance with high accuracy standards and ethical guidelines.

In the realm of \textbf{medical text generation}, Xie et al. \cite{xie2024doclens} used LLMs to evaluate the compduikeyi1leteness, conciseness, and attribution of medical texts at a fine-grained level. Similarly, Brake et al. \cite{brake2024comparing} leveraged LLMs, such as Llama2, to assess clinical note consistency, with results indicating agreement levels comparable to human annotators. When it comes to \textbf{medical question answering}, Krolik et al. \cite{krolik2024towards} explored the use of LLMs to automatically evaluate answer quality. Their focus was on evaluating adherence to medical knowledge and professional standards, completeness of information, accuracy of terminology, clarity of expression, and relevance to the question. 

In the area of \textbf{mental health counseling}, Li et al. \cite{li2024automatic} utilized LLMs to automate the evaluation of counseling effectiveness and quality. Key assessments included whether the counseling identified the client’s emotional needs, provided appropriate responses, demonstrated empathy, managed negative emotions, and met the overall goals of mental health support. Beyond these above applications, LLMs' judging capabilities have also been applied to assist in improving performance in specialized \textbf{medical reasoning tasks}. For instance, many studies \cite{jeong2024improving} employed LLMs to evaluate and filter medical information, thereby supporting enhanced medical reasoning.

\subsection{Legal}
\label{sec:Legal}

Due to the powerful evaluation capabilities of LLMs, LLMs-as-judges have been widely applied in the legal domain, covering multiple key scenarios, including evaluating the performance of law LLMs and relevance judgment in legal case retrieval. In the legal domain, the application of LLMs requires a deep understanding of the legal framework of specific jurisdictions, complex legal language, and rigorous logical reasoning abilities~\cite{li2024lexevalcomprehensivechineselegal}. At the same time, interpretability and transparency of the evaluation results are essential core requirements, as legal practice highly depends on clear logic and verifiable conclusions. Furthermore, the bias and fairness of the model are of significant concern, as any bias in legal evaluations could have a profound impact on judicial fairness. These unique demands set higher standards for LLM judges.

In response to these challenges, recent research has explored various ways in which LLMs can be effectively employed in legal evaluations. Some research used LLMs as judges to assist in \textbf{evaluating the performance of Law LLMs}. For example, Yue et al. \cite{yue2023disc} introduced DISC-LawLLM to provide a wide range of legal services and utilized GPT-3.5 as a referee to evaluate the model’s performance. They assessed three key criteria—accuracy, completeness, and clarity—by assigning a rating score from 1 to 5. Similarly, Ryu et al. \cite{ryu2023retrieval} applied retrieval-based evaluation to assess the performance of LLMs in Korean legal question-answering tasks, which applied RAG not for generation but for evaluation. What's more, LLMs have also been utilized to construct evaluation sets. Raju et al. \cite{raju2024constructing} explored methods for constructing these domain-specific evaluation sets, which are essential for enabling LLMs-as-judges to perform effective evaluation in legal domain. Beyond performance evaluation, LLMs have also been utilized for \textbf{relevance judgment} in legal case retrieval. For instance, Ma et al. \cite{ma2024leveraging} used LLMs to automate the evaluation of large numbers of retrieved legal documents, improving both the scalability and accuracy of legal case retrieval systems. In conclusion, the application of LLMs-as-judges in law holds significant promise in future.

\subsection{Financial}
\label{sec:Financial}

In the financial domain, LLM judges have been extensively explored in scenarios such as investment risk assessment and credit scoring, which presenting unique challenges. For example, the complexity of risk assessment requires LLMs to accurately capture the influence of multifaceted factors, including market volatility, regulatory changes, and geopolitical events. Real-time processing demands further elevate the challenge, requiring LLMs not only to be computationally efficient but also to deliver rapid response times. Additionally, the dynamic nature of high-frequency trading demanding that LLMs swiftly adapt to fluctuating market conditions.

In \textbf{investment risk assessment}, LLMs have proven effective due to their ability to process large amounts of data and make informed judgments. For instance, Xie et al. \cite{xie2023pixiu} developed a financial LLM, FinMA, fine-tuned on LLaMA to evaluate investment risks more effectively. Their model is designed to follow instructions for risk assessment and decision analysis, improving the accuracy and efficiency of financial evaluations.

Another key application in the financial domain is \textbf{credit scoring}, which predicts the future repayment ability and default risk of individuals or businesses. By analyzing a vast array of data, including credit history, financial status, and other relevant factors, LLMs can help financial institutions make more accurate credit scoring assessments. For example, Babaei et al. \cite{babaei2024gpt} demonstrated how LLMs can process unstructured text data, such as customer histories, contract terms, and news reports, to enhance the precision of credit assessments.

Furthermore, as the use of LLMs in finance continues to grow, there is a rising need to \textbf{evaluate the performance of these financial LLMs}. To address this, Son et al. \cite{son2024krx} developed an automated financial evaluation benchmark that leverages LLMs to extract valuable insights from both unstructured and structured data. This framework helps optimize the construction, updating, and compliance checks of financial benchmarks, supporting more efficient and scalable evaluation processes. Based on this, they facilitated the continuous optimization of financial LLMs, driving further advancements in the financial domain.

\subsection{Education}
\label{sec:Education}
LLMs-as-judges have found extensive applications in the education domain, covering a wide range of tasks, such as grading student assignments, evaluating essays, assessing mathematical reasoning, and judging debate performance. These applications present several key challenges, including the diversity of student responses and individual differences, as well as the need for multidimensional evaluation. Effective evaluation in education requires LLMs to consider not only correctness but also creativity, clarity, and logical coherence. Additionally, the interpretability and fairness of the evaluation results are crucial, as educational assessments significantly impact students' development and future opportunities.

In \textbf{assignment grading}, Chiang et al. \cite{chiang2024large} introduced
the concept of an LLM Teaching Assistant (LLM TA) in university classrooms, utilizing GPT-4 to
automate the grading of student assignments. By employing prompt engineering to define scoring
criteria and task descriptions, LLM TA generates quantitative scores and detailed feedback. Their
study emphasized the system’s ability to maintain consistency, adhere to grading standards, and
resist adversarial prompts, highlighting its robustness and practicality for classroom use.

In addition to assignment grading, LLMs-as-judges are also being explored for \textbf{automated essay scoring.} Wang et al. \cite{wang2024automated} proposed an advanced intelligent essay scoring system, integrating LLMs such as BERT and ChatGPT to enable automated scoring and feedback generation for essays across various genres. Similarly, Song et al. \cite{song2024automated} investigated a framework and methodology for automated essay scoring and revision based on open-source LLMs. Furthermore, Zhou et al. \cite{zhou2024llm} explored the potential of LLMs in academic paper reviewing tasks, assessing their reliability, effectiveness, and possible biases as reviewer. They found that while LLMs show certain promise in the domain of automated reviewing, they are not yet sufficient to fully replace human reviewers, particularly in areas with high technical complexity or strong innovation.

Another area where LLMs-as-judges are making an impact is in the evaluation of \textbf{math reasoning}. Unlike traditional mathematical task evaluation, which focuses solely on the correctness of the final results, Xia et al. \cite{xia2024evaluating} argued that additional aspects of the reasoning process should also be assessed, such as logical errors or unnecessary steps. In their work, the authors proposed ReasonEval, a new methodology for evaluating the quality of reasoning steps based on LLMs-as-judges.

LLMs have also been employed in \textbf{judging debate performance}. Liang et al. \cite{liang2024debatrix} proposed Debatrix, a new method which leverages LLMs to evaluate and analyze debates. The main aspects assessed include the logical consistency of arguments, the effectiveness of rebuttals, the appropriateness of emotional expression, and the coherence of the debate.

\subsection{Information Retrieval}
\label{sec: Information Retrieval}
Information retrieval refers to the process of effectively retrieving, filtering, and ranking relevant information from a large collection of data, matching information resources to users' needs (queries). However, evaluating these systems presents several challenges, particularly due to the the complexity of real-world data, the diversity of user needs, and personalization. To solve these challenges, LLMs-as-judges have been used across various applications, including relevance judgment, text ranking, recommendation explanations evaluation, and assessing retrieval-augmented generation (RAG) systems.

One key area in information retrieval is the evaluation of \textbf{the relevance of retrieved results to user queries}, a task that traditionally relies on manual annotations~\cite{soboroff2024don,rahmani2024llmjudge}. Rahmani et al. \cite{rahmani2024llmjudge} proposed a framework called LLMJudge which leveraged LLMs to assess the relevance of information retrieval system results to user queries, providing a more scalable and efficient evaluation approach.

Another important aspect of information retrieval is the \textbf{ranking of search results or recommendation lists}. Traditional ranking models often rely on shallow features or direct matching scores, which may not yield optimal results. To address this,  Qin et al. \cite{qin2023large} examined the performance of LLMs in text ranking tasks and proposed a novel method based on pairwise ranking prompting, utilizing LLMs for text ranking. Additionally, Niu et al. \cite{niu2024judgerank} introduced a framework called JudgeRank, which leveraged LLMs to rerank results in reasoning-intensive tasks. By evaluating the logic, relevance, and quality of candidate results, this approach tried to enhance ranking performance.

In recommendation systems, \textbf{explanation evaluation} plays a crucial role in helping users understand why a specific product, movie, or piece of content is recommended. Zhang et al. \cite{zhang2024large} investigated the potential of LLMs as automated evaluators of recommendation explanations, assessing them across multiple dimensions such as quality, clarity, and relevance. This approach provides a more efficient way to evaluate the effectiveness of explanations, which is essential for improving user trust and satisfaction.

Furthermore, with the growing use of \textbf{retrieval-augmented generation (RAG) systems} in tasks like question answering, fact-checking, and customer support, there is an increasing need to evaluate the quality of these systems. Traditional evaluation methods rely on large manually annotated datasets, which are time-consuming and costly. To address this, Saad et al. \cite{saad2023ares} proposed a new automated evaluation framework called ARES, which leveraged LLMs as the core evaluation tool to directly assess retrieval and generated content across multiple dimensions, including relevance, accuracy, coverage, fluency, and coherence.

\subsection{Others}
\label{sec: Others}
\subsubsection{Soft Engineering}
\label{sec: Soft Engineering}
The challenges that LLMs-as-judges need to overcome in the software engineering domain include complex code structures and the diversity of evaluation criteria. A numerous of articles \cite{patel2024aime, weyssow2024codeultrafeedback} used LLMs-as-judges to assess the quality of code generation. Moreover, Kumar et al. \cite{kumar2024llms} employed LLMs to evaluate the quality of Bug Report Summarization.

\subsubsection{Biology}
\label{sec: Biology}
The main evaluation challenges in biological field include the complexity and diversity of the data and the need for specialized biological knowledge~\cite{o2023bioplanner,hijazi2024using}. For example, Hijazi et al. \cite{hijazi2024using} used LLMs to evaluate Query-focused summarization (QFS), which refers to generating concise and accurate summaries from a large set of biomedical literature based on a specified query. In this context, the LLMs are used to assess whether these summaries accurately answer the specified query and whether they cover the correct biological knowledge.
\subsubsection{Social Science}
\label{sec: Social Science}
LLMs-as-Judges have also found applications in social sciences. On one hand, they are used in \textbf{real-world human social contexts}. For example, Tessler et al. \cite{tessler2024ai} used LLMs to participate in democratic discussions, assessing the quality of arguments, identifying fallacies, or providing a balanced view of an issue, thus helping people reach consensus on complex social and political matters. On the other hand, LLMs-as-judges are also used in \textbf{social scenarios constructed by language agents}. Zhou et al. \cite{zhou2023sotopia} proposed an interactive evaluation framework called Sotopia, which used LLMs to assess the social intelligence of language agents from multiple dimensions, such as emotional understanding, response adaptability, and other social skills.

In this section, we have outlined the specific applications of LLMs-as-judges across various domains. In these applications, LLMs leverage their powerful text understanding and generation capabilities to perform effective evaluations and judgments, providing accurate feedback and improvement suggestions. Although LLMs-as-judges have shown tremendous potential in these areas, especially in handling large-scale data and automating assessments, they still face challenges such as \textbf{the depth of domain-specific knowledge, limitations in reasoning abilities, and the diversity of evaluation criteria}. In the future, with continuous improvements in model performance and domain adaptation capabilities,, we believe the application of LLMs-as-judges will become more widespread and precise across various domains.

\section{Meta-evaluation}
\label{sec:Meta}
Meta-evaluation, the process of assessing the quality of the evaluator itself, is a crucial step in determining the reliability, consistency, and validity of LLM judges. Given the diverse applications of LLMs as evaluators, meta-evaluation methods have also been evolving. Researchers have proposed various datasets and metrics tailored to different tasks and evaluation objectives to assess the reliability and validity of LLM-based evaluations. This chapter will explore state-of-the-art Benchmarks (\S\ref{sec:benchmarks}) and evaluation Metrics (\S\ref{sec:metric}), categorize existing approaches, and discuss their advantages and limitations.

\begin{table*}[ht]
\footnotesize
\caption{Statistical information of different benchmarks (Part 1).}
\begin{tabular}{lccccc}
\hline
\textbf{Benchmark}                                                                   & \textbf{Task}      & \textbf{Type}                                                         & \textbf{Num}                                                    & \textbf{Evaluation Criteria}                                                                                                                    & \textbf{Language}     \\ \hline  \hline
\rowcolor{mygray}
HumanEval~\cite{chen2021evaluating}                                                          & Code        & Pointwise                                                    & 164                                                    & Functional correctness                                                                                                                 & English      \\
SWE-bench~\cite{jimenez2023swe}                                                                   & Code        & Pointwise                                                    & 2,294                                                  & Task solve                                                                                                                             & English      \\
\rowcolor{mygray}
DevAI~\cite{zhuge2024agent}                                                                     & Code        & Pointwise                                                    & 365                                                    & \begin{tabular}[c]{@{}c@{}}Disagreement, Task solve, \\ Requirements met\end{tabular}                                                  & English      \\
CrossCodeEval~\cite{ding2024crosscodeeval}                                                               & Code        & Pointwise                                                    & 1,002                                                  & Code match, Identifier match                                                                                                           & English      \\
\rowcolor{mygray}
CodeUltraFeedback~\cite{weyssow2024codeultrafeedback}                                                           & Code        & \begin{tabular}[c]{@{}c@{}}Pointwise\\ Pairwise\end{tabular} & 10k                                                    & \begin{tabular}[c]{@{}c@{}}Code explanation, \\ Code complexity and efficiency, \\ Code readability, Coding style\end{tabular}         & English      \\
\begin{tabular}[c]{@{}l@{}}Literary Translation \\ Comparisons~\cite{karpinska2023large}\end{tabular} & Translation & Pairwise                                                     & 720                                                    & Translation quality and errors                                                                                                         & Multilingual \\
\rowcolor{mygray}
MQM~\cite{freitag2021experts}                                                                         & Translation & Pointwise                                                    & 100k                                                   & \begin{tabular}[c]{@{}c@{}}Accuracy, Fluency, \\ Terminology, Style, Locale\end{tabular}                                               & Multilingual \\
\begin{tabular}[c]{@{}l@{}}WMT Metrics \\ Shared Task~\cite{freitag2021results}\end{tabular}          & Translation & Pointwise                                                    & -                                                      & Adequacy, Fluency                                                                                                                      & Multilingual \\
\rowcolor{mygray}
SummEval~\cite{fabbri2021summeval}                                                                    & Summary     & Pointwise                                                    & 1,600                                                  & \begin{tabular}[c]{@{}c@{}}Coherence, Consistency, \\ Fluency, Relevance\end{tabular}                                                  & English      \\
Opinsummeval~\cite{shen2023opinsummeval}                                                                & Summary     & Pointwise                                                    & 1,400                                                  & \begin{tabular}[c]{@{}c@{}}Aspect relevance, \\ Self-coherence, Readability\\ Sentiment consistency,\end{tabular}                      & English      \\
\rowcolor{mygray}
Frank ~\cite{pagnoni2021understanding}                                                                      & Summary     & Pointwise                                                    & 2,250                                                  & Factual errors                                                                                                                         & English      \\
Topical-Chat~\cite{gopalakrishnan2023topical}                                                                & Dialogue    & Pointwise                                                    & 60                                                     & \begin{tabular}[c]{@{}c@{}}Understandable, Natural, \\ Maintains context, Interesting, \\ Uses knowledge, Overall quality\end{tabular} & English      \\
\rowcolor{mygray}
Personal-Chat~\cite{zhang2018personalizing}                                                               & Dialogue    & Pointwise                                                    & 60                                                     & \begin{tabular}[c]{@{}c@{}}Understandable, Natural, \\ Maintains context, Interesting, \\ Uses knowledge, Overall quality\end{tabular} & English      \\
DSTC10 Hidden Set~\cite{zhang2021automatic}                                                           & Dialogue    & Pointwise                                                    & 9,500                                                  & \begin{tabular}[c]{@{}c@{}}Coherence, Appropriateness, \\ Naturalness, Toxicity control\end{tabular}                                   & English      \\
\rowcolor{mygray}
HANNA~\cite{chhun2022human}                                                                       & Story       & Pointwise                                                    & 1,056                                                  & \begin{tabular}[c]{@{}c@{}}Relevance, Coherence, \\ Empathy, Surprise, \\ Engagement, Complexity\end{tabular}                          & English      \\
MANS~\cite{guan2021openmeva}                                                                        & Story       & Pointwise                                                    & 2,000                                                  & Coherence                                                                                                                              & English      \\
\rowcolor{mygray}
StoryER~\cite{chen2023storyer}                                                                     & Story       & Pairwise                                                     & 100k                                                   & Upvoted                                                                                                                                & English      \\
Per-DOC~\cite{wang2023learning}                                                                     & Story       & Pointwise                                                    & 7,000                                                  & \begin{tabular}[c]{@{}c@{}}Interestingness, Adaptability,\\ Character developmentSurprise, Ending\end{tabular}                         & English      \\
\rowcolor{mygray}
PKU-SafeRLHF~\cite{ji2024pku}                                                                & Value       & Pairwise                                                     & 83.4K                                                  & Helpfulness, Harmlessness                                                                                                              & English      \\
HHH~\cite{askell2021general}                                                                         & Value       & Pairwise                                                     & 221                                                    & \begin{tabular}[c]{@{}c@{}}Helpfulness, Honesty, \\ Harmlessness\end{tabular}                                                          & English      \\
\rowcolor{mygray}
Cvalue~\cite{xu2023cvalues}                                                                      & Value       & Pairwise                                                     & 145k                                                   & Safety, Responsibility                                                                                                                 & Chinese      \\
Yelp  ~\cite{asghar2016yelp}                                                                      & Recom       & Pointwise                                                    & 8,630k                                                 & User perference                                                                                                                        & English      \\
\rowcolor{mygray}
Movielens\_Explanation~\cite{zhang2024large}                                                      & Recom       & Pointwise                                                    & 2,496                                                  & \begin{tabular}[c]{@{}c@{}}Persuasiveness, Transparency,   \\ Accuracy, Satisfaction\end{tabular}                                      & English      \\
Trec DL21\&22 ~\cite{DBLP:conf/trec/Craswell0YCL21,DBLP:conf/trec/Craswell0YCLVS22}                                                              & Search      & Pointwise                                                    & \begin{tabular}[c]{@{}c@{}}1,549/\\ 2,673\end{tabular} & Relevacne                                                                                                                              & English      \\
\rowcolor{mygray}
LeCarDv2~\cite{li2024lecardv2}                                                                    & Search      & Pointwise                                                    & 55,192                                                 & \begin{tabular}[c]{@{}c@{}}Characterization relevance, \\ Penalty relevance, \\ Procedure relevance\end{tabular}                       & English      \\ \hline
\end{tabular}
\end{table*}

\begin{table*}[ht]
\footnotesize
\caption{Statistical information of different benchmarks (Part 2).}
\begin{tabular}{lccccc}
\hline
\textbf{Benchmark}  & \textbf{Domain} & \textbf{Type}                                                           & \textbf{Num} & \textbf{Evaluation Criteria}                                                                                               & \textbf{Language}                                         \\ \hline \hline
\rowcolor{mygray}
UltraFeedback~\cite{cui2024ultrafeedback}        & Compre.         & Pairwise                                                                & 64k          & \begin{tabular}[c]{@{}c@{}}Helpfulness, Honesty,   \\ Instruction following, \\ Truthfulness\end{tabular}                  & English                                                   \\
AlpacaEval~\cite{dubois2024length}          & Compre.         & Pairwise                                                                & 20k          & Instruction-following                                                                                                      & English                                                   \\
\rowcolor{mygray}
Chatbot Arena~\cite{zheng2023judging}       & Compre.         & Pairwise                                                                & 33k          & User perference                                                                                                            & English                                                   \\
MTBench~\cite{zheng2023judging}             & Compre.         & Pairwise                                                                & 3,000        & \begin{tabular}[c]{@{}c@{}}Multi-turn conversational, \\ Instruction-following\end{tabular}                                & English                                                   \\
\rowcolor{mygray}
RewardBench~\cite{lambert2024rewardbench}        & Compre.         & Pairwise                                                                & 2,998        & User perference                                                                                                            & English                                                   \\
JudgerBench~\cite{cao2024compassjudger}         & Compre.         & Pairwise                                                                & 1,900        & Instruction following                                                                                                      & \begin{tabular}[c]{@{}c@{}}English\\ Chinese\end{tabular} \\
\rowcolor{mygray}
RM-Benchh~\cite{liu2024rm}            & Compre.         & Pairwise                                                                & 1,327        & Instruction following                                                                                                      & English                                                   \\
JUDGEBENCH~\cite{tan2024judgebench}          & Compre.         & Pairwise                                                                & 350          & Factual, Logical correctness                                                                                               & English                                                   \\
\rowcolor{mygray}
Infinity-Preference\footnote{\url{https://huggingface.co/datasets/BAAI/Infinity-Preference}} & Compre.         & Pairwise                                                                & 59.3k        & User perference                                                                                                            & \begin{tabular}[c]{@{}c@{}}English\\ Chinese\end{tabular}                                         \\
LLMeval~\cite{zhang2023wider}             & Compre.         & \begin{tabular}[c]{@{}c@{}}Pointwise\\ Pairwise\end{tabular}            & 453          & \begin{tabular}[c]{@{}c@{}}Correctness, Fluency, Logic, \\ Informativeness, Harmlessness\end{tabular}                      & Chinese                                                   \\
\rowcolor{mygray}
WildBench~\cite{lin2024wildbench}           & Compre.         & Pointwise                                                               & 1,024        & Checklists                                                                                                                 & English                                                   \\
Flask~\cite{ye2023flask}               & Compre.         & Pointwise                                                               & 1,740        & \begin{tabular}[c]{@{}c@{}}Logical thinking, \\ Background knowledge, \\ Problem handling, User alignment\end{tabular}     & English                                                   \\
\rowcolor{mygray}
AlignBench~\cite{liu2023alignbench}          & Compre.         & Pointwise                                                               & 683          & \begin{tabular}[c]{@{}c@{}}Task-oriented, Clarity \& Fluency, \\ Complexity \& Difficulty, \\ Desensitization\end{tabular} & Chinese                                                   \\
HELPSTEER~\cite{wang2023helpsteer}           & Compre.         & \begin{tabular}[c]{@{}c@{}}Pointwise\\ Pairwise\end{tabular}            & 37,120       & \begin{tabular}[c]{@{}c@{}}Helpfulness, Correctness, \\ Coherence, Complexity Verbosity\end{tabular}                       & English                                                   \\
\rowcolor{mygray}
HELPSTEER2~\cite{wang2024helpsteer2}          & Compre.         & \begin{tabular}[c]{@{}c@{}}Pointwise\\ Pairwise\end{tabular}            & 21,362       & \begin{tabular}[c]{@{}c@{}}Helpfulness, Correctness, \\ Coherence, Complexity, Verbosity\end{tabular}                      & English                                                   \\
MLLM-as-a-Judge~\cite{chen2024mllm}     & Compre.         & \begin{tabular}[c]{@{}c@{}}Pointwise\\ Pairwise\\ Listwise\end{tabular} & 17,000       & {\color[HTML]{4A4A4A} \begin{tabular}[c]{@{}c@{}}Relevance, Accuracy, \\ Creativity, Response granularity\end{tabular}}    & English                                                   \\
\rowcolor{mygray}
MM-EvalMM-Eval~\cite{son2024mm}             & Compre.         & Pairwise                                                                & 4,981        & Task-oriented                                                                                                              & Multilingual                                              \\ \hline
\end{tabular}
\end{table*}

\subsection{Benchmarks}
\label{sec:benchmarks}
To evaluate LLM-based judges, a common approach is to measure their alignment with human preferences, as human judgments are often considered the gold standard for quality and reliability. Given the diverse range of applications for LLM-based judges, different benchmarks have been created, each tailored to specific evaluation criteria and use cases.
In this section, we present a comprehensive collection of 40 widely-used benchmarks, each designed to capture different aspects of evaluation, such as language understanding, factual accuracy, coherence, creativity, and fairness. To enhance clarity and facilitate comparison, we categorize these benchmarks by application domain.

\subsubsection{Code Generation}
\label{sec:Code Generation}
Code generation aims to produce executable program code from natural language input. This task typically involves translating user requirements or descriptions into precise code. The applications of code generation are vast, including automated script creation, bug fixing, and the generation of complex programming tasks.Evaluating code generation is highly challenging, and LLMs are increasingly being used as evaluators for assessing code quality.

HumanEval~\cite{chen2021evaluating} is a widely used benchmark dataset designed to evaluate programming capabilities. It consists of 164 coding tasks, each accompanied by a brief natural language description. The tasks primarily involve algorithmic problems and data structure exercises, with difficulty levels ranging from basic to intermediate. 
One notable feature of HumanEval~\cite{chen2021evaluating} is the inclusion of input-output examples, which facilitate the assessment of functional correctness. However, the dataset's limited size and scope may not sufficiently capture the diversity of real-world programming challenges.
SWEBench~\cite{jimenez2023swe} targets more complex programming tasks that are closer to real-world software development scenarios. It includes 2,294 tasks requiring advanced operations such as reasoning, multi-step problem solving, and API usage. Unlike simpler benchmarks, SWEBench~\cite{jimenez2023swe} assesses the model's ability to handle comprehensive problem-solving and logical reasoning. However, the increased complexity also introduces challenges in establishing consistent evaluation criteria, particularly when it comes to subjective aspects like code style and efficiency. Moreover, DevAI~\cite{zhuge2024agent} was introduced to address the limitations of existing benchmarks, which often fail to capture the iterative nature of software development and lack adequate signals for measuring long-term progress. The dataset includes 365 task requirements, focusing on more complex and challenging programming scenarios.
CrossCodeEval~\cite{ding2024crosscodeeval} focuses on assessing cross-language programming models, containing over 1,000 tasks that involve translating code between different programming language pairs, such as Python to Java or JavaScript to C++. This dataset tests the model’s ability to adapt and transform code across languages, highlighting the challenges of understanding varied syntax and semantics. CodeUltraFeedback~\cite{weyssow2024codeultrafeedback} is designed to evaluate and enhance the alignment between LLMs and user-defined programming preferences. It includes 10,000 programming instructions, each paired with four responses from 14 different LLMs. These responses are scored by GPT-3.5 based on five distinct programming preferences, such as readability, efficiency, and adherence to user specifications. The dataset emphasizes fine-grained feedback and user-centered evaluation, making it a useful tool for analyzing preference alignment.

\subsubsection{Machine Translation}
\label{sec:Machine Translation}
Machine Translation (MT) refers to the process of automatically translating text from a source language to a target language. Over time, MT technology has progressed significantly, evolving from rule-based methods to Statistical Machine Translation (SMT), and more recently to Neural Machine Translation (NMT), which is now the dominant approach. With the widespread adoption of NMT and the emergence of LLMs, evaluating translation quality has become a complex task, requiring robust evaluation frameworks that can assess accuracy, fluency, and contextual relevance across diverse language pairs.

The Workshop on Machine Translation (WMT)~\cite{freitag2021results} is a prominent annual evaluation event in the field of MT. It provides large-scale, human-annotated datasets for a variety of language pairs, including English-French, English-German, and English-Russian. Each year, WMT releases benchmark datasets that include source texts, model-generated translations, reference translations, and human evaluation scores. These datasets are widely used for assessing the performance of automated evaluation metrics by comparing their outputs against human judgments. WMT covers a broad range of tasks, from sentence-level translation to document-level and domain-specific challenges, making it a comprehensive resource for evaluating the correlation between automated evaluators and human assessments. However, WMT primarily focuses on high-resource languages, which may limit its applicability to low-resource or underrepresented languages.
Literary Translation Comparisons~\cite{karpinska2023large} is designed to assess document-level translation quality, particularly in the context of literary works. It includes carefully selected paragraphs from various literary pieces, covering 18 language pairs such as Japanese-English, Polish-English, and French-English. Unlike sentence-level benchmarks, this dataset emphasizes the importance of evaluating translations in a broader context, as literary texts often require understanding of stylistic elements and cultural subtleties. This makes it particularly useful for evaluating the performance of LLMs, which may excel in capturing broader contextual information.
The MQM~\cite{freitag2021experts} study is the largest evaluation effort to date focusing on machine translation quality. It involves professional translators annotating the outputs of top-performing systems from the WMT 2020 shared task, specifically targeting English-German and Chinese-English translations. MQM introduces a multidimensional quality assessment framework that goes beyond traditional metrics like BLEU or ROUGE. It evaluates translations across multiple dimensions, including accuracy, fluency, terminology, style, and locale, providing a more nuanced understanding of translation quality.

\subsubsection{Text Summarization}
\label{sec:Text Summarization}
Text Summarization (TS) is the task of generating a concise and coherent summary from a given piece of text while preserving its essential meaning. The main goal is to provide a quick, accurate overview of the source content, capturing key information and eliminating unnecessary details. As LLMs have shown impressive capabilities in generating summaries, the need for robust meta-evaluation benchmarks is critical to effectively assess their performance across various dimensions like coherence, relevance, consistency, and fluency.

SummEval~\cite{fabbri2021summeval} is one of the most widely used benchmarks for evaluating summarization models. It includes summaries generated by 16 different models based on 100 news articles randomly sampled from the CNN/DailyMail test set. Each summary was annotated by five independent crowd-sourced workers and three expert evaluators, using a Likert scale from 1 to 5 across four key dimensions: coherence, consistency, fluency, and relevance. The dataset is valuable for analyzing the correlation between human judgments and automated evaluation metrics.
The FRANK~\cite{pagnoni2021understanding} dataset is dedicated to assessing the factual accuracy of summaries generated by automatic summarization systems. It provides detailed human annotations of factual errors, including semantic frame errors, discourse errors, and content verifiability issues. The dataset includes summaries from both the CNN/DailyMail and XSum datasets, making it a comprehensive resource for evaluating factual correctness. FRANK’s detailed categorization of errors offers valuable insights into the types of factual inaccuracies common in generated summaries, highlighting areas where LLMs often struggle. However, focusing solely on factual errors may overlook other aspects of summary quality, such as coherence and fluency.
OpinsummEval~\cite{shen2023opinsummeval} is a meta-evaluation benchmark specifically designed for opinion summarization tasks, where the goal is to extract and summarize opinions from a large volume of user reviews. This dataset includes outputs from 14 different opinion summarization models and provides human annotations across four dimensions: aspect relevance, self-consistency, sentiment consistency, and readability.

\subsubsection{Dialogue Generation}
\label{sec:Dialogue Generation}
Dialogue Generation is the task of automatically generating natural language conversations that are relevant to a given context. The primary goal is to develop dialogue systems that can understand context, generate fluent responses, and maintain logical consistency and contextual accuracy. Dialogue generation encompasses a wide range of applications, from chatbots and virtual assistants to social conversational agents. With the increasing capabilities of large language models (LLMs), evaluating dialogue generation has become more complex, requiring multi-faceted evaluation frameworks to assess various aspects of conversational quality.

In the field of dialogue generation, the most commonly used datasets include Topical-Chat~\cite{gopalakrishnan2023topical} and PERSONA-CHAT~\cite{zhang2018personalizing}. The Topical-Chat~\cite{gopalakrishnan2023topical} dataset aims to advance research in open-domain conversational AI, covering eight major topics such as entertainment, health, and technology. The PERSONA-CHAT~\cite{zhang2018personalizing} dataset, on the other hand, focuses on enhancing dialogue systems by incorporating predefined personas to generate more personalized responses. Each dialogue participant is assigned a persona profile, consisting of several descriptive sentences about their personality or preferences.
Mehri and Eskenazi~\cite{mehri2020usr} conducted a meta-evaluation study on these two widely-used open-domain dialogue corpora. They manually annotated 60 dialogue contexts from each dataset, with six responses per context for Topical-Chat and five for PERSONA-CHAT~\cite{zhang2018personalizing}, including both model-generated and human responses. Each response was evaluated across six key dimensions: naturalness, coherence, engagement, groundedness, understandability, and overall quality. This study highlights the importance of multi-dimensional evaluation in dialogue generation, providing valuable insights into the strengths and weaknesses of different dialogue models.
Additionally, the dataset from DSTC10 Track 5~\cite{yoshino2023overview,zhang2021automatic} focuses on evaluating open-domain dialogue systems and is designed for automatic evaluation and moderation of dialogue systems. The challenge aims to develop automatic evaluation mechanisms that accurately reflect human judgments while effectively handling harmful user inputs, maintaining conversational flow and engagement. The dataset includes annotations across four aspects: coherence, appropriateness, naturalness, and toxicity control.

\subsubsection{Automatic Story Generation}
\label{sec:Automatic Story Generation}
Automatic Story Generation (ASG) is a challenging task that aims to enable models to create coherent, engaging narratives based on a given prompt or context. It emulates human storytelling abilities by generating stories that exhibit a logical structure, compelling characters, and interesting plot developments. Evaluating story generation systems is inherently complex, as it involves assessing not only linguistic quality but also narrative elements like coherence, engagement, and surprise.

The HANNA~\cite{chhun2022human} dataset is tailored for evaluating automatic story generation (ASG), featuring 1,056 stories generated by 10 different systems from 96 prompts. Each story is annotated by three human reviewers across six criteria: relevance, coherence, resonance, surprise, engagement, and complexity. This comprehensive annotation framework provides a detailed assessment of narrative quality, making HANNA a valuable benchmark for comparing ASG models.
Another notable dataset is the MANS~\cite{guan2021openmeva}, which forms part of the OpenMEVA~\cite{guan2021openmeva} framework. It compiles stories from various natural language generation models using well-known corpora like ROCStories~\cite{mostafazadeh2016corpus} and WritingPrompts~\cite{fan2018hierarchical}. MANS~\cite{guan2021openmeva} focuses on manual annotations of narrative elements, serving as a robust testbed for exploring diverse evaluation metrics.
The StoryER~\cite{chen2023storyer} dataset offers a distinct approach to evaluating story generation by focusing on preference prediction and aspect-based rating. StoryER is divided into two primary components: the first is a 100k Story Ranking Data, which pairs stories from the WritingPrompts dataset. Each pair includes one story with high user engagement (upvotes $\ge$ 50) and another with low engagement (upvotes $\le$ 0). This component leverages real-world user feedback to capture implicit preferences, providing a practical basis for training models to predict story quality. The second component, Aspect Rating and Reasoning Data, contains 46,000 entries where annotators provide detailed ratings (on a scale of 1-5) for various story aspects such as introduction, character development, and plot description, along with explanatory comments. This combination of quantitative rankings and qualitative reasoning enables a nuanced evaluation of stories, making StoryER particularly useful for both automated scoring and interpretability research.
The PERSER~\cite{wang2023learning} dataset takes a different approach by addressing the subjectivity inherent in open-domain text generation evaluations. PERSER restructures existing datasets and introduces personalized tags, resulting in two sub-datasets: Per-MPST and Per-DOC. Per-MPST is an adapted version of the Movie Plot Synopsis Dataset, while Per-DOC includes 7,000 instances of paired stories generated from the same premise. These stories are evaluated based on dimensions such as interestingness, adaptability, surprise, character development, and the quality of the ending.

\subsubsection{Values Alignment}
\label{sec:Values Alignment}
Values alignment is a critical task in the development of AI systems, focused on ensuring that their behavior and decisions consistently reflect core human values and ethical standards. In the context of LLM-as-Judge, the alignment process is vital to verify that the model's outputs adhere to societal norms and ethical principles, minimizing risks related to harmful, biased, or unethical behavior. To support research and model development in values alignment, several datasets have been created, each with unique characteristics designed to evaluate or enhance the ethical behavior of LLMs.

One notable dataset is PKU-SafeRLHF~\cite{ji2024pku}, which was specifically curated for studying safe alignment in large language models. The dataset comprises 83.4K preference entries, focusing on two primary dimensions: harmlessness and usefulness. In each sample, the dataset presents a pair of model responses to a given prompt, annotated with safety meta-labels and preferences based on the levels of safety and utility. 
Another influential dataset is the HHH~\cite{askell2021general} (Honesty, Helpfulness, and Harmlessness) dataset, designed to evaluate LLM performance across various human-model interaction scenarios. The dataset emphasizes three core human-centered values: honesty, helpfulness, and harmlessness. It includes a diverse collection of conversational examples where models are tested on their adherence to these values. By exposing models to a wide range of contexts, the HHH dataset serves as a comprehensive benchmark for assessing whether LLMs align with essential ethical standards and effectively mitigate risks of misinformation, harmful advice, or biased outputs.
Moreover, the CVALUES~\cite{xu2023cvalues} benchmark is a more recent contribution aimed at evaluating human values alignment specifically for Chinese LLMs. It represents the first comprehensive framework tailored to assess values alignment in the Chinese language context, focusing on two critical criteria: Safety and Responsibility.

\subsubsection{Recommendation}
\label{sec:Recommendation}
Recommendation systems aim to provide personalized suggestions based on users' preferences and historical behavior. As the use of large language models (LLMs) expands, their role in evaluating the performance of recommendation systems has garnered increasing attention. LLMs can serve as versatile evaluators, offering insights into multiple aspects of recommendation systems beyond traditional metrics like accuracy. They can assess factors such as user engagement, satisfaction, and the quality of generated explanations.

The MovieLens~\cite{harper2015movielens} dataset is a widely-used public dataset for movie recommendations, available in multiple versions with varying scales, ranging from thousands of users and ratings to millions. Zhang et al.~\cite{zhang2024large} further annotated the MovieLens~\cite{harper2015movielens} data to create a sub-dataset featuring user self-explanation texts. In this sub-dataset, users write explanatory texts after being presented with a recommended movie. These explanations are then rated on a five-point Likert scale across four dimensions: Persuasiveness, Transparency, Accuracy, and Satisfaction. This annotated data provides valuable reference texts for LLMs in the context of explainability evaluation.

Another commonly used dataset is the Yelp dataset~\cite{asghar2016yelp}, which contains detailed review data from 11 metropolitan areas, covering approximately 150,000 businesses, nearly 7 million user reviews, and over 200,000 images. User reviews include ratings for businesses, such as hotel ratings (1-5 stars), as well as additional feedback like ``cool'' and ``funny'' votes. Furthermore, the Yelp dataset provides extensive business attribute information (e.g., operating hours, parking availability, and delivery options), offering rich contextual information that can be leveraged for developing and evaluating recommendation systems.

\subsubsection{Search}
\label{sec:Search}
The search task is a fundamental component of information retrieval (IR), focusing on identifying the most relevant documents from extensive text collections based on user queries. Traditionally, relevance assessments in search tasks have been conducted by human annotators following established guidelines. However, recent advances in large language models (LLMs) have opened up new opportunities for utilizing these models as evaluators, offering an automated and scalable approach to relevance assessment.
With the advent of retrieval-augmented generation (RAG) models, the role of LLMs as evaluators has expanded. There is now a growing need to assess various dimensions of retrieved contexts, including context relevance, answer faithfulness, and answer relevance. This shift highlights the potential of LLMs to provide nuanced judgments that go beyond simple topical relevance.

A key resource for evaluating the performance of LLMs as relevance assessors is the series of datasets from the Text Retrieval Conference (TREC). TREC workshops aim to advance research in IR by offering large-scale test collections, standardized evaluation procedures, and a platform for benchmarking retrieval models. 
The datasets from the TREC Deep Learning Track~\cite{lawrie2024overview}, specifically from 2021 (DL21)~\cite{DBLP:conf/trec/Craswell0YCL21} and 2022 (DL22)~\cite{DBLP:conf/trec/Craswell0YCLVS22}, are commonly used for this purpose. These datasets are derived from the expanded MS MARCO v2 collection~\cite{bajaj2016ms}, which contains approximately 138 million passages. Relevance judgments are provided by assessors from the National Institute of Standards and Technology (NIST) using a 4-point scale (0 to 3). This structured and fine-grained annotation scheme allows for a detailed comparison between LLM-generated relevance scores and human judgments.
While general-purpose datasets offer valuable benchmarks, specialized retrieval tasks often require domain-specific datasets that reflect unique relevance criteria. One notable example is LeCaRDv2~\cite{li2024lecardv2}, a large-scale dataset tailored for legal case retrieval. LeCaRDv2 enriches the concept of relevance by incorporating three distinct aspects: characterization, penalty, and procedure. These additional criteria provide a more comprehensive perspective on relevance.

\subsubsection{Comprehensive Data}
\label{sec:Comprehensive Data}
To thoroughly assess the role of LLMs-as-Judges and better align them with human preferences, a diverse set of comprehensive datasets has been developed. These datasets provide large-scale, well-annotated data, allowing for the effective training and evaluation of LLMs in complex, real-world contexts. As a result, they contribute to improving the models’ reliability and effectiveness in their role as evaluators.

Datasets such as HelpSteer~\cite{wang2023helpsteer} and HelpSteer2~\cite{wang2024helpsteer2} are designed to improve the alignment and usefulness of LLMs. They provide multi-attribute data, enabling the training of models that can generate responses that are factually correct, coherent, and tailored to diverse user preferences. These open-source datasets support adjustments in response complexity and verbosity, catering to varying user needs. Additionally, UltraFeedback~\cite{cui2024ultrafeedback} offers a large-scale dataset with around 64,000 prompts from sources like UltraChat~\cite{ding2023enhancing}, ShareGPT~\cite{chiang2023vicuna}, and TruthfulQA~\cite{lin2021truthfulqa}. It includes multiple responses per prompt generated by different LLMs, with high-quality preference labels and textual feedback covering aspects like instruction-following, truthfulness, and helpfulness. UltraFeedback’s fine-grained annotations and diverse prompts provide a robust resource for training reward and critic models, enhancing the evaluative capabilities of LLMs.

In exploring instruction following and dialogue capabilities, specialized tools like AlpacaEval~\cite{dubois2024length}, alongside interactive platforms such as Chatbot Arena~\cite{zheng2023judging} and benchmarks like MT-Bench~\cite{zheng2023judging}, provide critical insights. AlpacaEval is an automated evaluation tool using GPT-4 or Claude as evaluators. It assesses chat-based LLMs against the AlpacaFarm dataset, providing win-rate calculations across a variety of tasks, enabling rapid and cost-effective comparisons with baseline models like GPT-3.5 (Davinci-003). Chatbot Arena, on the other hand, offers a user-driven evaluation framework where participants interact with anonymous models and vote based on their preferences. The platform has collected over 1,000,000 user votes, using the Bradley-Terry model to rank LLMs and chatbots, providing valuable insights into user preferences and model performance in open-domain dialogue.

Benchmarks like WildBench~\cite{lin2024wildbench} and FLASK~\cite{ye2023flask} aim to evaluate LLMs on tasks more reflective of real-world applications. WildBench~\cite{lin2024wildbench} collects challenging examples from real users via the AI2 WildChat project, providing fine-grained annotations, task types, and checklists for response quality evaluation, and employs length-penalized Elo ratings to ensure unbiased assessments. FLASK~\cite{ye2023flask} introduces a fine-grained evaluation protocol that decomposes overall scoring into skill set-level scoring for each instruction, enhancing interpretability and reliability in both human-based and model-based evaluations. Additionally, comprehensive evaluations covering multiple domains—including factual question answering, reading comprehension, summarization, mathematical problem-solving, reasoning, poetry generation, and programming—have been conducted. These evaluations involve assessing models across multiple criteria such as correctness, fluency, informativeness, logicality, and harmlessness.

Reward models and LLM-based judges face the crucial task of ensuring alignment with human expectations, a challenge addressed by datasets like RewardBench~\cite{lambert2024rewardbench}, RM-Bench~\cite{liu2024rm}, and JudgerBench~\cite{cao2024compassjudger}.
RewardBench~\cite{lambert2024rewardbench} focuses on assessing models through complex prompt-choice trios, covering diverse areas like chat, reasoning, and safety, with a particular emphasis on out-of-distribution scenarios.  RM-Bench~\cite{liu2024rm} introduces a new benchmark for evaluating reward models based on their sensitivity to subtle content differences and resistance to stylistic biases, emphasizing the need for refined assessments that correlate highly with aligned language models' performance.
JudgerBench~\cite{cao2024compassjudger}, with its dual components (JDB-A and JDB-B), offers a structured framework for evaluating alignment and critique abilities. By including data from human voting results and combining insights from varied sources, JudgerBench~\cite{cao2024compassjudger} provides a nuanced understanding of model performance across different languages and dialogue formats.

With the growing complexity of tasks handled by LLMs, there is an increasing demand for more objective and reliable evaluation frameworks. JUDGEBENCH~\cite{tan2024judgebench} proposes a novel approach to assessing LLM-based judges on challenging response pairs across domains like knowledge, reasoning, mathematics, and coding. It addresses the limitations of existing benchmarks by introducing preference labels that reflect objective correctness, providing a robust platform for evaluating the capabilities of advanced LLM-based judges.

As LLMs evolve beyond text-only tasks, evaluation frameworks have expanded to encompass multimodal and multilingual contexts. MLLM-as-a-Judge~\cite{chen2024mllm} serves as a benchmark for assessing Multimodal LLMs, covering tasks like image description, mathematical reasoning, and infographic interpretation. By integrating human annotations, it provides a comprehensive evaluation across visual and textual domains, reflecting the growing demand for models capable of processing diverse inputs. In a parallel effort, MM-Eval~\cite{son2024mm} addresses the multilingual aspect, offering extensive analysis across 18 languages. With core subsets like Chat, Reasoning, and Linguistics, alongside a broader Language Resource subset spanning 122 languages, MM-Eval~\cite{son2024mm} highlights performance discrepancies, especially in low-resource languages where models tend to default to neutral scores.

\subsection{Metric}
\label{sec:metric}
The evaluation of LLMs-as-Judges models centers around assessing the extent to which the model's judgments align with human evaluations, which are typically considered the benchmark for quality. Given the complexity and subjectivity of many evaluation tasks, achieving high agreement with human ratings is a key indicator of the LLM's performance. To quantify this agreement, a range of statistical metrics is employed. Below, we outline these metrics and their applications in evaluating LLMs-as-Judges models.

\subsubsection{Accuracy}

Accuracy is a fundamental metric used to assess the proportion of correct judgments made by the LLM compared to human evaluations. In classification tasks, it is defined as:

\begin{equation}
    \text{Accuracy} = \frac{\text{Number of Correct Predictions}}{\text{Total Number of Predictions}},
\end{equation}

where the number of correct predictions corresponds to instances where the LLM's judgment matches the human evaluator's judgment. While accuracy is simple to compute and intuitive, it may not fully capture the quality of the model, especially when dealing with tasks that involve nuanced or continuous evaluations.

\subsubsection{Pearson Correlation Coefficient}

The Pearson Correlation Coefficient~\cite{cohen2009pearson} measures the linear relationship between two continuous variables, in this case, the evaluation scores assigned by the LLM and those assigned by human evaluators. It is defined as:

\begin{equation}
    r = \frac{\sum (x_i - \bar{x})(y_i - \bar{y})}{\sqrt{\sum (x_i - \bar{x})^2 \sum (y_i - \bar{y})^2}},
\end{equation}

where $x_i$ and $y_i$ are the scores from the LLM and the human, respectively, and $\bar{x}$ and $\bar{y}$ are their means. Pearson correlation values range from $-1$ to $1$.

\subsubsection{Spearman's Rank Correlation Coefficient}

Spearman's Rank Correlation Coefficient ($\rho$) ~\cite{sedgwick2014spearman} assesses the monotonic relationship between two variables by comparing their ranked values rather than the raw scores. It is defined as:

\begin{equation}
    \rho = 1 - \frac{6 \sum d_i^2}{n(n^2 - 1)},
\end{equation}

where $d_i$ is the difference between the ranks of corresponding scores from the LLM and the human evaluator, and $n$ is the number of paired scores.
Spearman's $\rho$ is less sensitive to outliers and non-linear relationships compared to Pearson's correlation, making it a robust choice for evaluating tasks where the relative order of scores is more important than the exact values. It is commonly used in ranking-based evaluations such as preference judgments or ranking tasks.

\subsubsection{Kendall's Tau}

Kendall's Tau ($\tau$) ~\cite{sen1968estimates} is another rank-based correlation metric that measures the ordinal association between two ranked lists. It is defined as:

\begin{equation}
    \tau = \frac{C - D}{\frac{1}{2}n(n-1)},
\end{equation}

where $C$ is the number of concordant pairs (where the rank order agrees between the LLM and human), and $D$ is the number of discordant pairs. Kendall's $\tau$ is particularly useful when evaluating the consistency of rankings produced by LLMs and human evaluators. It is often preferred when the dataset contains many ties, as it provides a more nuanced measure of agreement than Spearman's $\rho$.

\subsubsection{Cohen's Kappa}

Cohen's Kappa ($\kappa$) ~\cite{warrens2015five} measures the level of agreement between two raters (in this case, the LLM and the human) beyond what would be expected by chance. It is defined as:

\begin{equation}
    \kappa = \frac{p_o - p_e}{1 - p_e},
\end{equation}

where $p_o$ is the observed agreement and $p_e$ is the expected agreement by chance.
Cohen's Kappa is particularly effective in classification tasks where both the LLM and the human evaluators assign categorical labels. It accounts for the possibility of random agreement, making it a more robust metric than simple accuracy.

\subsubsection{Intraclass Correlation Coefficient (ICC)}

The Intraclass Correlation Coefficient (ICC)~\cite{bartko1966intraclass} assesses the reliability of ratings when there are multiple evaluators. It evaluates the consistency or conformity of measurements made by different raters, including LLMs and human annotators. ICC is defined based on the variance components derived from a one-way or two-way ANOVA model.
The ICC is particularly useful when comparing multiple LLMs or when evaluating the consistency of an LLM across different subsets of data, providing a broader view of its reliability as an evaluator.

\begin{table*}[t]
    \centering
    \small
    \caption{Summary of Common Metrics for Evaluating LLMs-as-Judges Models}
    \begin{tabular}{l|l|l|l}
        \hline
        \textbf{Metric} & \textbf{Type} & \textbf{Use Case} & \textbf{Robustness to Outliers} \\
        \hline
        \hline
        Accuracy & Agreement measure & Proportion of correct judgments & Sensitive \\
        \hline
        Pearson & Linear correlation & Continuous score comparison & Sensitive \\
        \hline
        Spearman & Rank correlation & Rank-based evaluation tasks & Robust \\
        \hline
        Kendall's Tau & Rank correlation & Consistency in ordinal rankings & Robust, handles ties \\
        \hline
        Cohen's Kappa & Agreement measure & Two raters, consistency analysis & Adjusts for chance \\
        \hline
        ICC & Agreement measure & Multiple raters, consistency analysis & Robust for group ratings \\
        \hline
    \end{tabular}
    
\end{table*}

\section{Limitation}
\label{sec:Limitation}
Although the application of LLMs-as-judges holds great promise, there are still several significant limitations that can affect their effectiveness, reliability, and fairness~\cite{thakur2024judging,stureborg2024large}.
These limitations arise from the inherent characteristics of LLMs, including their reliance on large-scale data for training and token-based decoding mechanisms.
In this section, we will primarily explore the limitations in the following three key aspets: \textbf{Biases} (\S\ref{sec:biases}), \textbf{Adversarial Attacks} (\S\ref{sec:attacks}), and \textbf{Inherent Weaknesses} (\S\ref{sec:weaknesses}).

\begin{table*}[t]
\caption{Overview of different biases.}
\begin{tabular}{ll}
\hline
\textbf{Bias}                   & \textbf{Description}                                                                                                                                                                                                        \\ \hline \hline
\multicolumn{2}{l}{\textbf{Presentation-Related Biases} (\S\ref{sec:presentationbiases})}                                                                                                                                                                                             \\ \hline 
\rowcolor{mygray}
Position Bias          & \begin{tabular}[c]{@{\ }p{0.7\textwidth}@{\ }}A tendency to make judgments based on the position of the input, where responses earlier or later in the  sequence are favored over those in other positions.\end{tabular}         \\
Verbosity Bias         & \begin{tabular}[c]{@{\ }p{0.7\textwidth}@{\ }}A tendency to favor longer responses, potentially equating length with quality, regardless of the content’s actual value.\end{tabular}                                               \\ \hline \hline
\multicolumn{2}{l}{\textbf{Social-Related Biases} (\S\ref{sec:socialbiases})}                                                                                                                                                                                                   \\ \hline
\rowcolor{mygray}
Authority Bias         & \begin{tabular}[c]{@{\ }p{0.7\textwidth}@{\ }}A tendency to be swayed by references to authoritative sources, such as books, websites, or famous individuals.\end{tabular}                                                         \\
Bandwagon-Effect Bias  & \begin{tabular}[c]{@{\ }p{0.7\textwidth}@{\ }}A tendency to align with majority opinions, where LLMs-as-judges favor prevailing views over objectively assessing the content.\end{tabular}                                      \\
\rowcolor{mygray}
Compassion-Fade Bias   & \begin{tabular}[c]{@{\ }p{0.7\textwidth}@{\ }}A tendency to be influenced by anonymization strategies, such as the removal of model names, affecting the judgments of LLMs.\end{tabular}                                        \\
Diversity Bias         & \begin{tabular}[c]{@{\ }p{0.7\textwidth}@{\ }}A tendency to shift judgments based on identity-related markers, such as gender, ethnicity, or other social categorizations.\end{tabular}                                            \\ \hline \hline
\multicolumn{2}{l}{\textbf{Content-Related Biases} (\S\ref{sec:contentbiases})}                                                                                                                                                                                                  \\ \hline
\rowcolor{mygray}
Sentiment Bias         & \begin{tabular}[c]{@{\ }p{0.7\textwidth}@{\ }}A tendency to favor responses with specific emotional tones, such as cheerful or neutral, over negative or fearful ones.\end{tabular}                                                \\
Token Bias             & \begin{tabular}[c]{@{\ }p{0.7\textwidth}@{\ }}A tendency of LLMs to favor more frequent or prominent tokens during pre-training, leading to skewed judgments.\end{tabular}                                                         \\
\rowcolor{mygray}
Context Bias           & \begin{tabular}[c]{@{\ }p{0.7\textwidth}@{\ }}A tendency of LLMs to produce biased judgments influenced by contextual examples or cultural contexts, potentially leading to biased or culturally insensitive outcomes.\end{tabular} \\ \hline \hline 
\multicolumn{2}{l}{\textbf{Cognitive-Related Biases} (\S\ref{sec:cognitivebiases})}                                                                                                                                                                                                \\ \hline
\rowcolor{mygray}
Overconfidence Bias    & \begin{tabular}[c]{@{\ }p{0.7\textwidth}@{\ }}A tendency to exhibit inflated confidence in evaluation judgments, leading to overly assertive but potentially incorrect conclusions\end{tabular}                                 \\
Self-Enhancement Bias  & \begin{tabular}[c]{@{\ }p{0.7\textwidth}@{\ }}A tendency to favor outputs generated by the same model acting as a judge, undermining objectivity.\end{tabular}                                                                     \\
\rowcolor{mygray}
Refinement-Aware Bias  & \begin{tabular}[c]{@{\ }p{0.7\textwidth}@{\ }}A tendency for scoring variations influenced by whether an answer is original, refined, or accompanied by conversation history during evaluation.\end{tabular}                    \\
Distraction Bias       & \begin{tabular}[c]{@{\ }p{0.7\textwidth}@{\ }}A tendency to be influenced by irrelevant content, which can detract from the quality of judgments by diverting attention from critical elements.\end{tabular}                    \\
\rowcolor{mygray}
Fallacy-Oversight Bias & \begin{tabular}[c]{@{\ }p{0.7\textwidth}@{\ }}A tendency to overlook logical fallacies, which can undermine the accuracy of judgments.\end{tabular}                                                                                \\ \hline
\end{tabular}
\label{tab:biases}
\end{table*}

\subsection{Biases}
\label{sec:biases}
Essentially, LLMs are trained on vast amounts of data gathered from diverse sources. While this allows them to generate human-like responses, it also makes them inherit to the biases inherent in the training data. 
These biases are presented in various forms, which can significantly affect evaluation results, compromising the fairness and accuracy of decisions.

To gain a deeper understanding of the impact of bias, we have provided a detailed classification of bias. As shown in Table~\ref{tab:biases}, the biases exhibited by LLMs-as-judges can be systematically categorized into four groups based on their underlying causes and manifestations: \textbf{Presentation-Related Biases} (\S\ref{sec:presentationbiases}), \textbf{Social-Related Biases} (\S\ref{sec:socialbiases}), \textbf{Content-Related Biases} (\S\ref{sec:contentbiases}), and \textbf{Cognitive-Related Biases} (\S\ref{sec:cognitivebiases}). In this section, we provide a detailed overview of the definition, impact, and solutions to these biases.

\subsubsection{Presentation-Related Biases}  
\label{sec:presentationbiases}
Presentation-Related Biases refer to tendencies in LLMs where judgments are influenced more by the structure or presentation of information than by its substantive content. 
For example, models may prioritize certain formats, styles, or patterns of expression, which can affect the quality of the input. Next, we introduce two biases related to Presentation-Related Biases: position bias and verbosity bias.

\noindent\textbf{Position bias} is a prevalent issue not only in the context of LLMs-as-judges but also in human decision-making and across various machine learning domains.
Research has shown that humans are often influenced by the order of options presented to them, leading to biased decision that can impact fairness and objectivity~\cite{shi2024judging,blunch1984position,raghubir2006center,zhao2024measuring}. 
Similarly, in other ML applications, models trained on ordered data exhibit a positional preference, skewing outcomes based on the sequence of input~\cite{ko2020look,wang2018position}. 
Position bias in LLMs-as-judges refers to the tendency of LLMs to favor certain answers based on their position in the response set. 
For example, when presented with multiple answer choices or compared pairwise, LLMs disproportionately select options that appear earlier in the list, leading to skewed judgment. 

Recent studies have further examined position bias in the LLMs-as-judges context. 
For instance, a framework~\cite{llmsjuding2024openreview} is proposed to investigate position bias in pairwise comparisons, introducing metrics such as repetition stability, position consistency, and preference fairness to better understand how positions affect LLM judgments. 
Another study~\cite{zheng2023judging} explores the limitations of LLMs-as-judges, including position biases, and verifies agreement between LLM judgments and human preferences across multiple benchmarks. 
These findings underscore the need for robust debiasing strategies to enhance the fairness and reliableness of LLMs-as-judges.

Several methods are proposed to mitigate position bias. 
The naive approach involves excluding inconsistent judgments by swapping the positions of the candidate answers and verifying whether the LLM's judgment remains consistent. Inconsistent judgments are then filtered out~\cite{zheng2023judging,chen2024humans,wang2023large,li2023generative,zheng2023large,li2024calibraeval}.
The \textbf{swap-based} debiasing method can be further divided into two categories: \textit{score-based} and \textit{comparison-based}. Both approaches start by swapping the positions of the candidate answers. The difference lies in how the final judgment is determined. In the score-based method, each candidate answer is scored, and the average score across multiple swaps is taken as the final score for that answer~\cite{zheng2023judging,wang2023large,li2023generative,raina2024llm,hou2024large}.In contrast, the comparison-based method considers the outcome a tie if the LLM's judgments are inconsistent after swapping.
The conclusion of a tie is based on an analysis of the quality gap between answers. The larger the quality gap between candidate answers, the smaller the impact of position bias, resulting in higher consistency in predictions after swapping their positions, which is detailed in a recently study~\cite{llmsjuding2024openreview}.
In addition to the aforementioned methods, PORTIA~\cite{li2023split} employs an \textbf{alignment-based} approach that simulates human comparison strategies. It divides each answer into multiple segments, aligns similar content across candidate answers, and then merges these aligned segments into a single prompt for the LLM to evaluate. By presenting content in a balanced and aligned format, PORTIA enables the model to make more consistent and unbiased judgments, focusing on answer quality rather than order. This approach is effective across various LLMs, significantly improving evaluation consistency and reducing costs.
Further efforts to enhance LLM-based evaluations have explored new techniques to address position bias and other judgment inconsistencies. \textbf{Discussion-based} methods~\cite{li2023prd,khan2024debating} incorporate peer ranking and discussion to improve evaluation accuracy. Instead of relying solely on a single LLM’s judgment, these methods prompt multiple LLMs to compare answers and discuss preferences to reach a consensus, thereby reducing individual positional bias and enhancing alignment with human judgments. This collaborative evaluation approach represents a promising direction for mitigating biases inherent in LLM assessments.

\noindent\textbf{Verbosity bias}~\cite{khan2024debating,chen2024humans,zheng2023judging,nasrabadi2024juree} refers to the tendency of a judge, whether human or model-based, to favor lengthier responses over shorter ones, irrespective of the actual content quality or relevance. This bias may cause LLMs prefer longer responses, even if the extended content does not contribute substantively to the correctness of the judgments. 

To mitigate verbosity bias in LLMs-as-judges, several approaches~\cite{khan2024debating,ye2024justice,ye2024beyond} have been proposed. One approach~\cite{khan2024debating} employs persuasive debating techniques, structuring responses to prioritize substance. By training LLMs-as-judges to engage in a debate-like format, this method encourages clarity and relevance, reducing the tendency to favor verbose arguments that lack substantive content. Additionally, the CALM~\cite{ye2024justice} framework introduces controlled modifications to systematically assess and quantify verbosity’s impact on judgments, using automated perturbations to evaluate robustness against verbosity bias and refine LLMs-as-judges toward objective and concise assessments. Complementing these methods, the contrastive judgments (Con-J)~\cite{ye2024beyond} approach trains models with structured rationale pairs instead of scalar scores, encouraging LLMs-as-judges to focus on well-reasoned content rather than associating verbosity with quality.

\subsubsection{Social-Related Biases} 
\label{sec:socialbiases}
Social-Related Biases refer to biases in language models that resemble social phenomena~\cite{zhao2023mind}. These biases may manifest when models are swayed by references to authoritative sources (Authority Bias), align with prevailing majority opinions without independent evaluation (Bandwagon-Effect Bias), or adjust their judgments based on anonymization strategies or identity markers such as gender and ethnicity (Compassion-Fade Bias and Diversity Bias). Next, we present the details of these biases.

\noindent\textbf{Authority bias}~\cite{chen2024humans,ye2024justice} in the context of LLMs-as-judges refers to the tendency of the model to attribute greater credibility to statements associated with authoritative references, regardless of the actual evidence supporting them. For instance, LLMs-as-judges may favor responses that include references to well-known sources or experts, even when the content is inaccurate or irrelevant. This bias highlights a critical vulnerability where the appearance of authority can unduly influence judgment outcomes.

    While specific solutions to mitigate authority bias in LLMs-as-judges are still under active exploration, potential approaches include using retrieval-augmented generation (RAG) techniques to verify the validity of authoritative claims against external knowledge bases. This approach allows the model to cross-check referenced information and ensure its alignment with factual evidence. Another possible strategy is to design prompts that explicitly emphasize semantic accuracy and relevance over perceived authority. Further research is needed to validate these approaches and develop robust methods for addressing authority bias effectively in evaluative contexts.    

\noindent\textbf{Bandwagon-effect bias}~\cite{koo2023benchmarking,ye2024justice}  in the context of LLMs-as-judges refers to the tendency of the model to align its judgments with the majority opinion or prevailing trends, regardless of the actual quality or correctness of the evaluated content. For instance, when multiple responses are presented with indications of popular support or consensus, LLMs-as-judges may disproportionately favor these responses over alternatives, even when the consensus is flawed or biased. This bias reflects a susceptibility to groupthink dynamics, undermining the objectivity and fairness of the judgment process.

    Solutions to address bandwagon-effect bias include designing evaluation prompts that anonymize information about majority opinions, ensuring that judgments are based solely on the intrinsic quality of the responses rather than external indicators of popularity. Further exploration of debiasing strategies tailored to specific evaluative contexts is necessary to mitigate the impact of bandwagon-effect bias effectively.

\noindent\textbf{Compassion-fade bias}~\cite{koo2023benchmarking,ye2024justice} occurs when the anonymity of model names or the absence of identifiable contextual cues affects the judgments made by LLMs-as-judges. For example, anonymizing model names or using neutral identifiers may lead to shifts in evaluation outcomes. This bias highlights how the lack of personalized or contextual information can diminish the model’s sensitivity to equitable considerations.

    To mitigate compassion-fade bias, it is important to design evaluation prompts that standardize judgment criteria, ensuring that assessments remain consistent regardless of whether identifying details are present. Additionally, fairness-driven frameworks that explicitly address anonymization effects can further enhance the reliability of LLMs-as-judges.

\noindent\textbf{Diversity bias}~\cite{chen2024humans,ye2024justice} in the context of LLMs-as-judges refers to the model’s tendency to exhibit judgment shifts based on identity-related markers, such as gender, ethnicity, religion, or other social categorizations. For example, LLMs-as-judges might favor responses associated with certain demographic groups over others, leading to unfair or skewed judgments. This bias reflects the model’s susceptibility to implicit stereotypes or unequal treatment of diverse identities present in the training data.
Continued efforts to address this bias are crucial for ensuring fairness and inclusivity in the judgments conducted by LLMs-as-judges.

\subsubsection{Content-Related Biases}
\label{sec:contentbiases}
Content-Related Biases involve preferences or skewed judgments based on the content’s characteristics. An LLM might favor responses with certain emotional tones (Sentiment Bias), prefer frequently occurring words from its training data (Token Bias), or be influenced by specific cultural or domain contexts leading to insensitive outcomes (Context Bias). We present the details of these biases in the following.

\noindent\textbf{Sentiment bias}~\cite{ye2024justice} in the context of LLMs-as-judges refers to the tendency of the model to favor responses that exhibit certain emotional tones, such as positive or neutral sentiments, over others, regardless of their actual content quality or relevance. For instance, LLMs-as-judges may disproportionately reward responses that are cheerful or optimistic while penalizing those that are negative or emotionally intense, even if the latter are more contextually appropriate or accurate.

To address sentiment bias, potential solution is the use of sentiment-neutralizing mechanisms, such as filtering or adjusting responses to remove sentiment-driven influences during evaluation.

\noindent\textbf{Token Bias}~\cite{jiang2024peek,li2024calibraeval,pezeshkpour2023large,raina2024llm} refers to that LLMs favor certain tokens during the evaluation process. This bias often arises from the model's pre-training data, where more frequently occurring tokens are prioritized over less common ones, regardless of the contextual appropriateness or correctness in judgment.

\noindent\textbf{Contextual Bias} refers to the tendency of LLMs to produce skewed or biased judgments based on the specific context in which they are applied. For instance, models used in healthcare may propagate biases found in medical datasets, potentially influencing diagnoses or treatment recommendations~\cite{poulain2024bias}, while in finance, they might reflect biases in credit scoring or loan approval processes~\cite{zhou2024large}. 
In addition, the selection of contextual examples may also introduce bias~\cite{zhou2023batch,fei2023mitigating,zhao2021calibrate,han2022prototypical}.

\subsubsection{Cognitive-Related Biases} 
\label{sec:cognitivebiases}
Cognitive-Related Biases pertain to the inherent cognitive tendencies of LLMs in processing information. This includes exhibiting unwarranted confidence in judgments (Overconfidence Bias), favoring outputs generated by themselves (Self-Enhancement Bias), varying scores based on whether an answer is original or refined (Refinement-Aware Bias), being distracted by irrelevant information (Distraction Bias), or overlooking logical fallacies (Fallacy-Oversight Bias). The details of these biases are presented in the following.

\noindent\textbf{Overconfidence bias}~\cite{khan2024debating,jung2024trust} in the context of LLMs-as-judges refers to the tendency of models to exhibit an inflated level of confidence in their judgments, often resulting in overly assertive evaluations that may not accurately reflect the true reliability of the answer. This bias is particularly concerning in evaluative contexts, as it can lead LLMs-as-judges to overstate the correctness of certain outputs, compromising the objectivity and dependability of assessments.

To address Overconfidence bias, researchers have proposed several methods.
Cascaded Selective Evaluation~\cite{jung2024trust} addresses overconfidence by using Simulated Annotators to estimate confidence. This involves simulating diverse annotator preferences through in-context learning, which provides a more realistic measure of the likelihood that a human would agree with the LLM’s judgment. By analyzing multiple simulated responses, this method offers a confidence metric that reflects human-like disagreement, which helps to avoid overconfidence bias. 
Another method uses an adversarial debate mechanism, where two LLMs argue for different outcomes. Through structured debate rounds, each model is required to substantiate its position, which can reveal overconfidence by prompting self-reflection and critical analysis. This approach has been shown to improve truthfulness and reduces overconfidence by fostering a balanced evaluation, aligning LLMs' judgments more closely with accurate and reasoned conclusions.

\noindent\textbf{Self-enhancement bias} is the tendency to favor their own outputs~\cite{liu2023g,zheng2023judging,li2023prd,liu2023g,panickssery2024llm}. 
This concept of self-enhancement is drawn from social psychology, as discussed in Brown’s work in social cognition literature~\cite{brown1986evaluations}.
In the context of LLMs-as-judges, this bias manifests when a LLM evaluates its own generated outputs more favorably than those of other LLMs. Such bias is particularly concerning in applications involving self-assessment or feedback generation, as it compromises the objectivity of the LLMs-as-judges.

To address self-enhancement bias, PRD~\cite{li2023prd} introduces Peer Rank (PR) and Peer Discussion (PD) mechanisms. PR mitigates bias by using multiple LLMs as reviewers, each assessing pairwise comparisons between responses from different LLMs. By aggregating evaluations from several peer LLMs and weighting their preferences based on consistency with human judgments, PR reduces the impact of any single LLM’s self-enhancement bias, as more reliable reviewers have a greater influence. PD further alleviates self-enhancement bias by enabling two LLMs to engage in a dialogue to reach a mutual agreement on their preference between two answers. This multi-turn discussion encourages models to re-evaluate their initial judgments and consider alternative perspectives, focusing on content quality rather than self-generated responses. By promoting collaborative assessment and accountability, PR and PD effectively mitigate self-enhancement bias, aligning evaluations more closely with human standards.
Recently, an automated bias quantification framework named CALM~\cite{ye2024justice} has been proposed to systematically evaluate biases in LLMs-as-judges. CALM’s findings suggest that one effective way to reduce self-enhancement bias is to avoid using the same model to both generate and judge answers, thereby ensuring that evaluation remains more impartial.
Moreover, the Reference-Guided Verdict method~\cite{badshah2024reference} further addresses self-enhancement bias by providing a definitive gold-standard answer as a reference for LLM judges. This reference anchor helps align judgments to objective criteria, even when an LLM evaluates its own output, thus reducing the tendency to favor self-generated answers. Through structured prompts, this method has been shown to enhance reliability and mitigate variability in judgments, especially when multiple LLMs are used collectively. The integration of multiple LLMs, trained on varied datasets or fine-tuned with different parameters, has proven instrumental in producing less biased, more balanced evaluations, highlighting the effectiveness of model diversity and reference-guided criteria in combating self-enhancement bias.

\noindent\textbf{Refinement-aware bias}~\cite{ye2024justice} in the context of LLMs-as-judges refers to the tendency of the model to evaluate responses differently based on whether they are original, refined, or include revision history. For instance, an answer that has been iteratively refined may be judged more favorably than an original response, even if the refinement process does not significantly improve the content quality. Similarly, responses that explicitly present their improvement process or revision rationale might receive undue preference, skewing the evaluation outcomes.

While research on refinement-aware bias in the specific context of LLMs-as-judges remains limited, solution~\cite{xu2024pride} developed for general LLMs offer valuable insights. One potential solution involves incorporating external feedback mechanisms during judgment, as it introduces an objective and independent judgment mechanism that is not influenced by the LLM’s internal iterations or self-perception.

\noindent\textbf{Distraction bias} in the context of LLMs-as-judges refers to the model’s tendency to be influenced by irrelevant or unimportant details when making judgments. For instance, introducing unrelated information, such as a meaningless statement like “System Star likes to eat oranges and apples,”~\cite{ye2024justice,koo2023benchmarking,shi2023large} can significantly alter the model’s evaluation outcomes. This bias highlights the vulnerability of LLMs to attentional diversion caused by inconsequential content.

While existing studies~\cite{ye2024justice,koo2023benchmarking} have analyzed and discussed distraction bias, effective strategies to mitigate this issue in LLMs-as-judges remain underexplored. Potential solutions could involve input sanitization to preprocess and remove irrelevant information before presenting it to the model, ensuring that evaluations focus solely on relevant content. Additionally, explicit prompting with clear and strict guidelines could be designed to direct the LLM to evaluate only task-related aspects, reducing its susceptibility to distractions. Further research is needed to develop and validate robust methods that can systematically address distraction bias in the context of LLMs-as-judges.

\noindent\textbf{Fallacy-oversight bias}~\cite{chen2024humans,ye2024justice} refers to the tendency of LLMs-as-judges to overlook logical fallacies or inconsistencies within the evaluated responses. For instance, when presented with arguments or answers containing reasoning errors—such as circular reasoning, false dilemmas, or strawman arguments—LLMs-as-judges may fail to identify these issues and treat the responses as valid, potentially compromising the integrity of their evaluations.

In summary, while LLMs-as-judges have garnered significant attention for their effectiveness in diverse scenarios, the exploration of various biases that impact their performance remains relatively underdeveloped. 
% Various biases, such as bandwagon-effect bias, fallacy-oversight bias, and refinement-aware bias, have been identified, yet comprehensive studies and mitigation strategies specific to the LLMs-as-judges context are still lacking. 
These biases pose significant challenges to ensuring fair, objective, and reliable judgments across tasks, particularly in various applications where the implications of biased judgments can be severe.
Future research must focus on systematically identifying, quantifying, and addressing these biases within the LLMs-as-judges framework. 
Drawing from methodologies developed for general LLMs, such as external feedback mechanisms, balanced datasets, and fairness-aware prompting, could offer initial insights. However, domain-specific challenges require tailored solutions that align with the unique demands of LLMs-as-judges.

% Addressing these biases is critical to advancing the reliability and trustworthiness of LLMs-as-judges, paving the way for their broader adoption in practical.
\subsection{Adversarial Attacks}
\label{sec:attacks}
Adversarial attacks involve carefully crafted inputs designed to deceive the model into producing incorrect or unintended outputs. For LLM judges, attackers may subtly modify the input content, alter the wording of questions, or introduce misleading context to influence the model's evaluation results.
Researchers have found that for LLMs, even small, seemingly insignificant changes to the input data, such as adding or removing words, changing word order, or introducing ambiguous phrasing, can significantly affect the model's response~\cite{shen2023anything,jiang2023prompt,zou2023universal}. Such attacks can lead to inaccurate ratings or assessments, particularly when evaluating complex or high-risk tasks.

In this section, we first review research on adversarial attacks on LLMs within general domains. Then, we specifically focus on adversarial attacks in the context of LLMs-as-judges.

\subsubsection{Adversarial Attacks on LLMs}
\label{sec:Adversarial Attacks on LLMs}
Adversarial attacks on LLMs focus on exploiting vulnerabilities within the general framework of language model functionality. These attacks can be classified into three main categories based on the manipulation level: text-level manipulations, structural and semantic distortions, and optimization-based attacks.

\textbf{Text-Level Manipulations} involve subtle changes to the input text to deceive the model. Character-level perturbations, such as introducing typos, swapping letters, or inserting unnecessary characters, can cause significant changes in predictions despite minimal visible alterations~\cite{ebrahimi2017hotflip,jiang2023prompt}. Sentence-level modifications, such as rearranging phrases, adding irrelevant information, or paraphrasing inputs, further exploit the model’s sensitivity to surface-level changes~\cite{branch2022evaluating,perez2022ignore}.

\textbf{Structural and Semantic Distortions} focus on the syntactic and semantic properties of the input. Syntactic attacks rewrite sentence structures while preserving semantic meaning, targeting the model’s reliance on specific linguistic patterns~\cite{xu2023llm}. Semantic preservation with perturbations modifies critical tokens identified through saliency analysis, ensuring the attack minimally affects meaning but significantly alters predictions. 

\textbf{Optimization-Based Attacks} leverage algorithmic techniques to craft adversarial inputs. Gradient-based methods utilize the model’s gradients to identify and manipulate influential input features, causing substantial shifts in predictions~\cite{sun2020natural,sun2020adv}. Population-based optimization techniques iteratively generate adversarial examples in black-box settings, exploiting the model’s outputs to refine attacks~\cite{lee2022query}. 

These attacks highlight the vulnerabilities in LLMs, demonstrating their susceptibility to subtle manipulations. Studying these adversarial attacks is essential, as it provides insights that can guide the development of robust defense mechanisms, ensuring that LLMs maintain reliability against such manipulations.

\subsubsection{Adversarial Attacks on LLMs-as-judges}
\label{sec:Adversarial Attacks on LLMs-as-judges}
Recent studies have unveiled significant vulnerabilities in LLMs-as-judges to adversarial attacks~\cite{zheng2024cheating, doddapaneni2024finding, raina2024llm, shi2024optimization}. 
Zheng et al.~\cite{zheng2024cheating} and Doddapaneni et al.~\cite{doddapaneni2024finding} demonstrated that automatic benchmarking systems like MT-Bench~\cite{zheng2023judging} can be easily deceived to yield artificially high scores. These findings highlight that malicious inputs can manipulate evaluation metrics, undermining the reliability of such benchmarks.

Building on this, Raina et al.~\cite{raina2024llm} investigated the robustness of LLMs-as-judges against universal adversarial attacks. Their work showed that appending short, carefully crafted phrases to evaluated texts can effortlessly manipulate LLM scores, inflating them to their maximum regardless of the actual quality. Remarkably, these universal attack phrases are transferable across models; phrases optimized on smaller surrogate models (e.g., FlanT5-xl~\cite{chung2024scaling}) can successfully deceive larger models like GPT-3.5 and Llama2~\cite{touvron2023llama}.

Furthermore, Shi et al.~\cite{shi2024optimization} introduced \textit{JudgeDeceiver}, an optimization-based prompt injection attack tailored for the LLMs-as-judges framework. Unlike handcrafted methods, JudgeDeceiver formulates a precise optimization objective to efficiently generate adversarial sequences. These sequences can mislead LLMs-as-judges into selecting biased or incorrect responses among candidate answers, thereby compromising the evaluation process.

Although preliminary studies~\cite{shi2024optimization, raina2024llm} have highlighted the vulnerability of LLMs-as-judges to adversarial manipulations, this field remains largely underexplored. 
It is imperative to advance our understanding of these weaknesses and devise effective defense strategies. 
As the use of LLMs-as-judges grows across diverse applications, future research should focus on uncovering new attack methods and strengthening the models against such adversarial threats.

\subsection{Inherent Weaknesses}
\label{sec:weaknesses}
Despite the remarkable capabilities of LLMs, they possess several inherent weaknesses that can compromise their reliability and robustness in LLMs-as-judges. This subsection discusses key limitations, including issues related to knowledge recency, hallucination, and other domain-specific knowledge gaps~\cite{zhao2023survey}.

\subsubsection{Knowledge Recency}
\label{sec:Knowledge Recency}
One significant limitation of LLMs is their inability to access or incorporate up-to-date information reliably. LLMs are generally trained on static datasets that may become outdated over time, limiting their ability to evaluate scenarios that require knowledge of recent events, legislation, or rapidly evolving fields. The most straightforward solution is to retrain the model on new data; however, this approach is resource-intensive and risks catastrophic forgetting~\cite{luo2023empirical}, where previously learned knowledge is overwritten during training.
This temporal disconnect can lead to judgments based on invalid data, or obsolete practices, compromising their reliability in real-world, time-sensitive applications. Consider a case where LLMs-as-judges are used to evaluate which of two responses from LLMs better answers a prompt about the COVID-19 pandemic. Suppose one response references the WHO guidelines updated timely, while the other relies on outdated 2020 guidelines. If the LLM-as-Judge has not been updated with the latest guidelines, it might erroneously prefer the outdated response, incorrectly deeming it more accurate. This failure to account for recent developments highlights the importance of addressing knowledge recency in LLMs-as-judges.

Addressing the issue of knowledge recency can involve integrating retrieval-augmented generation (RAG) methods~\cite{gao2023retrieval,lewis2020retrieval}, which enable LLMs to query external, dynamically updated databases or knowledge sources during evaluation. 
Additionally, periodic fine-tuning with updated datasets or leveraging continual learning frameworks~\cite{wu2024continual} can ensure that LLMs-as-judges remain aligned with the latest information. Combining these approaches with robust fact-checking mechanisms~\cite{dierickx2024striking} can further enhance temporal reliability in judgment contexts.

\subsubsection{Hallucination}
\label{sec:Hallucination}
Another critical issue in LLMs is the hallucination problem, where models generate incorrect or fabricated information with high confidence. In the context of LLMs-as-judges, hallucination can manifest as the invention of non-existent precedents, misinterpretation of facts, or fabrication of sources, which can severely undermine the reliability of their judgments. This issue is particularly concerning in various applications, where such errors can lead to unfair or harmful outcomes.

Employing fact-checking mechanisms~\cite{dierickx2024striking,ji2023survey,tonmoy2024comprehensive} during evaluation is crucial to mitigate hallucination. By cross-verifying the outputs of LLMs-as-judges with trusted databases and external knowledge sources, hallucinated information can be identified and corrected.

\subsubsection{Domain-Specific Knowledge Gaps}
\label{sec:Domain-Specific Knowledge Gaps}
While LLMs demonstrate broad generalization capabilities, they often lack the depth of understanding required for specialized domains~\cite{feng2023knowledge,pan2024unifying,szymanski2024limitations,dorner2024limitsscalableevaluationfrontier}. For instance, legal judgments demand intricate knowledge of statutes, precedents, and contextual nuances, which may not be adequately captured in the training data of general-purpose LLMs. This limitation can lead to shallow or incorrect judgments in domain-specific contexts.

Domain adaptation techniques, such as integrating LLMs with domain-specific knowledge graphs~\cite{feng2023knowledge,pan2024unifying} or leveraging RAG systems~\cite{gao2023retrieval}, can substantially improve their performance in specialized domains. Knowledge graphs provide structured, expert-curated information that enhances context-awareness, while RAG enables LLMs to dynamically retrieve relevant knowledge from specific domain.

The inherent weaknesses of LLMs highlight the need for continued research and innovation. Addressing these limitations through RAG, enhanced training methods, and knowledge graph techniques is crucial for ensuring that LLMs-as-judges deliver reliable, accurate, and trustworthy evaluations in diverse applications.

\section{Future Work}
\label{sec:Future}
In this section, we will explore the key directions for future work, focusing on how to build more efficient, more effective, and more reliable LLM judges. These directions aim to address the bottlenecks and challenges in current technologies and practices, while also promoting broader applications and deeper integration of LLM judges in diverse scenarios.

\subsection{More Efficient LLMs-as-Judges}
\label{sec:more-Efficient}

\subsubsection{Automated Construction of Evaluation Criteria and Tasks}

Current LLM judges often rely on manually predefined evaluation criteria, lacking the ability to adapt dynamically during the assessment process. Designing prompts for these systems is not only tedious and time-consuming but also struggles to address the diverse requirements of various task scenarios~\cite{bai2024benchmarking,zhao2024auto,yu2024kieval,zhang2024talec,wang2024revisiting}. To overcome these limitations, future LLM judges could incorporate enhanced adaptability by tailoring evaluation criteria based on task types, target audiences, and domain-specific knowledge. Such advancements would significantly streamline the configuration process of LLM judges, while also greatly improving their practicality and efficiency in real-world applications.

Moreover, existing static evaluation datasets are prone to issues such as training data contamination, which can compromise their effectiveness in accurately assessing the evolving capabilities of LLMs. To address this, future LLM judges could focus on dynamically constructing more suitable evaluation tasks and continuously optimizing the evaluation process, thereby enhancing applicability and precision~\cite{bai2024benchmarking,zhao2024auto}.

\subsubsection{Scalable Evaluation Systems}
Existing LLM judges often exhibit limited adaptability in practical applications. While these judges may perform effectively on specific downstream tasks, they frequently struggle in cross-domain or multi-task settings, thereby falling short of meeting the diverse and broader demands of real-world applications.

To address these limitations, future research could focus on modular design principles to create scalable evaluation frameworks~\cite{xu2024perfect}. Such frameworks would allow users to flexibly add or customize evaluation modules to suit their specific needs. This modular approach not only enhances the usability and flexibility of the system but also significantly reduces the cost and complexity of transferring the framework across different domains.

\subsubsection{Accelerating Evaluation Processes}

Existing LLM systems often face significant computational costs when performing evaluation tasks. For example, pairwise comparison methods require multiple rounds of comparisons for each candidate, which becomes extremely time-consuming as the number of candidates grows.  In resource-constrained environments, such high-cost evaluation methods are challenging to deploy effectively. To address this issue, future research could focus on developing more efficient candidate selection algorithms, thereby unlocking new opportunities for the use of LLMs in low-resource settings~\cite{lee2024aligning,liu2024aligning}.

Similarly, the multi-LLM evaluation paradigm, which relies on multiple rounds of interaction, further exacerbates computational demands. To mitigate these challenges, future efforts could explore streamlined communication frameworks that support high-quality evaluation tasks while minimizing resource requirements~\cite{chen2024internet}. Advances in these areas could lead to the development of more efficient and scalable evaluation systems, making LLM-based evaluations more practical across diverse and resource-limited scenarios.

\subsection{More Effective LLMs-as-Judges}
\label{sec:more-Effective}

\subsubsection{Integration of Reasoning and Judge Capabilities}

Current LLMs-as-judges systems often treat reasoning and evaluation capabilities as distinct and independent modules, which can hinder effectiveness when addressing complex tasks. As the demand for evaluating increasingly complex systems grows, future LLM-as-Judge systems should prioritize the deep integration of reasoning and evaluation capabilities to achieve a seamless synergy~\cite{zhuo2023ice,yi2024protocollm,stephan2024calculation}. For instance, in legal scenarios, the model could first infer the relevant legal provisions and then assess the case's relevance, making the evaluation process more effective.

\subsubsection{Establishing a Collective Judgment Mechanism}
Current LLMs-as-Judge systems typically rely on a single model for evaluation. While this approach is straightforward, it is prone to biases inherent in individual models, leading to reduced accuracy and stabilitys. Moreover, a single model often struggles to comprehensively address the diverse requirements of such tasks. Future research could investigate collaborative multi-agent mechanisms to enable ``collective judgment'' where multiple LLMs work together, leveraging their respective strengths in reasoning and knowledge~\cite{chan2023chateval,chu2024pre},. Additionally, ensemble techniques could be employed to dynamically balance the contributions of different models, leading to more stable and reliable judgment outcomes.

\subsubsection{Enhancing Domain Knowledge}

Current LLMs-as-Judge systems often fall short when handling tasks in specialized fields due to insufficient domain knowledge. Furthermore, as domain knowledge continues to evolve, these models struggle to keep up with the latest developments, further limiting their effectiveness and applicability in real-world scenarios.

To address these challenges, future LLMs-as-judges systems should focus on integrating comprehensive domain knowledge to enhance their performance in specialized tasks~\cite{raju2024constructing}. This can be achieved by utilizing knowledge graphs, embedding domain-specific expertise, and fine-tuning models based on feedback from subject-matter experts.
In addition, these systems should incorporate dynamic knowledge updating capabilities. For instance, in the legal domain, models could regularly acquire and integrate updates on new statutes, case law, and policy changes, ensuring that their judgments remain current and aligned with the latest legal standards.

\subsubsection{Cross-Domain and Cross-Language Transferability}
Current LLMs-as-Judge systems are often confined to specific domains or languages, making it challenging for them to transfer across different fields. For instance, an LLM proficient in processing legal texts may struggle to effectively handle evaluation tasks in the medical or financial domains. This limitation greatly restricts the applicability of such systems.

Future research can focus on exploring cross-domain and cross-language transfer learning techniques to enhance the adaptability of LLMs in diverse fields. By leveraging shared general knowledge across fields, models can quickly adapt to new tasks with minimal additional training costs~\cite{son2024mm,hada2023large,watts2024pariksha}. For example, evaluation capabilities developed in English could be transferred to contexts in German, thereby improving the evaluation performance in these new areas.

\subsubsection{Multimodal Integration Evaluation}
Current LLM-as-Judge systems primarily focus on processing textual data, with limited attention to integrating other modalities like images, audio, and video. This single-modal approach falls short in complex scenarios requiring multimodal analysis, such as combining visual and textual information in medical assessments. Future systems should develop cross-modal integration capabilities to process and evaluate multimodal data simultaneously~\cite{chen2024mllm}. Leveraging cross-modal validation can enhance evaluation accuracy. Key research areas include efficient multimodal feature extraction, integration, and the design of unified frameworks for more comprehensive and precise evaluations.

\subsection{More Reliable LLMs-as-Judges}
\label{sec:more-Reliable}

\subsubsection{Enhancing Interpretability and Transparency}

Current LLM-as-Judge systems often operate as black boxes, with their rulings lacking transparency and a clear reasoning process. This opacity is particularly concerning in high-stakes domains such as legal judgments, where users cannot fully understand the basis of the model's decisions or trust its outputs.
Future research should focus on improving the interpretability of LLMs~\cite{liu2024hd}. For example, the LLM judges should not only provide evaluation results but also present a clear explanation. Research could explore designing validation models based on logical frameworks to make the decision-making process more transparent.

\subsubsection{Mitigating Bias and Ensuring Fairness}
LLMs may be influenced by biases present in their training data, leading to unfair judgments in different social, cultural, or legal contexts. These biases could be amplified by the model and compromise the fairness of its decisions. Future research could focus on ensuring fairness in model outputs through debiasing algorithms and fairness constraints~\cite{li2024calibraeval}. Targeted approaches, such as adversarial debiasing training or bias detection tools, can dynamically identify and mitigate potential biases during the model's reasoning process.

\subsubsection{Enhancing Robustness}
LLMs are sensitive to noise, incompleteness, or ambiguity in input instructions, which may lead to errors or instability in evaluation results when handling complex or highly uncertain texts. This lack of robustness significantly limits their reliability in practical applications.
Future research can adopt several methods to enable LLMs robust and reliable performance in real-world environments~\cite{shi2024optimization,elangovan2024beyond}. For instance, introducing more advanced data augmentation techniques to generate diverse and uncertain simulated cases can help train models to adapt to various complex input conditions.

\section{Conclusion}
\label{sec:Conclusion}
This survey systematically examined the LLMs-as-Judges framework across five dimensions: functionality, methodology, applications, meta-evaluation, and limitations, providing a comprehensive understanding of its advantages, limitations, practical implementations, applications, and methods for evaluating its effectiveness.
To advance research in this field, we also outlined several promising directions for future exploration, including the development of more efficient, effective, and reliable LLM judges. We hope to promote the ongoing development of this field by providing foundational resources and will continue to update relevant content.

\newpage
\bibliographystyle{ACM-Reference-Format}
\bibliography{sample-base}

\end{document}